\documentclass[10pt,twocolumn,letterpaper]{article}

\usepackage{cvpr}
\usepackage{times}
\usepackage{epsfig}
\usepackage{graphicx}
\usepackage{amsmath}
\usepackage{amssymb}

\usepackage{xcolor, soul}
\sethlcolor{pink}

\usepackage{xparse, mathtools, array}
\DeclarePairedDelimiterX{\rvect}[1]{[}{]}{\,\makervect{#1}\,}
\ExplSyntaxOn
\NewDocumentCommand{\makervect}{m}
 {
  \seq_set_split:Nnn \l_tmpa_seq { , } { #1 }
  \begin{matrix}
  \seq_use:Nn \l_tmpa_seq { & }
  \end{matrix}
 }
\ExplSyntaxOff

\newcommand{\Transp}{\mathsf{T}}

\usepackage{caption}
\usepackage{subcaption}

\usepackage[title]{appendix}

\usepackage{booktabs}
\usepackage{multirow}

\usepackage{algorithm}
\usepackage[noend]{algpseudocode}
\makeatletter
\def\BState{\State\hskip-\ALG@thistlm}
\makeatother
\algnewcommand\Or{\textbf{or}}

\usepackage{amsmath,amsfonts,bm}

\def\eqref#1{equation~\ref{#1}}

\def\1{\bm{1}}

\DeclareMathAlphabet{\mathsfit}{\encodingdefault}{\sfdefault}{m}{sl}
\SetMathAlphabet{\mathsfit}{bold}{\encodingdefault}{\sfdefault}{bx}{n}

\newcommand{\parsection}[1]{\vspace{2mm}\noindent\textbf{#1}~ }

\newcolumntype{C}[1]{>{\centering\let\newline\\\arraybackslash\hspace{0pt}}m{#1}}

\usepackage{pgfplots}
    \usepgfplotslibrary{colorbrewer}
    \pgfplotsset{
        cycle list/Dark2,
        cycle multiindex* list={
            mark list*\nextlist
            Dark2\nextlist
        },
    }
\pgfplotsset{compat=1.14}

\newcommand{\Softmax}{\textrm{Softmax}}
\newcommand{\Cat}{\textrm{Cat}}

\usepackage[pagebackref=true,breaklinks=true,letterpaper=true,colorlinks,bookmarks=false]{hyperref}

\cvprfinalcopy 

\def\cvprPaperID{1} 

\begin{document}

\title{Evaluating Scalable Bayesian Deep Learning\\ Methods for Robust Computer Vision}


\author{Fredrik K.~Gustafsson$^{1}$
\and
Martin Danelljan$^{2}$
\and
Thomas B.~Sch\"on$^{1}$ \vspace{1.0mm}
\and
$^{1}$Department of Information Technology, Uppsala University, Sweden
\and
$^{2}$Computer Vision Lab, ETH Zurich, Switzerland
}

\maketitle

\begin{abstract}
    While deep neural networks have become the go-to approach in computer vision, the vast majority of these models fail to properly capture the uncertainty inherent in their predictions. Estimating this predictive uncertainty can be crucial, for example in automotive applications. In Bayesian deep learning, predictive uncertainty is commonly decomposed into the distinct types of aleatoric and epistemic uncertainty. The former can be estimated by letting a neural network output the parameters of a certain probability distribution. Epistemic uncertainty estimation is a more challenging problem, and while different scalable methods recently have emerged, no extensive comparison has been performed in a real-world setting. We therefore accept this task and propose a comprehensive evaluation framework for scalable epistemic uncertainty estimation methods in deep learning. Our proposed framework is specifically designed to test the robustness required in real-world computer vision applications. We also apply this framework to provide the first properly extensive and conclusive comparison of the two current state-of-the-art scalable methods: ensembling and MC-dropout. Our comparison demonstrates that ensembling consistently provides more reliable and practically useful uncertainty estimates. Code is available at \url{https://github.com/fregu856/evaluating_bdl}.
\end{abstract}

\section{Introduction}
\label{introduction}

\begin{figure*}[t]
\newcommand{\wid}{1.6cm}%
\newcommand{\imwid}{0.333\textwidth}%
\centering
    \begin{minipage}{0.04125\textwidth}
		\vspace{5mm}%
		\rotatebox{90}{\resizebox{7cm}{!}{%
            \begin{tabular}{ C{\wid} C{\wid} C{\wid} C{\wid} }
                \multicolumn{2}{c}{\textbf{Semantic Segmentation}} & \multicolumn{2}{c}{\textbf{Depth Completion}} \\
                Real & Synthetic & Real & Synthetic
            \end{tabular}}}%
    \end{minipage}%
    \begin{minipage}{0.78375\textwidth}
			\begin{tabular}{@{}C{\imwid}@{}C{\imwid}@{}C{\imwid}@{}}
				Input & Prediction & Predictive Uncertainty 
	\end{tabular}
      \includegraphics[width=\imwid]{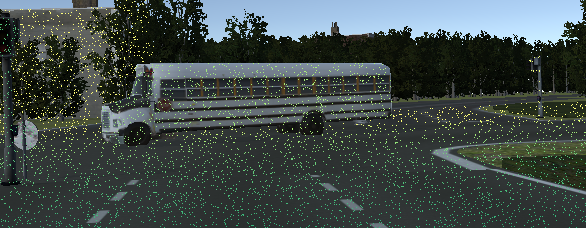}%
      \includegraphics[width=\imwid]{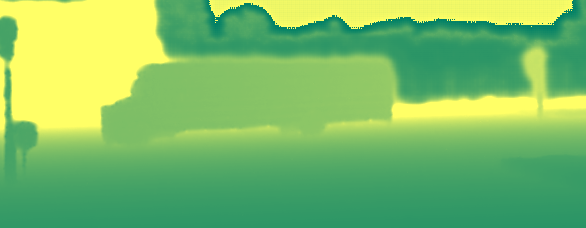}%
      \includegraphics[width=\imwid]{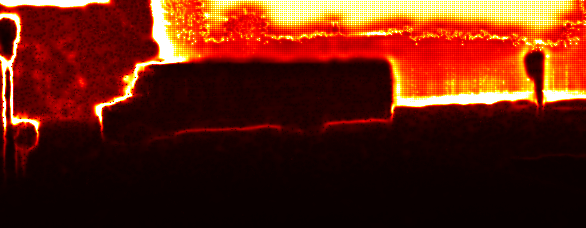}
	     \includegraphics[width=\imwid]{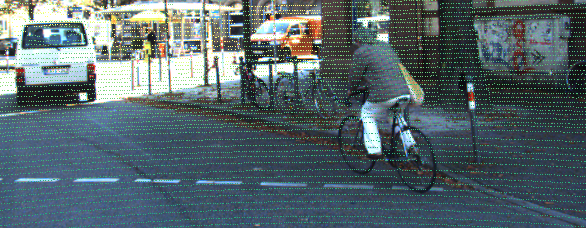}%
	     \includegraphics[width=\imwid]{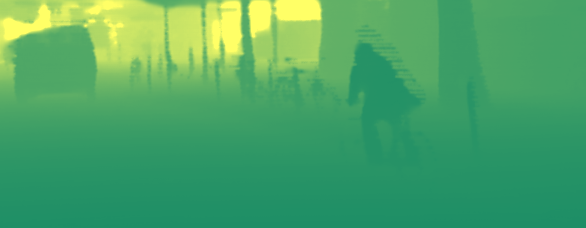}%
	     \includegraphics[width=\imwid]{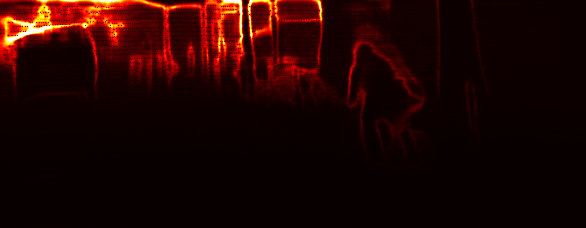}\vspace{1mm}
     	\includegraphics[width=\imwid]{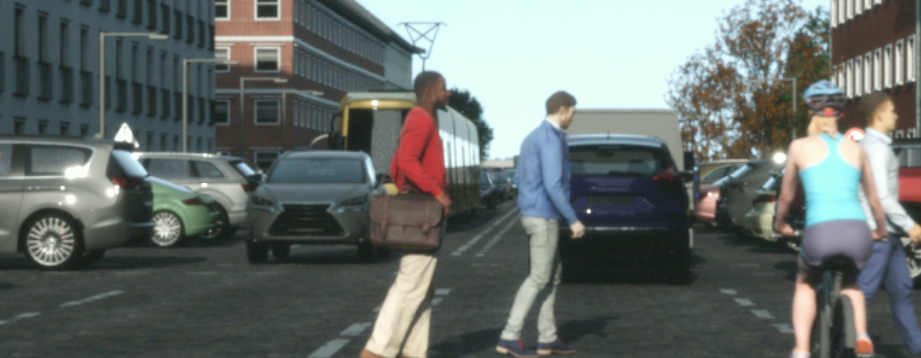}%
     	\includegraphics[width=\imwid]{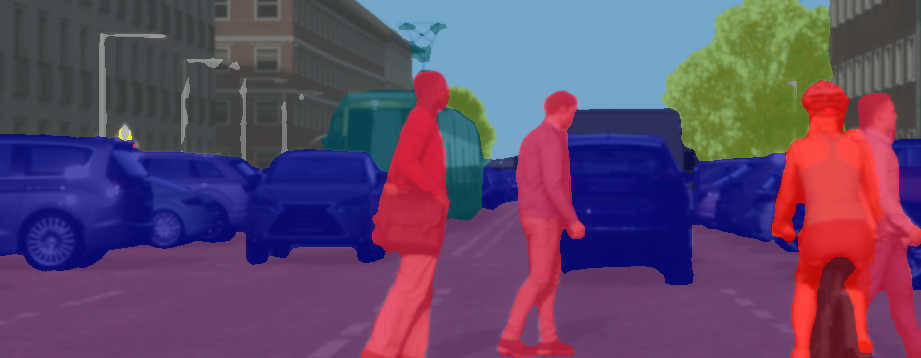}%
     	\includegraphics[width=\imwid]{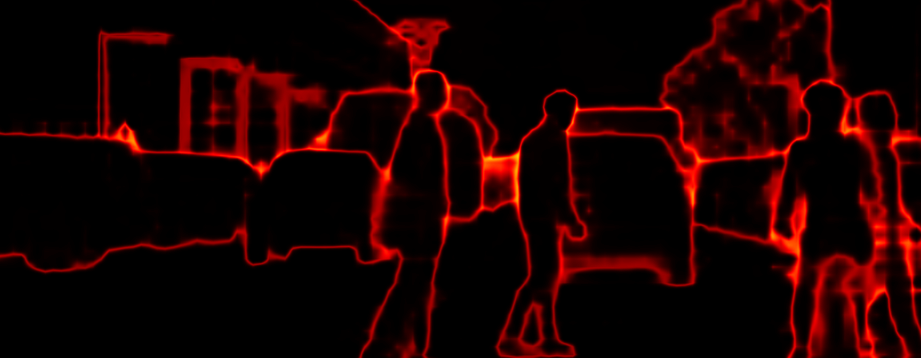}
     	\includegraphics[width=\imwid]{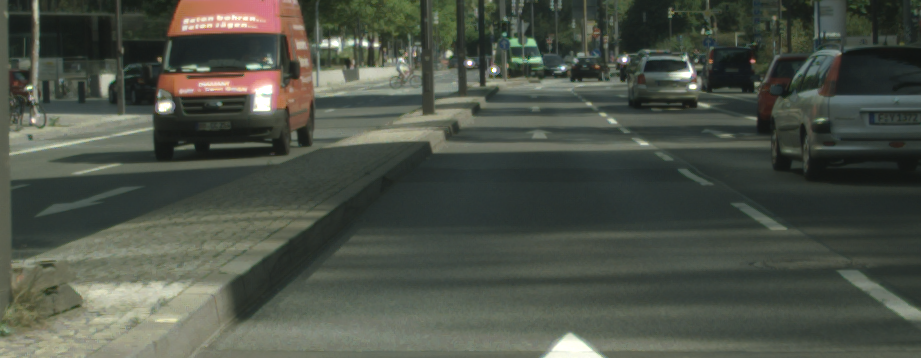}%
     	\includegraphics[width=\imwid]{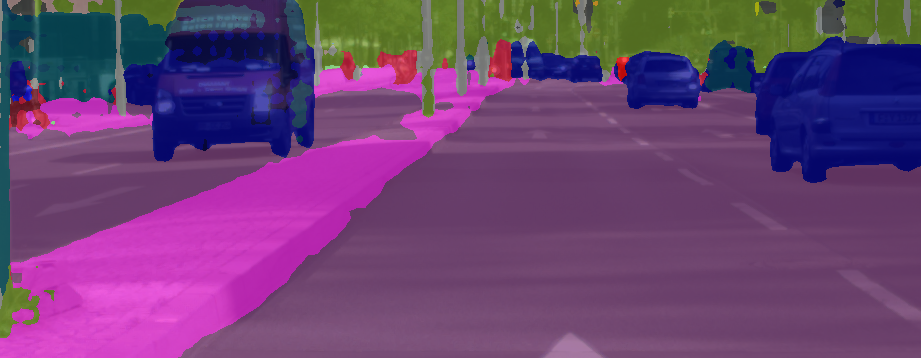}%
     	\includegraphics[width=\imwid]{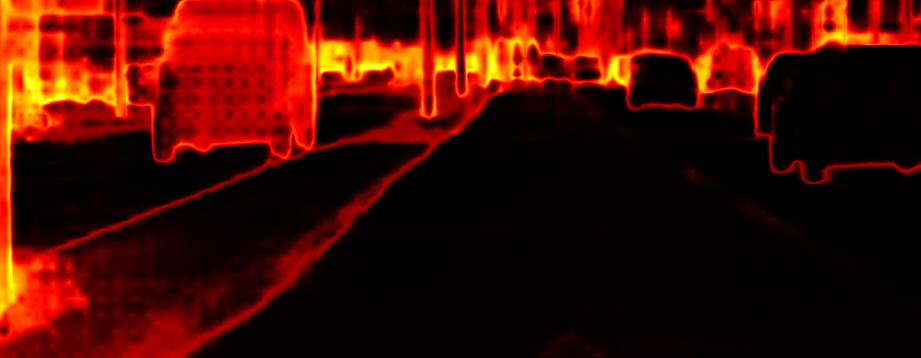}
\end{minipage}
\vspace{-0.5mm}
\caption{We propose a comprehensive evaluation framework for \emph{scalable} epistemic uncertainty estimation methods in deep learning. The proposed framework employs state-of-the-art DNN models on the tasks of depth completion and street-scene semantic segmentation. All models are trained exclusively on synthetic data (the Virtual KITTI~\cite{Gaidon:Virtual:CVPR2016} and Synscapes~\cite{wrenninge2018synscapes} datasets). We here show the input (left), prediction (center) and estimated predictive uncertainty (right) for ensembling with $M=8$ ensemble members, on both synthetic and real (the KITTI~\cite{Geiger2013IJRR, Uhrig2017THREEDV} and Cityscapes~\cite{cordts2016cityscapes} datasets) example validation images. Black pixels correspond to minimum predictive uncertainty, white pixels to maximum uncertainty.}\vspace{-3.5mm}
\label{fig:overview}
\end{figure*}

Deep Neural Networks (DNNs) have become the standard paradigm within most computer vision problems due to their astonishing predictive power compared to previous alternatives. Current applications include many safety-critical tasks, such as street-scene semantic segmentation~\cite{cordts2016cityscapes, chen2017rethinking, yuan2018ocnet}, 3D object detection~\cite{shi2018pointrcnn, lang2018pointpillars} and depth completion~\cite{Uhrig2017THREEDV, ma2018self}. Since erroneous predictions can have disastrous consequences, such applications require an accurate measure of the predictive uncertainty. The vast majority of these DNN models do however fail to properly capture the uncertainty inherent in their predictions. They are thus not fully capable of the type of \emph{uncertainty-aware} reasoning that is highly desired e.g. in automotive applications.

The approach of \emph{Bayesian deep learning} aims to address this issue in a principled manner. Here, predictive uncertainty is commonly decomposed into two distinct types, which both should be captured by the learned DNN~\cite{gal2016thesis, kendall2017uncertainties}. \emph{Epistemic} uncertainty accounts for uncertainty in the DNN model parameters, while \emph{aleatoric} uncertainty captures inherent and irreducible data noise. Input-dependent \emph{aleatoric} uncertainty about the target $y$ arises due to e.g. noise and ambiguities inherent in the input $x$. This is present for instance in street-scene semantic segmentation, where image pixels at object boundaries are inherently ambiguous, and in 3D object detection where the location of a distant object is less certain due to noise and limited sensor resolution. In many computer vision applications, this aleatoric uncertainty can be effectively estimated by letting a DNN directly output the parameters of a certain probability distribution, modeling the conditional distribution $p(y | x)$ of the target given the input. For classification tasks, a predictive categorical distribution is commonly realized by a softmax output layer, although recent work has also explored Dirichlet models~\cite{gast2018lightweight, sensoy2018evidential, malinin2018predictive}. For regression, Laplace and Gaussian models have been employed~\cite{ilg2018uncertainty, chua2018deep, kendall2017uncertainties, lakshminarayanan2017simple}. 

Directly predicting the conditional distribution $p(y | x)$ with a DNN does however not capture \emph{epistemic} uncertainty, as information about the uncertainty in the model parameters is disregarded. This often leads to highly confident predictions that are incorrect, especially for inputs $x$ that are not well-represented by the training distribution~\cite{guo2017calibration, lakshminarayanan2017simple}. For instance, a DNN can fail to generalize to unfamiliar weather conditions or environments in automotive applications, but still generate confident predictions. Reliable estimation of epistemic uncertainty is thus of great importance. However, this task has proven to be highly challenging, largely due to the vast dimensionality of the parameter space, which renders standard Bayesian inference approaches intractable. To tackle this problem, a wide variety of approximations have been explored \cite{neal1995bayesian, hinton1993keeping, blundell2015weight, welling2011bayesian, chen2014stochastic, hernandez2015probabilistic, wang2016natural}, but only a small number have been demonstrated to be applicable even to the large-scale DNN models commonly employed in \emph{real-world} computer vision tasks. Among such \emph{scalable} methods, MC-dropout \cite{gal2016thesis, kendall2017uncertainties, kendall2015bayesian, mukhoti2018evaluating} and ensembling \cite{lakshminarayanan2017simple, chua2018deep, ilg2018uncertainty} are clearly the most widely employed, due to their demonstrated effectiveness and simplicity. While \emph{scalable} techniques for epistemic uncertainty estimation recently have emerged, the research community however lacks a common and comprehensive evaluation framework for such methods. Consequently, both researchers and practitioners are currently unable to properly assess and compare newly proposed methods. In this work, we therefore accept this task and set out to design exactly such an evaluation framework, aiming to benefit and inspire future research in the field.

Previous studies have provided only partial insight into the performance of different \emph{scalable} methods for epistemic uncertainty estimation. Kendall and Gal~\cite{kendall2017uncertainties} evaluated MC-dropout alone on the tasks of semantic segmentation and monocular depth regression, providing mainly qualitative results. Lakshminarayanan et al.~\cite{lakshminarayanan2017simple} introduced ensembling as a non-Bayesian alternative and found it to generally outperform MC-dropout. Their experiments were however based on relatively small-scale models and datasets, limiting the real-world applicability. Ilg et al.~\cite{ilg2018uncertainty} compared ensembling and MC-dropout on the task of optical-flow estimation, but only in terms of the AUSE metric which is a \emph{relative} measure of the uncertainty estimation quality. While finding ensembling to be advantageous, their experiments were also limited to a fixed number ($M = 8$) of ensemble members and MC-dropout forward passes, not allowing a completely fair comparison. Ovadia et al.~\cite{snoek2019can} also fixed the number of ensemble members, and moreover only considered classification tasks. We improve upon this previous work and propose an evaluation framework that actually enables a conclusive ranking of the compared methods.

\parsection{Contributions}
We propose a comprehensive evaluation framework for \emph{scalable} epistemic uncertainty estimation methods in deep learning. The proposed framework is specifically designed to test the robustness required in \emph{real-world} computer vision applications, and employs state-of-the-art DNN models on the tasks of depth completion (regression) and street-scene semantic segmentation (classification).\ It also employs a novel combination of quantitative evaluation metrics which explicitly measures the reliability and practical usefulness of estimated predictive uncertainties. We apply our proposed framework to provide the first properly extensive and conclusive comparison of the two current state-of-the-art \emph{scalable} methods: ensembling and MC-dropout.\ This comparison demonstrates that ensembling consistently outperforms the highly popular MC-dropout method. Our work thus suggests that ensembling should be considered the new go-to approach, and encourages future research to understand and further improve its efficacy. Figure~\ref{fig:overview} shows example predictive uncertainty estimates generated by ensembling. Our framework can also directly be applied to compare other \emph{scalable} methods, and we encourage external usage with publicly available code.

In our proposed framework, we predict the conditional distribution $p(y | x)$ in order to estimate input-dependent aleatoric uncertainty. The methods for epistemic uncertainty estimation are then compared by quantitatively evaluating the estimated predictive uncertainty in terms of the relative AUSE metric and the \emph{absolute} measure of uncertainty calibration. Our evaluation is the first to include \emph{both} these metrics, and furthermore we apply them to both regression and classification tasks. To provide a deeper and more fair analysis, we also study all metrics as functions of the number of samples $M$, enabling a highly informative comparison of the rate of improvement. Moreover, we simulate challenging real-world conditions found e.g.\ in automotive applications, where robustness to out-of-domain inputs is required to ensure safety, by training our models exclusively on synthetic data and evaluating the predictive uncertainty on real-world data. By analyzing this important domain shift problem, we significantly increase the practical applicability of our evaluation. We also complement our real-world analysis with experiments on illustrative toy regression and classification problems. Lastly, to demonstrate the evaluation rigor necessary to achieve a conclusive comparison, we repeat each experiment multiple times and report results together with the observed variation.
\section{Predictive Uncertainty Estimation using Bayesian Deep Learning}
\label{predictive_uncertainty_estimation}

DNNs have been shown to excel at a wide variety of supervised machine learning problems, where the task is to predict a target value $y \in \mathcal{Y}$ given an input $x \in \mathcal{X}$. In computer vision, the input space $\mathcal{X}$ often corresponds to the space of images. For classification problems, the target space $\mathcal{Y}$ consists of a finite set of $C$ classes, while a regression problem is characterized by a continuous target space, e.g. $\mathcal{Y}~=~\mathbb{R}^K$. For our purpose, a DNN is defined as a function $f_{\theta}: \mathcal{X} \rightarrow \mathcal{U}$, parameterized by $\theta \in \mathbb{R}^{P}$, that maps an input $x \in \mathcal{X}$ to an output $f_{\theta}(x) \in \mathcal{U}$. Next, we cover alternatives for estimating both the aleatoric and epistemic uncertainty in the predictions of DNN models.

\begin{figure*}
\newcommand{\wid}{0.221\textwidth}
     \centering
     \begin{subfigure}[b]{\wid}
         \centering
         \includegraphics[width=\textwidth]{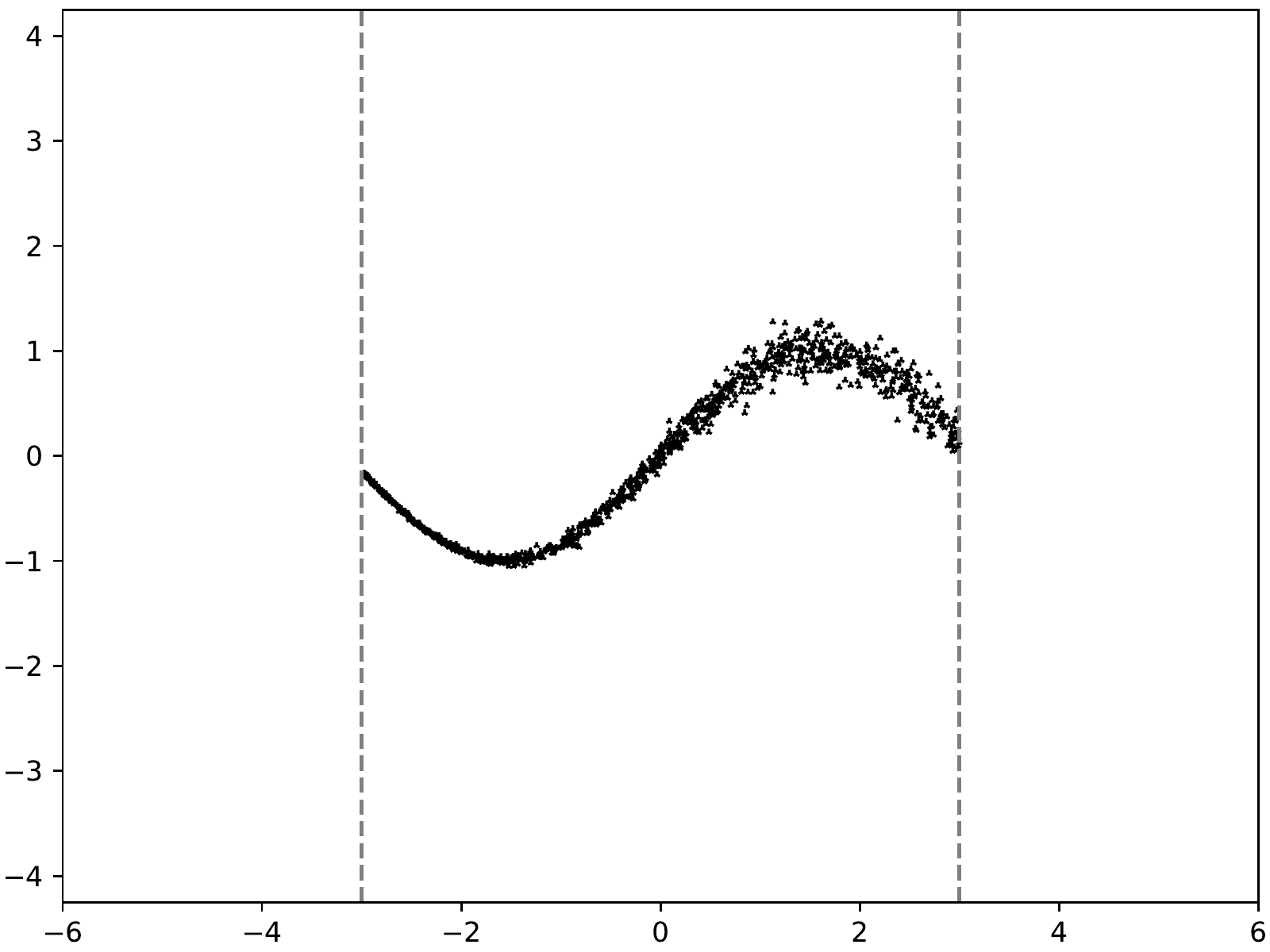}\vspace{-1mm}%
         \caption{}
        \label{fig:illustrative_1d_regression_training_dataset}
     \end{subfigure}
     \vspace{-0.75mm}
     \begin{subfigure}[b]{\wid}
         \centering
         \includegraphics[width=\textwidth]{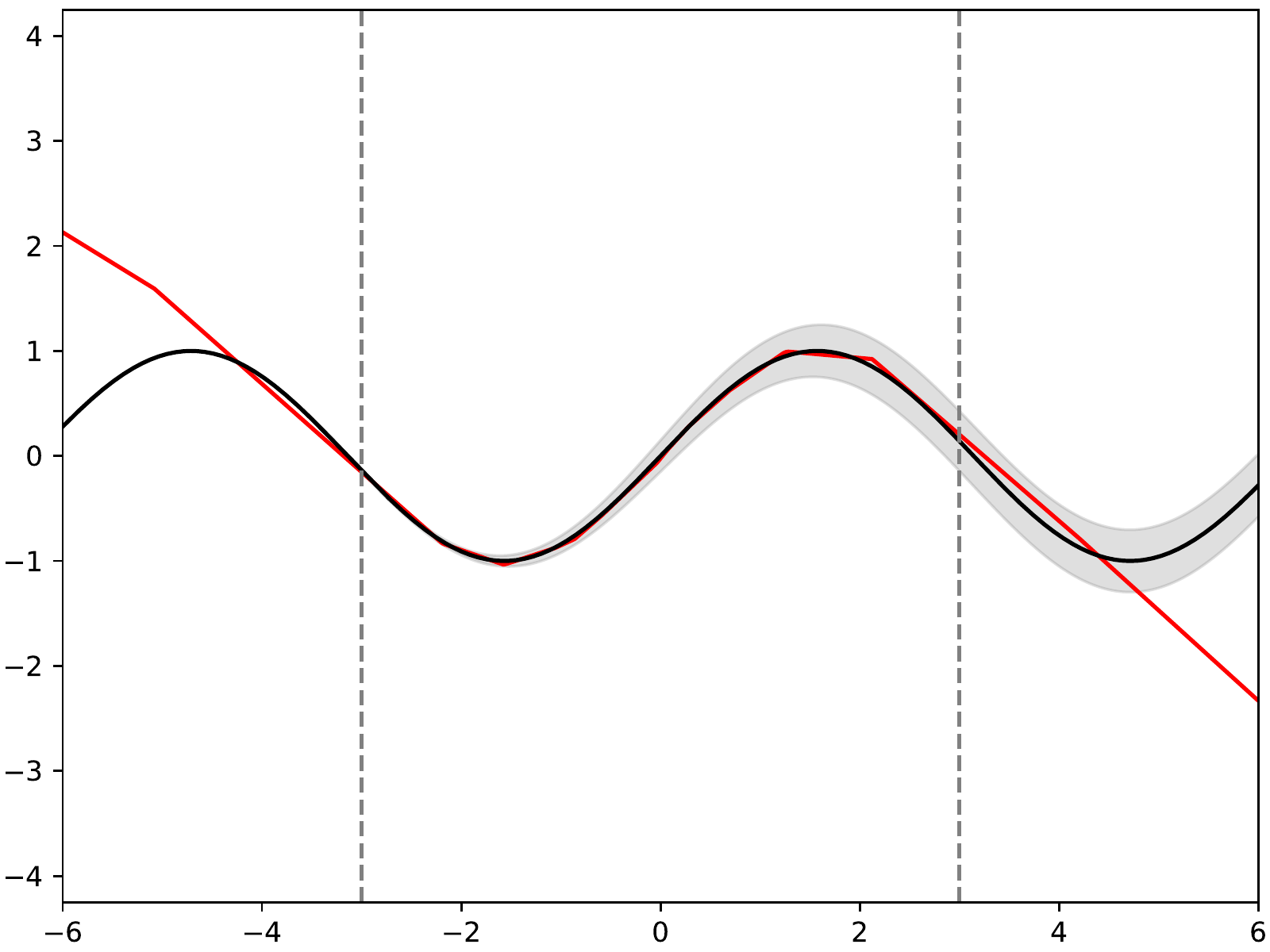}\vspace{-1mm}%
         \caption{}
        \label{fig:illustrative_1d_regression_deterministic}
     \end{subfigure}
     \vspace{-0.75mm}
     \begin{subfigure}[b]{\wid}
         \centering
         \includegraphics[width=\textwidth]{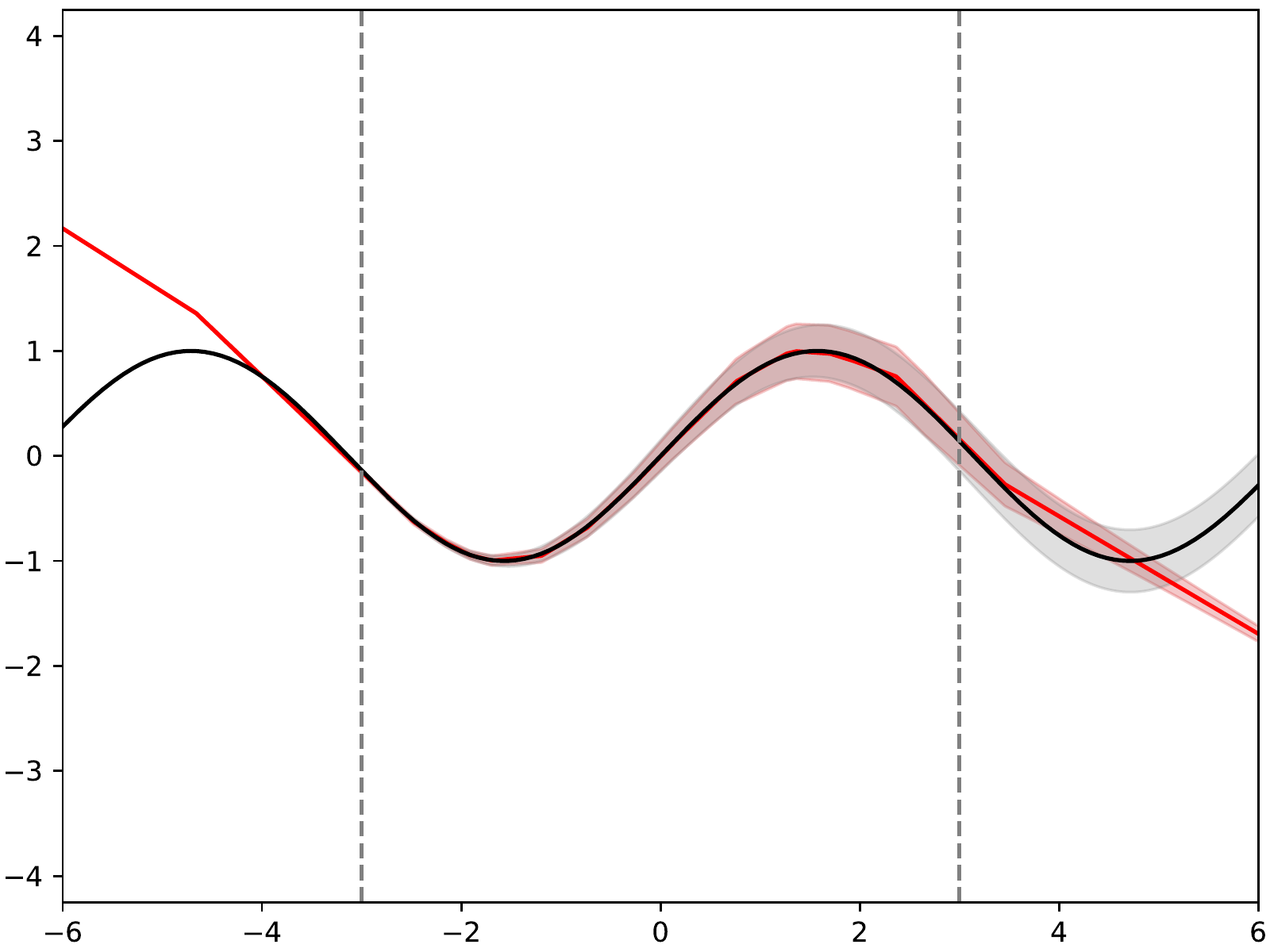}\vspace{-1mm}%
         \caption{}
         \label{fig:illustrative_1d_regression_mle}
     \end{subfigure}
     \vspace{-0.75mm}
     \begin{subfigure}[b]{\wid}
         \centering
         \includegraphics[width=\textwidth]{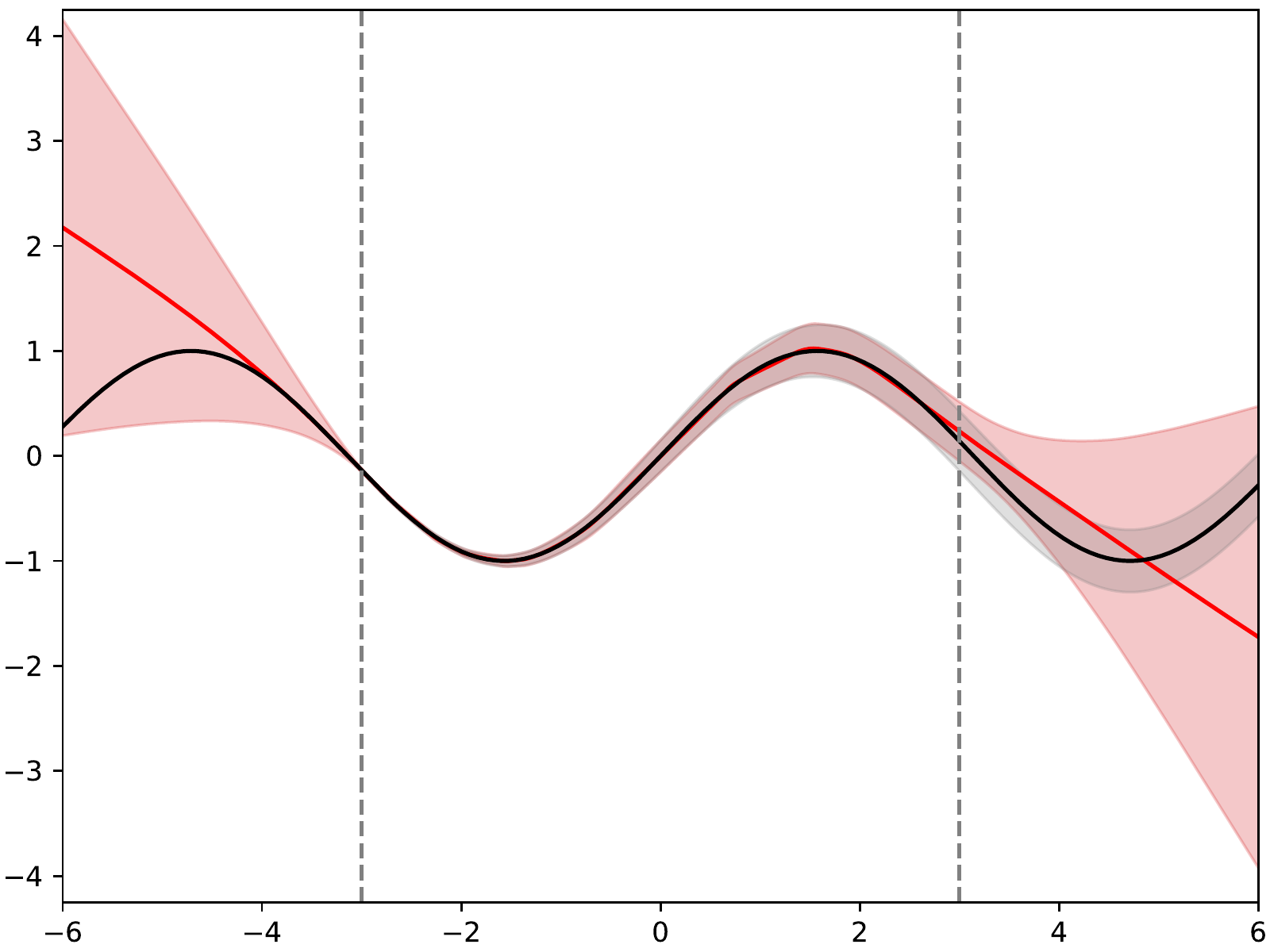}\vspace{-1mm}%
         \caption{}
         \label{fig:illustrative_1d_regression_bayesian}
     \end{subfigure}
     \vspace{-0.75mm}
        \caption{Toy regression problem illustrating the task of predictive uncertainty estimation with DNNs. The true data generator $p(y|x)$ is a Gaussian, where the mean is given by the solid black line and the variance is represented in shaded gray. The predictive mean and variance are given by the solid red line and the shaded red area, respectively. \textbf{(a)} Training dataset with $N\!=\!1\thinspace000$ examples. \textbf{(b)} A DNN trained to directly predict the target $y$ captures no notion of uncertainty. \textbf{(c)} A corresponding Gaussian DNN model (\ref{eq:gaussian_model}) trained via maximum-likelihood captures \emph{aleatoric} but not epistemic uncertainty. \textbf{(d)} The Gaussian model instead trained via approximate Bayesian inference~(\ref{eq:approx_predictive_distribution_Bayesian}) captures both \emph{aleatoric} and \emph{epistemic} uncertainty.}
        \label{fig:illustrative_1d_regression}\vspace{-3.5mm}
\end{figure*}

\parsection{Aleatoric Uncertainty}
In classification problems, aleatoric uncertainty is commonly captured by predicting a categorical distribution $p(y | x, \theta)$. This is implemented by letting the DNN predict logit scores $f_{\theta}(x) \in \mathbb{R}^C$, which are then normalized by a $\Softmax$ function,
\begin{equation}
\begin{gathered}
    p(y | x, \theta) = \Cat(y; s_{\theta}(x)),\\
    s_{\theta}(x) = \Softmax(f_{\theta}(x)).
\label{eq:categorical_model}
\end{gathered}
\end{equation}
Given a training set of i.i.d.\ sample pairs $\mathcal{D} = \{X, Y\} = \{(x_i, y_i)\}_{i=1}^{N}$, $(x_i, y_i) \sim p(x, y)$, the data likelihood is obtained as $p(Y | X, \theta) = \prod_{i=1}^{N} p(y_i | x_i, \theta)$. The maximum-likelihood estimate of the model parameters, $\hat{\theta}_{\mathrm{MLE}}$, is obtained by minimizing the negative log-likelihood $- \sum_i \log p(y_i | x_i, \theta)$. For the Categorical model~(\ref{eq:categorical_model}), this is equivalent to minimizing the well-known cross-entropy loss. At test time, the trained model predicts the distribution $p(y^\star | x^\star, \hat{\theta}_{\mathrm{MLE}})$ over the target class variable~$y^\star$, given a test input $x^\star$. These DNN models are thus able to capture input-dependent aleatoric uncertainty, by outputting less confident predictions for inherently ambiguous cases.

In regression, the most common approach is to let the DNN directly predict targets, $y^\star = f_{\hat{\theta}}(x^\star)$. The parameters $\hat{\theta}$ are learned by minimizing e.g. the $L^2$ or $L^1$ loss over the training dataset \cite{shi2018pointrcnn, lang2018pointpillars}. However, such direct regression does not model aleatoric uncertainty. Instead, recent work \cite{ilg2018uncertainty, kendall2017uncertainties, lakshminarayanan2017simple} has explored predicting the distribution $p(y | x, \theta)$, similar to the classification case above. For instance, $p(y | x, \theta)$ can be parameterized by a Gaussian distribution~\cite{chua2018deep, lakshminarayanan2017simple}, giving the following model in the 1D case,
\begin{equation}
\begin{gathered}
    p(y | x, \theta) = \mathcal{N}\big(y; \mu_{\theta}(x),\,\sigma^2_{\theta}(x)\big),\\
    f_{\theta}(x) = \rvect{\mu_{\theta}(x), \log\sigma^2_{\theta}(x)}^{\Transp} \in \mathbb{R}^2.
\label{eq:gaussian_model}
\end{gathered}
\end{equation}
Here, the DNN predicts the mean $\mu_{\theta}(x)$ and variance $\sigma^2_{\theta}(x)$ of the target $y$. The variance is naturally interpreted as a measure of input-dependent aleatoric uncertainty. As in classification, the model parameters $\theta$ are learned by minimizing the negative log-likelihood $- \sum_i \log p(y_i | x_i, \theta)$.

\parsection{Epistemic Uncertainty}
While the above models can capture aleatoric uncertainty, stemming from the data, they are agnostic to the uncertainty in the model parameters $\theta$.\ A principled means to estimate this epistemic uncertainty is to perform Bayesian inference. The aim is to utilize the posterior distribution $p(\theta | \mathcal{D})$, which is obtained from the data likelihood and a chosen prior $p(\theta)$ by applying Bayes' theorem. The uncertainty in the parameters $\theta$ is then marginalized out to obtain the predictive posterior distribution,
\begin{equation}
\begin{gathered}
    p(y^\star | x^\star, \mathcal{D}) = \int p(y^\star | x^\star, \theta)p(\theta | \mathcal{D}) d\theta\\ 
    \approx \frac{1}{M} \sum_{i=1}^{M}  p(y^\star | x^\star, \theta^{(i)}), \quad \theta^{(i)} \sim p(\theta | \mathcal{D})\,.
\label{eq:predictive_distribution_Bayesian}
\end{gathered}
\end{equation}
Here, the generally intractable integral in (\ref{eq:predictive_distribution_Bayesian}) is approximated using $M$ Monte Carlo samples $\theta^{(i)}$, ideally drawn from the posterior. In practice however, obtaining samples from the true posterior $p(\theta | \mathcal{D})$ is virtually impossible, requiring an approximate posterior $q(\theta) \approx p(\theta | \mathcal{D})$ to be used. We thus obtain the approximate predictive posterior as,
\begin{equation}
    \hat{p}(y^\star | x^\star, \mathcal{D}) \triangleq \frac{1}{M} \sum_{i=1}^{M}  p(y^\star | x^\star, \theta^{(i)}), \quad \theta^{(i)} \sim q(\theta) \,,
\label{eq:approx_predictive_distribution_Bayesian}
\end{equation}
which enables us to estimate both aleatoric and epistemic uncertainty of the prediction. The quality of the approximation (\ref{eq:approx_predictive_distribution_Bayesian}) depends on the number of samples $M$ and the method employed for generating $q(\theta)$. Prior work on such approximate Bayesian inference methods is discussed in Section~\ref{related_work}. For the Categorical model~(\ref{eq:categorical_model}), $\hat{p}(y^\star | x^\star, \mathcal{D}) = \Cat(y^\star; \hat{s}(x^\star))$, $\hat{s}(x^\star) = \frac{1}{M} \sum_{i=1}^{M} s_{\theta^{(i)}}(x^\star)$.\ For the Gaussian model~(\ref{eq:gaussian_model}), $\hat{p}(y^\star | x^\star, \mathcal{D})$ is a uniformly weighted mixture of Gaussian distributions. We approximate this mixture with a single Gaussian, see Appendix~\ref{appendix:gaussian_approximation_of_the_predictive_distribution} for details.

\parsection{Illustrative Example}
To visualize and provide intuition for the problem of predictive uncertainty estimation with DNNs, we consider the problem of regressing a sinusoid corrupted by \emph{input-dependent} Gaussian noise,
\begin{equation}
\begin{gathered}
y \sim \mathcal{N}\big(\mu(x), \sigma^2(x)\big),\\
\mu(x) = \sin(x), \quad \sigma(x) = 0.15(1 + e^{-x})^{-1}.
\end{gathered}
\label{eq:toy_regression_problem}
\end{equation}
Training data $\{(x_i, y_i)\}_{i=1}^{1000}$ is only given in the interval $[-3, 3]$, see Figure~\ref{fig:illustrative_1d_regression_training_dataset}. A DNN trained to directly predict the target $y$ is able to accurately regress the mean for $x^\star~\in~[-3, 3]$, see Figure~\ref{fig:illustrative_1d_regression_deterministic}. However, this model does not capture any notion of uncertainty. A corresponding Gaussian DNN model~(\ref{eq:gaussian_model}) trained via maximum-likelihood obtains a predictive distribution that closely matches the ground truth for $x^\star \in [-3, 3]$, see Figure~\ref{fig:illustrative_1d_regression_mle}. While correctly accounting for aleatoric uncertainty, this model generates overly confident predictions for inputs $|x^\star| > 3$ not seen during training. Finally, the Gaussian DNN model trained via approximate Bayesian inference~(\ref{eq:approx_predictive_distribution_Bayesian}), with a prior distribution $p(\theta)~=~\mathcal{N}(0, I_{P})$ and $M=1\thinspace000$ samples obtained via Hamiltonian Monte Carlo~\cite{neal2011mcmc}, is additionally able to predict more reasonable uncertainties in the region with no available training data, see Figure~\ref{fig:illustrative_1d_regression_bayesian}.
\section{Related Work}
\label{related_work}

Here, we discuss prior work on approximate Bayesian inference. We also note that ensembling, which is often considered a non-Bayesian alternative, in fact can naturally be viewed as an approximate Bayesian inference method.

\begin{figure*}
     \centering
     \begin{subfigure}[b]{0.221\textwidth}
         \centering
         \includegraphics[width=\textwidth]{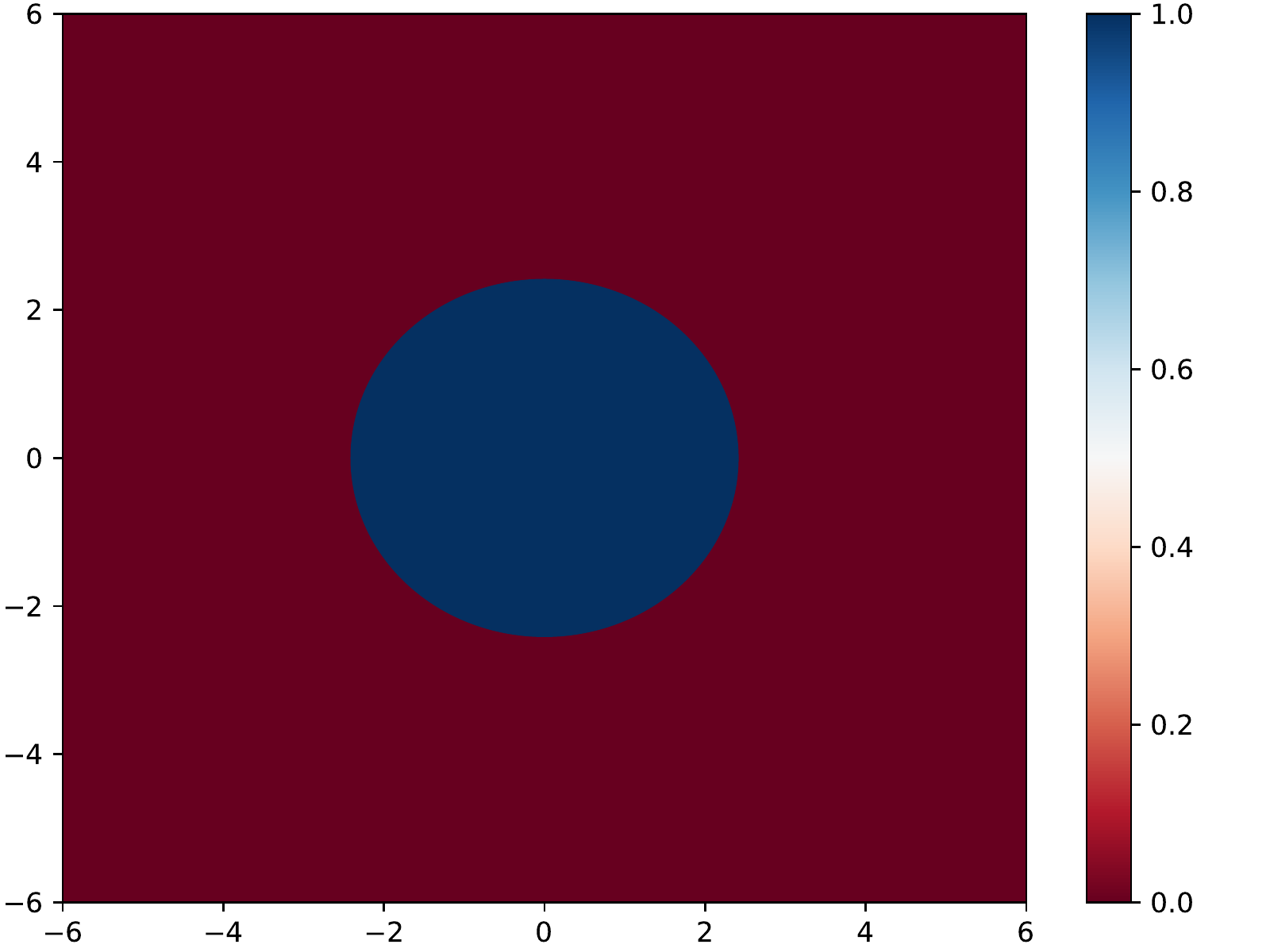}\vspace{-1mm}%
         \caption{}
         \label{fig:illustrative_toy_classificaton_a}
     \end{subfigure}
     \vspace{-0.75mm}
     \begin{subfigure}[b]{0.221\textwidth}
         \centering
         \includegraphics[width=0.82\textwidth]{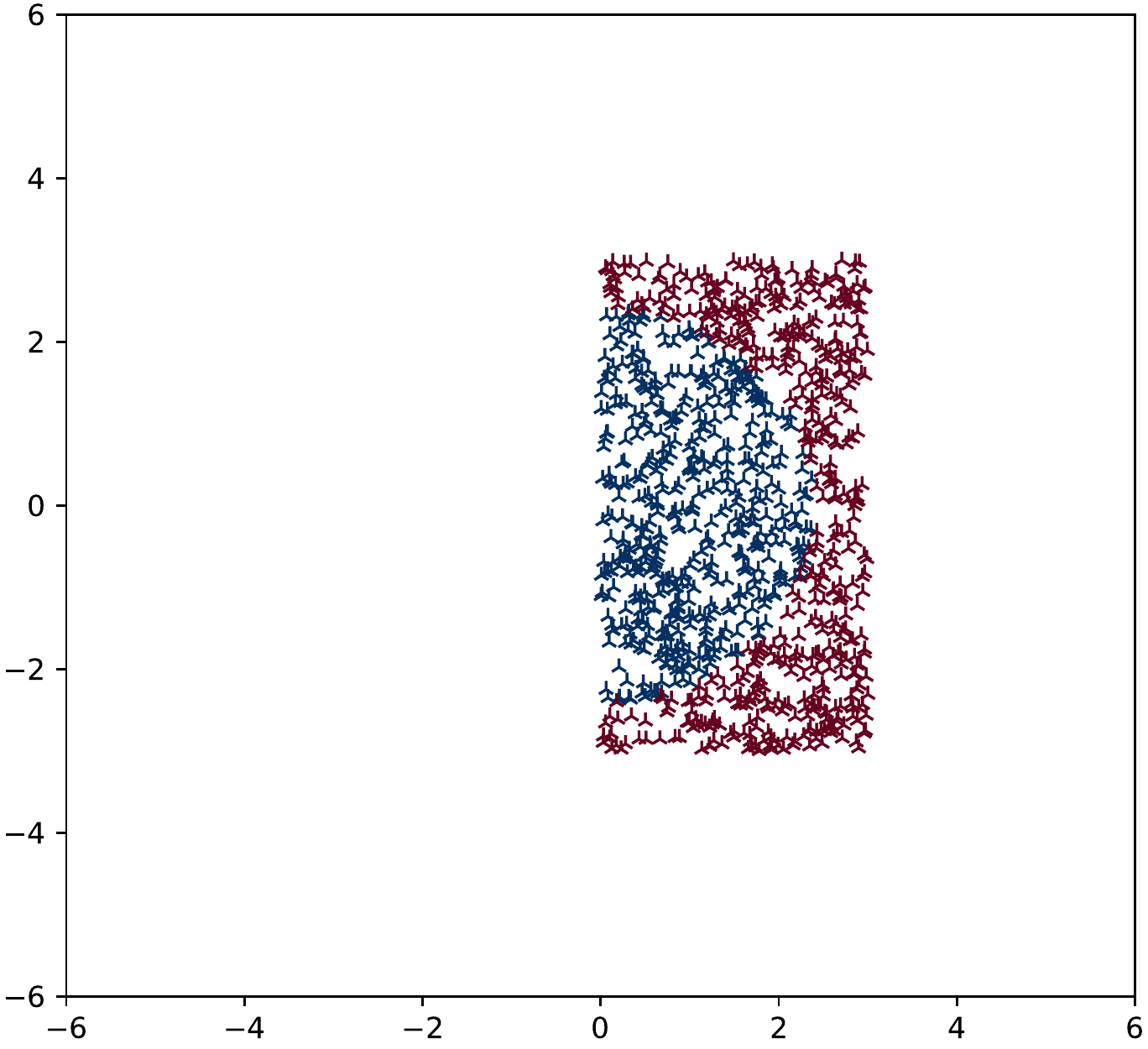}\vspace{-1mm}%
         \caption{}
         \label{fig:illustrative_toy_classificaton_b}
     \end{subfigure}
     \vspace{-0.75mm}
     \begin{subfigure}[b]{0.221\textwidth}
         \centering
         \includegraphics[width=\textwidth]{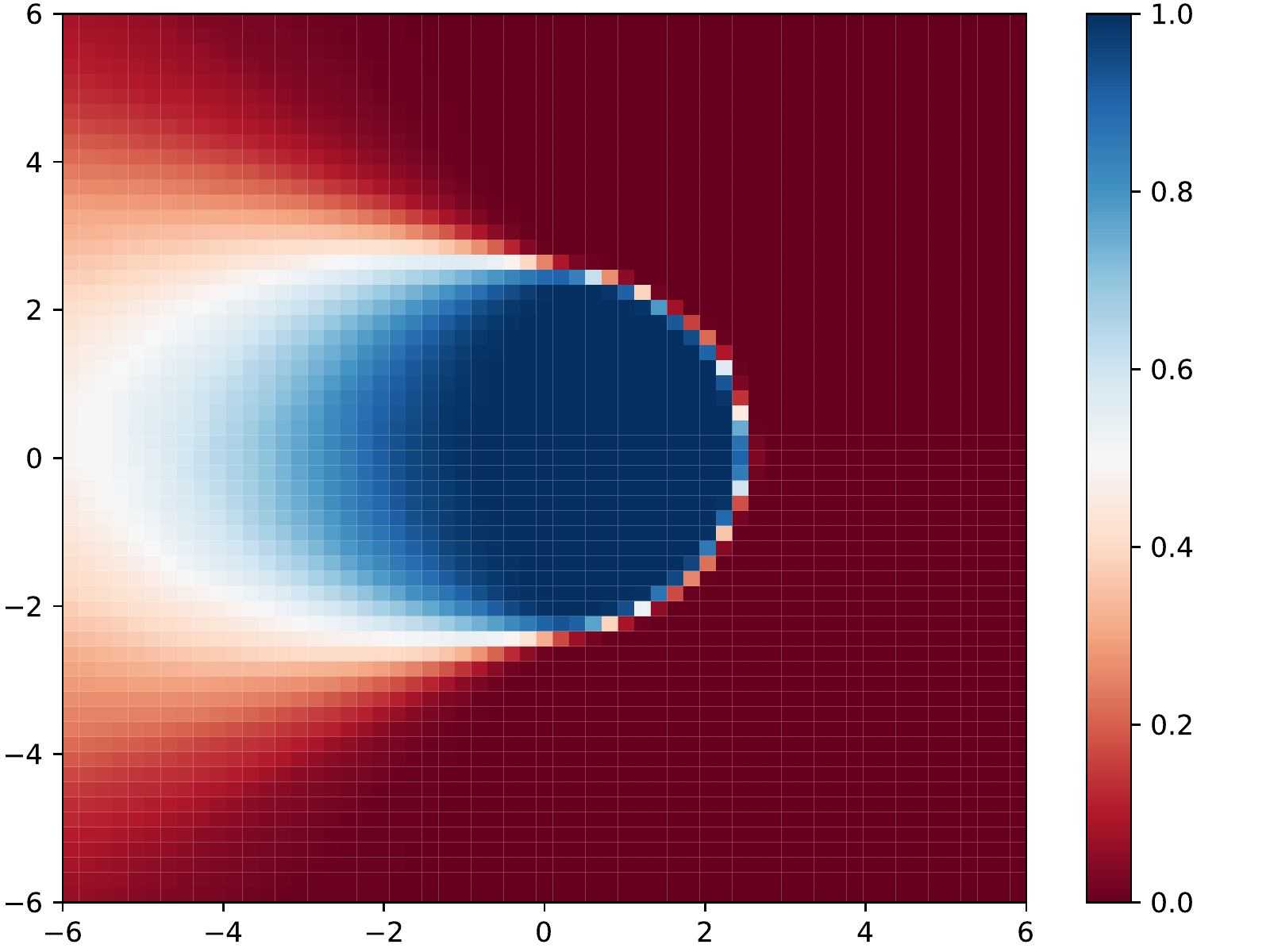}\vspace{-1mm}%
         \caption{}
         \label{fig:illustrative_toy_classificaton_c}
     \end{subfigure}
     \vspace{-0.75mm}
        \caption{Toy binary classification problem. \textbf{(a)} True data generator, red and blue represents the two classes. \textbf{(b)} Training dataset with $N\!=\!1\thinspace040$ examples. \textbf{(c)} ``Ground truth'' predictive distribution, obtained using HMC~\cite{neal2011mcmc}.}\vspace{-3.5mm}
        \label{fig:illustrative_toy_classificaton}
\end{figure*}

\parsection{Approximate Bayesian Inference}
The method employed for approximating the posterior $q(\theta) \approx p(\theta | \mathcal{D}) = p(Y | X, \theta)p(\theta)/p(Y | X)$ is a crucial choice, determining the quality of the approximate predictive posterior $\hat{p}(y^\star | x^\star, \mathcal{D})$ in~(\ref{eq:approx_predictive_distribution_Bayesian}). There exists two main paradigms for constructing  $q(\theta)$, the first one being \textit{Markov chain Monte Carlo (MCMC)} methods. Here, samples $\theta^{(i)}$ approximately distributed according to the posterior are obtained by simulating a Markov chain with $p(\theta | \mathcal{D})$ as its stationary distribution. For DNNs, this approach was pioneered by Neal~\cite{neal1995bayesian}, who employed Hamiltonian Monte Carlo (HMC) on small feed-forward neural networks. HMC entails performing Metropolis-Hastings~\cite{metropolis1953equation, hastings1970monte} updates using Hamiltonian dynamics based on the potential energy $U(\theta) \triangleq -\log p(Y | X, \theta)p(\theta)$. To date, it is considered a ``gold standard'' method for approximate Bayesian inference, but does not scale to large DNNs or large-scale datasets.\ Therefore, \textit{Stochastic Gradient MCMC (SG-MCMC)}~\cite{ma2015complete} methods have been explored, in which stochastic gradients are utilized in place of their full-data counterparts. SG-MCMC variants include Stochastic Gradient Langevin Dynamics (SGLD)~\cite{welling2011bayesian}, where samples $\theta^{(i)}$ are collected from the parameter trajectory given by the update equation $\theta_{t+1} = \theta_{t} - \alpha_{t}\nabla_{\theta}\widetilde{U}(\theta_{t}) + \sqrt{2\alpha_{t}}\epsilon_{t}$, where $\epsilon_{t} \sim \mathcal{N}(0, 1)$ and $\nabla_{\theta}\widetilde{U}(\theta)$ is the stochastic gradient of $U(\theta)$. Save for the noise term $\sqrt{2\alpha_{t}}\epsilon_{t}$, this update is identical to the conventional SGD update when minimizing the maximum-a-posteriori (MAP) objective $-\log p(Y | X, \theta)p(\theta)$. Similarly, Stochastic Gradient HMC (SGHMC)~\cite{chen2014stochastic} corresponds to SGD with momentum injected with properly scaled noise. Given a limited computational budget, SG-MCMC methods can however struggle to explore the high-dimensional and highly multi-modal posteriors of large DNNs. To mitigate this problem, Zhang et al.~\cite{zhang2019cyclical} proposed to use a cyclical stepsize schedule to help escaping local modes in $p(\theta | \mathcal{D})$.

The second paradigm is that of \textit{Variational Inference (VI)}~\cite{hinton1993keeping, barber1998ensemble, graves2011practical, blundell2015weight}. Here, a distribution $q_{\phi}(\theta)$ parameterized by variational parameters $\phi$ is explicitly chosen, and the best possible approximation is found by minimizing the Kullback-Leibler (KL) divergence with respect to the true posterior $p(\theta | \mathcal{D})$. While principled, VI methods generally require sophisticated implementations, especially for more expressive variational distributions $q_{\phi}(\theta)$~\cite{louizos2016structured, louizos2017multiplicative, zhang2018noisy}. A particularly simple and scalable method is MC-dropout~\cite{gal2016dropout}. It entails using dropout~\cite{srivastava2014dropout} also at test time, which can be interpreted as performing VI with a Bernoulli variational distribution~\cite{gal2016dropout, kendall2015bayesian, mukhoti2018evaluating}. The approximate predictive posterior $\hat{p}(y^\star | x^\star, \mathcal{D})$ in~(\ref{eq:approx_predictive_distribution_Bayesian}) is obtained by performing $M$ stochastic forward passes on the same input.

\parsection{Ensembling}
Lakshminarayanan et al.~\cite{lakshminarayanan2017simple} created a parametric model $p(y | x, \theta)$ of the conditional distribution using a DNN $f_{\theta}$, and learned multiple point estimates $\{\hat{\theta}^{(m)}\}_{m=1}^{M}$ by repeatedly minimizing the MLE objective $-\log p(Y | X, \theta)$ with random initialization. They then averaged over the corresponding parametric models to obtain the following predictive distribution,
\begin{equation}
    \hat{p}(y^\star | x^\star) \triangleq \frac{1}{M} \sum_{m=1}^{M}  p(y^\star | x^\star, \hat{\theta}^{(m)}).
\label{eq:predictive_distribution_ensembling}
\end{equation}
The authors considered this a non-Bayesian alternative to predictive uncertainty estimation. However, since the point estimates $\{\hat{\theta}^{(m)}\}_{m=1}^{M}$ always can be seen as samples from some distribution $\hat{q}(\theta)$, we note that (\ref{eq:predictive_distribution_ensembling}) is virtually identical to the approximate predictive posterior in~(\ref{eq:approx_predictive_distribution_Bayesian}). Ensembling can thus naturally be viewed as approximate Bayesian inference, where the level of approximation is determined by how well the implicit sampling distribution $\hat{q}(\theta)$ approximates the posterior $p(\theta | \mathcal{D})$. Ideally, $\{\hat{\theta}^{(m)}\}_{m=1}^{M}$ should be distributed exactly according to $p(\theta | \mathcal{D})~\propto~p(Y | X, \theta)p(\theta)$. Since $p(Y | X, \theta)$ is highly \emph{multi-modal} in the parameter space for DNNs~\cite{auer1996exponentially, choromanska2015loss}, so is $p(\theta | \mathcal{D})$. By minimizing $-\log p(Y | X, \theta)$ multiple times, starting from \emph{randomly chosen} initial points, we are likely to find different local optima. Ensembling can thus generate a compact set of samples $\{\hat{\theta}^{(m)}\}_{m=1}^{M}$ that, even for small values of $M$, captures this important aspect of multi-modality in $p(\theta | \mathcal{D})$.
\begin{figure*}
\centering
     \begin{subfigure}[c]{0.215\textwidth}
         \centering
         \includegraphics[width=\textwidth]{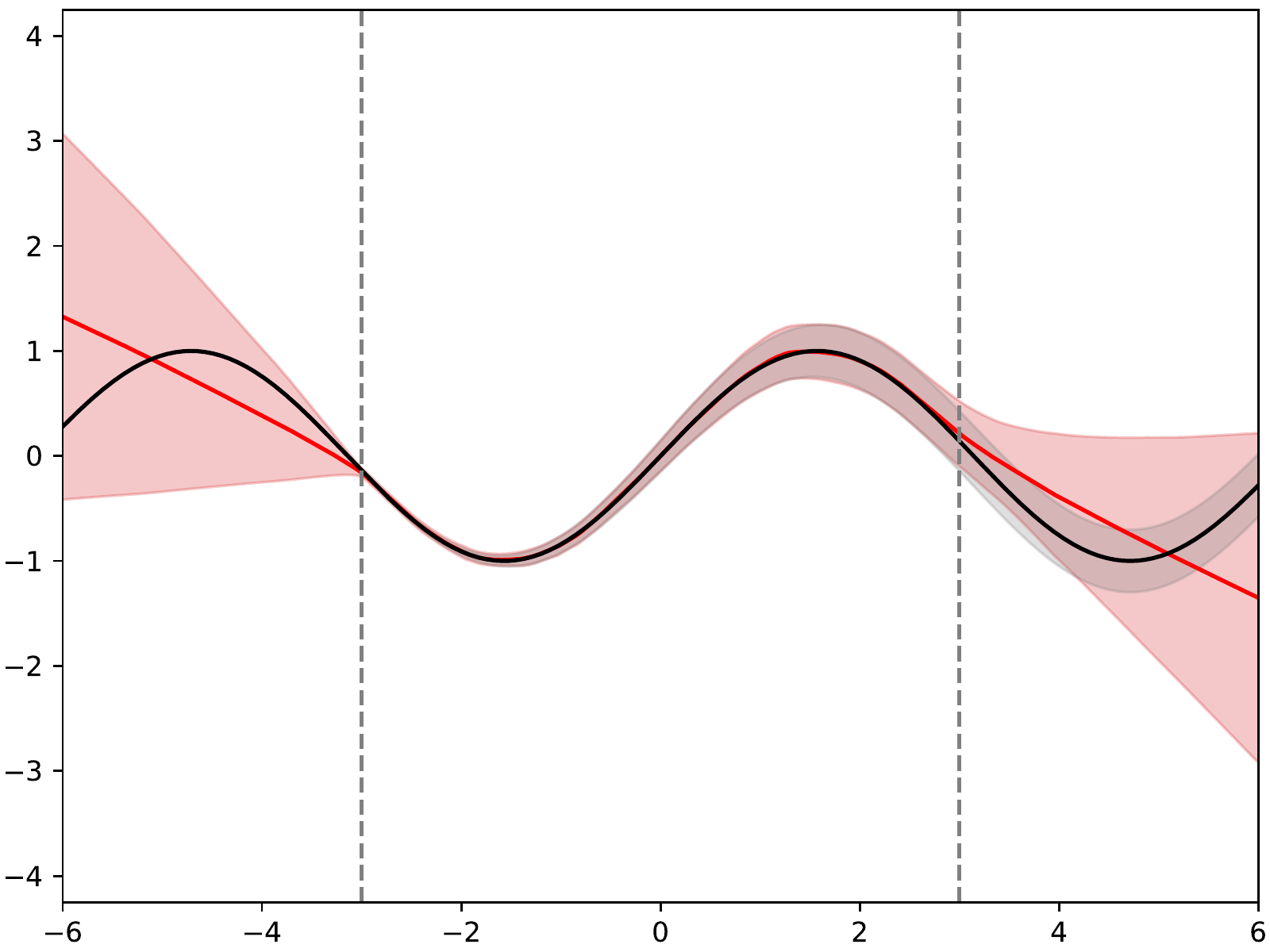}
         \caption{Ensembling.}
     \end{subfigure}
     \vspace{-0.75mm}
     \begin{subfigure}[c]{0.215\textwidth}
         \centering
         \includegraphics[width=\textwidth]{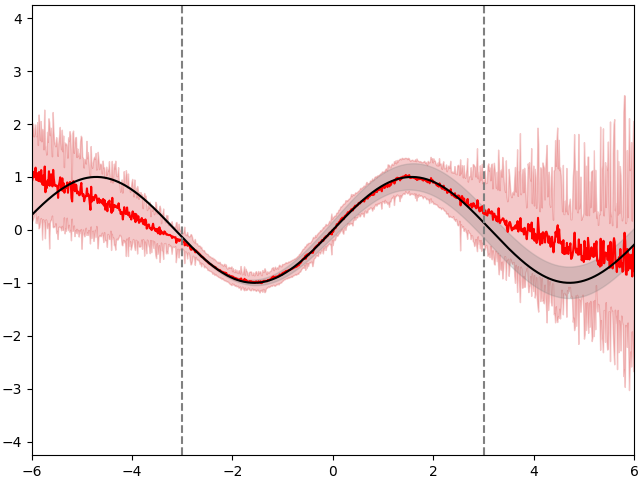}
         \caption{MC-dropout.}
     \end{subfigure}
     \vspace{-0.75mm}
     \begin{subfigure}[c]{0.215\textwidth}
         \centering
         \includegraphics[width=\textwidth]{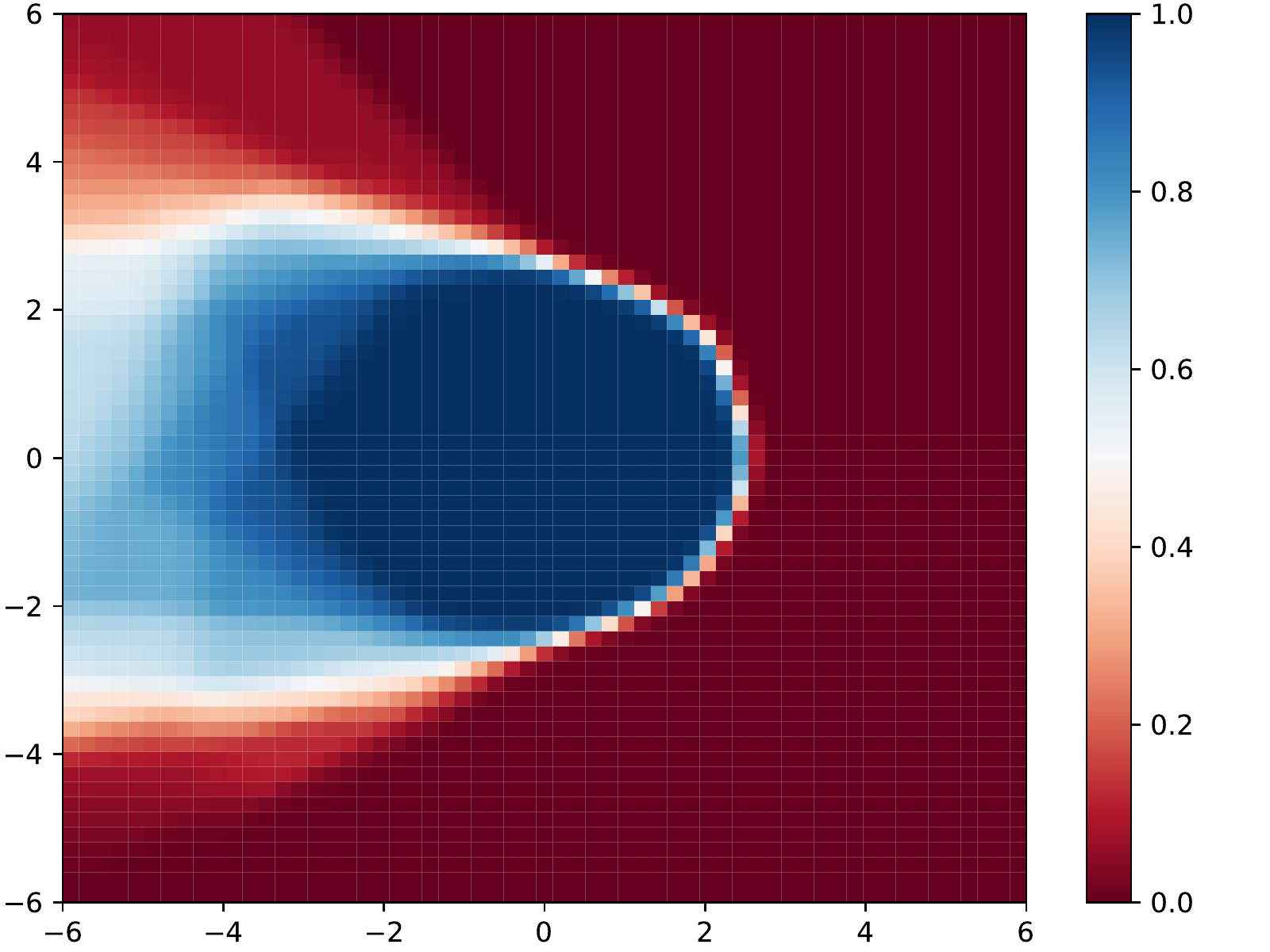}
         \caption{Ensembling.}
     \end{subfigure}
     \vspace{-0.75mm}
     \begin{subfigure}[c]{0.215\textwidth}
         \centering
         \includegraphics[width=\textwidth]{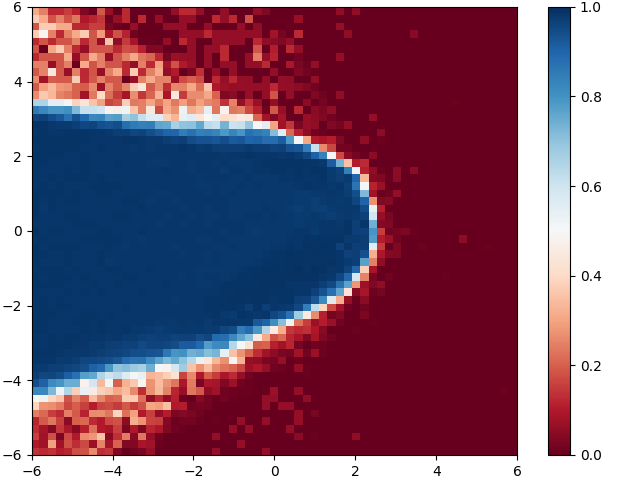}
         \caption{MC-dropout.}
     \end{subfigure}
    \vspace{-0.75mm}
  \caption{Illustrative toy problems - example predictive distributions for ensembling and MC-dropout with $M\!=\!16$ samples.}\vspace{-3.5mm}
  \label{fig:toy_problems_qualitative_results}
\end{figure*}

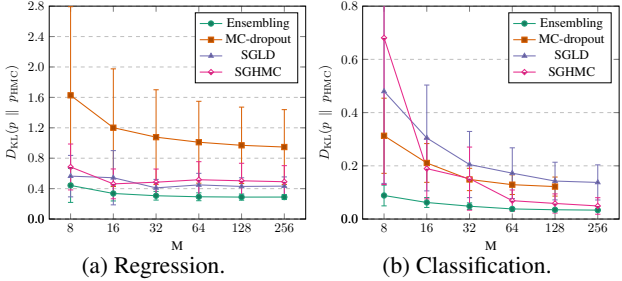
\begin{figure}
\centering
    \begin{subfigure}[c]{0.49\columnwidth}
      \centering
            \begin{tikzpicture}[scale=0.4975, baseline]
                \begin{axis}[
                    xmode=log,
                    log ticks with fixed point,
                    xlabel={M},
                    ylabel={$D_\mathrm{KL}(p~\parallel~p_{\mathrm{HMC}})$},
                    ymin=0.0, ymax=2.8,
                    xtick={8, 16, 32, 64, 128, 256},
                    ytick={0,0, 0.4, 0.8, 1.2, 1.6, 2.0, 2.4, 2.8},
                    legend pos=north east,
                    ymajorgrids=true,
                    grid style=dashed,
                    y tick label style={
                        /pgf/number format/.cd,
                            fixed,
                            fixed zerofill,
                            precision=1,
                        /tikz/.cd
                    },
                    every axis plot/.append style={thick},
                ]
            \addplot
             plot [error bars/.cd, y dir = both, y explicit]
             table[row sep=crcr, x index=0, y index=1, y error index=2,]{
            8 0.442268 0.2233641\\
            16 0.336318 0.0984614\\
            32 0.306449 0.0571394\\
            64 0.292924 0.0518983\\
            128 0.288727 0.0408139\\
            256 0.28913 0.0322919\\
            };
            \addlegendentry{Ensembling} 
            
            \addplot
             plot [error bars/.cd, y dir = both, y explicit]
             table[row sep=crcr, x index=0, y index=1, y error index=2,]{
            8 1.62725 1.16544\\
            16 1.20143 0.77387\\
            32 1.07685 0.623371\\
            64 1.01024 0.538445\\
            128 0.969594 0.502551\\
            256 0.94703 0.491882\\
            };
            \addlegendentry{MC-dropout} 
            
            \addplot
             plot [error bars/.cd, y dir = both, y explicit]
             table[row sep=crcr, x index=0, y index=1, y error index=2,]{
            8 0.564368 0.272595\\
            16 0.543245 0.356657\\
            32 0.410131 0.111204\\
            64 0.449033 0.152189\\
            128 0.428836 0.113887\\
            256 0.433093 0.121152\\
            };
            \addlegendentry{SGLD} 
            
            \addplot
             plot [error bars/.cd, y dir = both, y explicit]
             table[row sep=crcr, x index=0, y index=1, y error index=2,]{
            8 0.684939 0.301885\\
            16 0.46253 0.196306\\
            32 0.484642 0.17294\\
            64 0.516942 0.237645\\
            128 0.503064 0.230595\\
            256 0.490817 0.211457\\
            };
            \addlegendentry{SGHMC} 
            
            \end{axis}
            \end{tikzpicture}\vspace{-2.35mm}%
      \caption{Regression.}
      \label{fig:toy_problems_quantative_results_regression}
    \end{subfigure}
    \vspace{-0.95mm}
    \begin{subfigure}[c]{0.49\columnwidth}
      \centering
            \begin{tikzpicture}[scale=0.4975, baseline]
                \begin{axis}[
                    xmode=log,
                    log ticks with fixed point,
                    xlabel={M},
                    ylabel={$D_\mathrm{KL}(p~\parallel~p_{\mathrm{HMC}})$},
                    ymin=0.0, ymax=0.8,
                    xtick={8, 16, 32, 64, 128, 256},
                    ytick={0,0, 0.2, 0.4, 0.6, 0.8},
                    legend pos=north east,
                    ymajorgrids=true,
                    grid style=dashed,
                    y tick label style={
                        /pgf/number format/.cd,
                            fixed,
                            fixed zerofill,
                            precision=1,
                        /tikz/.cd
                    },
                    every axis plot/.append style={thick},
                ]
            \addplot 
             plot [error bars/.cd, y dir = both, y explicit]
             table[row sep=crcr, x index=0, y index=1, y error index=2,]{
            8 0.0885124 0.0389856\\
            16 0.0622361 0.0186496\\
            32 0.0482295 0.0124783\\
            64 0.0380383 0.00531272\\
            128 0.0348956 0.00400926\\
            256 0.0335332 0.00138975\\
            };
            \addlegendentry{Ensembling} 
            
            \addplot
             plot [error bars/.cd, y dir = both, y explicit]
             table[row sep=crcr, x index=0, y index=1, y error index=2,]{
            8 0.312933 0.141196\\
            16 0.210738 0.072815\\
            32 0.148762 0.0419286\\
            64 0.129146 0.0369342\\
            128 0.122003 0.0358212\\
            };
            \addlegendentry{MC-dropout} 
            
            \addplot
             plot [error bars/.cd, y dir = both, y explicit]
             table[row sep=crcr, x index=0, y index=1, y error index=2,]{
            8 0.479821 0.345935\\
            16 0.304685 0.198831\\
            32 0.204824 0.124369\\
            64 0.172374 0.0952615\\
            128 0.142729 0.0707726\\
            256 0.137614 0.0660084\\
            };
            \addlegendentry{SGLD} 
            
            \addplot
             plot [error bars/.cd, y dir = both, y explicit]
             table[row sep=crcr, x index=0, y index=1, y error index=2,]{
            8 0.680847 0.550977\\
            16 0.189506 0.113788\\
            32 0.151961 0.118518\\
            64 0.0690482 0.0401916\\
            128 0.0586131 0.0359652\\
            256 0.0491939 0.0310982\\
            };
            \addlegendentry{SGHMC} 
            
            \end{axis}
            \end{tikzpicture}\vspace{-2.35mm}%
      \caption{Classification.}
      \label{fig:toy_problems_quantative_results_classification}
    \end{subfigure}
    \vspace{-0.95mm}
  \caption{Illustrative toy problems - quantitative results. The plots show the KL divergence ($\downarrow$) between the predictive distribution estimated by each method and the HMC ``ground truth'', for different number of samples $M$.}\vspace{-3.5mm}
  \label{fig:toy_problems_quantative_results}
\end{figure}

\section{Experiments}
\label{experiments}

We conduct experiments both on illustrative toy regression and classification problems (Section~\ref{experiments-illustrative_toy_problems}), and on the real-world computer vision tasks of depth completion (Section~\ref{experiments-depth_completion}) and street-scene semantic segmentation (Section~\ref{experiments-semantic_segmentation}). Our evaluation is motivated by real-world conditions found e.g. in automotive applications, where robustness to varying environments and weather conditions is required to ensure safety. Since images captured in these different circumstances could all represent distinctly different regions of the vast input image space, it is infeasible to ensure that all encountered inputs will be well-represented by the training data. Thus, we argue that robustness to out-of-domain inputs is crucial in such applications. To simulate these challenging conditions and test the robustness required for such real-world scenarios, we train all models on synthetic data and evaluate them on real-world data. To improve rigour of our evaluation, we repeat each experiment multiple times and report results together with the observed variation. A more detailed description of all results are found in the Appendix (Appendix~\ref{appendix:illustrative_toy_problems_results_description}, \ref{appendix:depth_completion_results_description}, \ref{appendix:street-scene_semantic_segmentation_results_description}). All experiments are implemented in PyTorch~\cite{paszke2017automatic}.

\subsection{Illustrative Toy Problems}
\label{experiments-illustrative_toy_problems}

We first present results on illustrative toy problems to gain insights into how ensembling and MC-dropout fare against other approximate Bayesian inference methods. For regression, we conduct experiments on the 1D problem defined in~(\ref{eq:toy_regression_problem}) and visualized in Figure~\ref{fig:illustrative_1d_regression}. We use the Gaussian model~(\ref{eq:gaussian_model}) with two separate feed-forward neural networks outputting $\mu_{\theta}(x)$ and $\log\sigma^2_{\theta}(x)$. We evaluate the methods by quantitatively measuring how well the obtained predictive distributions approximate that of the ``gold standard'' HMC~\cite{neal2011mcmc} with $M\!=\!1\thinspace000$ samples and prior $p(\theta)=\mathcal{N}(0, I_{P})$. We thus consider the predictive distribution visualized in Figure~\ref{fig:illustrative_1d_regression_bayesian} ground truth, and take as our metric the KL divergence $D_\mathrm{KL}(p~\parallel~p_{\mathrm{HMC}})$ with respect to this target distribution $p_\mathrm{HMC}$.\ For classification, we conduct experiments on the binary classification problem in Figure~\ref{fig:illustrative_toy_classificaton}. The true data generator is visualized in Figure~\ref{fig:illustrative_toy_classificaton_a}, where red and blue represents the two classes. The training dataset contains $520$ examples of each class, and is visualized in Figure~\ref{fig:illustrative_toy_classificaton_b}. We use the Categorical model~(\ref{eq:categorical_model}) with a feed-forward neural network. As for regression, we quantitatively measure how well the obtained predictive distributions approximate that of HMC, which is visualized in Figure~\ref{fig:illustrative_toy_classificaton_c}. Further details are provided in Appendix~\ref{appendix:illustrative_toy_problems}.

\parsection{Results}
A comparison of ensembling, MC-dropout, SGLD and SGHMC in terms of $D_\mathrm{KL}(p~\parallel~p_{\mathrm{HMC}})$ is found in Figure~\ref{fig:toy_problems_quantative_results}. The Adam optimizer~\cite{kingma2014adam} is here used for both ensembling and MC-dropout. We observe that ensembling consistently outperforms the compared methods, and MC-dropout in particular. Even compared to SG-MCMC variants such as SGLD and SGHMC, ensembling thus provides a better approximation to the MCMC method HMC. This result is qualitatively supported by visualized predictive distributions found in Appendix~\ref{appendix:illustrative_toy_problem_qualitative_results}. Example predictive distributions for ensembling and MC-dropout with $M\!=\!16$ are shown in Figure~\ref{fig:toy_problems_qualitative_results}. We observe that ensembling provides reasonable approximations to HMC even for quite small values of $M$, especially compared to MC-dropout.

\subsection{Depth Completion}
\label{experiments-depth_completion}

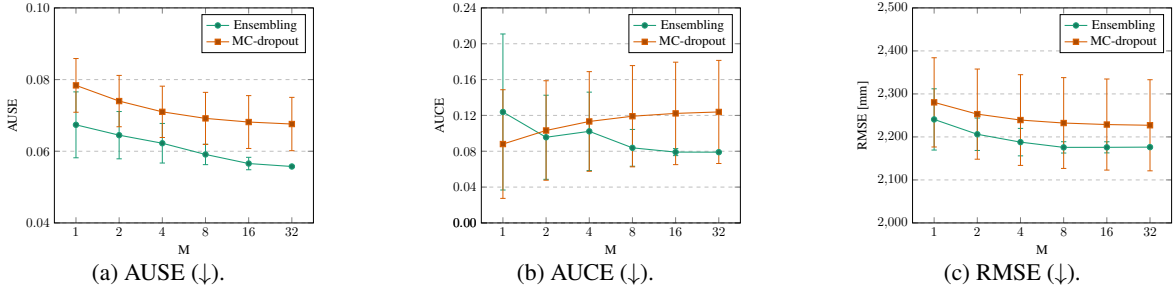
\begin{figure*}
\centering
    \begin{subfigure}[t]{0.32\textwidth}
      \centering
            \begin{tikzpicture}[scale=0.5025, baseline]
                \begin{axis}[
                    xmode=log,
                    log ticks with fixed point,
                    xlabel={M},
                    ylabel={AUSE},
                    ymin=0.04, ymax=0.1,
                    xtick={1, 2, 4, 8, 16, 32},
                    ytick={0.04, 0.06 ,0.08, 0.1},
                    legend pos=north east,
                    ymajorgrids=true,
                    grid style=dashed,
                    y tick label style={
                        /pgf/number format/.cd,
                            fixed,
                            fixed zerofill,
                            precision=2,
                        /tikz/.cd
                    },
                    every axis plot/.append style={thick},
                ]
            \addplot
             plot [error bars/.cd, y dir = both, y explicit]
             table[row sep=crcr, x index=0, y index=1, y error index=2,]{
            1 0.0673736 0.00919524\\
            2 0.0644875 0.00659769\\
            4 0.0622355 0.00552193\\
            8 0.0591118 0.00284625\\
            16 0.056574 0.00173449\\
            32 0.0557493 0.000257961\\
            };
            \addlegendentry{Ensembling}
            \addplot
             plot [error bars/.cd, y dir = both, y explicit]
             table[row sep=crcr, x index=0, y index=1, y error index=2,]{
            1 0.0783795 0.00749478\\
            2 0.0740023 0.00716145\\
            4 0.0710013 0.00715852\\
            8 0.0691715 0.00726264\\
            16 0.0681523 0.00737443\\
            32 0.0675977 0.0074391\\
            };
            \addlegendentry{MC-dropout}
            \end{axis}
            \end{tikzpicture}\vspace{-2mm}%
      \caption{AUSE ($\downarrow$).}
      \label{fig:depth_completion_ause}
    \end{subfigure}
    \vspace{-0.75mm}
    \begin{subfigure}[t]{0.32\textwidth}
      \centering
            \begin{tikzpicture}[scale=0.5025, baseline]
                \begin{axis}[
                    xmode=log,
                    log ticks with fixed point,
                    xlabel={M},
                    ylabel={AUCE},
                    ymin=0.0, ymax=0.24,
                    xtick={1, 2, 4, 8, 16, 32},
                    ytick={0,0, 0.04, 0.08 ,0.12, 0.16, 0.20, 0.24},
                    legend pos=north east,
                    ymajorgrids=true,
                    grid style=dashed,
                    y tick label style={
                        /pgf/number format/.cd,
                            fixed,
                            fixed zerofill,
                            precision=2,
                        /tikz/.cd
                    },
                    every axis plot/.append style={thick},
                ]
            \addplot
             plot [error bars/.cd, y dir = both, y explicit]
             table[row sep=crcr, x index=0, y index=1, y error index=2,]{
            1 0.123815 0.0870771\\
            2 0.0956072 0.0469372\\
            4 0.10223 0.0437496\\
            8 0.083811 0.0205284\\
            16 0.0790672 0.0039251\\
            32 0.0789799 0.000497125\\
            };
            \addlegendentry{Ensembling}
            \addplot
             plot [error bars/.cd, y dir = both, y explicit]
             table[row sep=crcr, x index=0, y index=1, y error index=2,]{
            1 0.0880315 0.0606418\\
            2 0.103185 0.0555137\\
            4 0.113254 0.0556156\\
            8 0.1191 0.0565115\\
            16 0.122245 0.0571666\\
            32 0.123869 0.0575381\\
            };
            \addlegendentry{MC-dropout}
            \end{axis}
            \end{tikzpicture}\vspace{-2mm}%
      \caption{AUCE ($\downarrow$).}
      \label{fig:depth_completion_auce}
    \end{subfigure}
    \vspace{-0.75mm}
    \begin{subfigure}[t]{0.32\textwidth}
      \centering
            \begin{tikzpicture}[scale=0.5025, baseline]
                \begin{axis}[
                    xmode=log,
                    log ticks with fixed point,
                    xlabel={M},
                    ylabel={RMSE [mm]},
                    ymin=2000, ymax=2500,
                    xtick={1, 2, 4, 8, 16, 32},
                    ytick={2000, 2100, 2200, 2300, 2400, 2500},
                    legend pos=north east,
                    ymajorgrids=true,
                    grid style=dashed,
                    every axis plot/.append style={thick},
                ]
            \addplot
             plot [error bars/.cd, y dir = both, y explicit]
             table[row sep=crcr, x index=0, y index=1, y error index=2,]{
            1 2240.68 71.2081\\
            2 2206.18 37.6691\\
            4 2187.89 31.9101\\
            8 2175.69 13.1603\\
            16 2175.8 12.9214\\
            32 2176.29 2.04356\\
            };
            \addlegendentry{Ensembling}
            \addplot
             plot [error bars/.cd, y dir = both, y explicit]
             table[row sep=crcr, x index=0, y index=1, y error index=2,]{
            1 2280.34 103.747\\
            2 2253.01 104.82\\
            4 2239.22 105.467\\
            8 2232.29 105.577\\
            16 2228.77 105.89\\
            32 2227.08 105.912\\
            };
            \addlegendentry{MC-dropout}
            \end{axis}
            \end{tikzpicture}\vspace{-2mm}%
      \caption{RMSE ($\downarrow$).}
      \label{fig:depth_completion_rmse}
    \end{subfigure}
    \vspace{-0.75mm}
  \caption{Depth completion - quantitative results. The plots show a comparison of ensembling and MC-dropout in terms of AUSE, AUCE and RMSE on the KITTI depth completion validation dataset, for different number of samples $M$.}\vspace{-3.5mm}
  \label{fig:depth_completion_ause_auce}
\end{figure*}

\begin{figure}
     \centering
     \begin{subfigure}[b]{0.48\columnwidth}
         \centering
         \includegraphics[width=\textwidth]{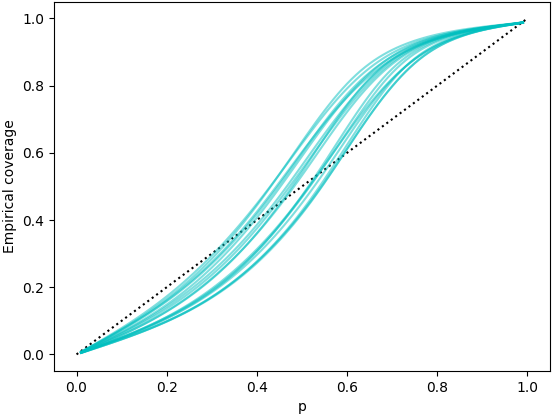}
         \caption{Ensembling.}
     \end{subfigure}
     \vspace{-0.95mm}
     \begin{subfigure}[b]{0.48\columnwidth}
         \centering
         \includegraphics[width=\textwidth]{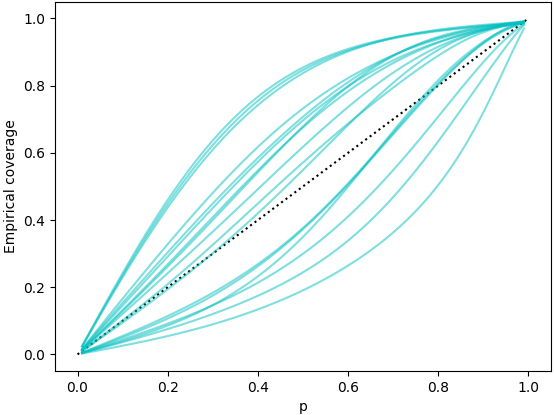}
         \caption{MC-dropout.}
     \end{subfigure}
     \vspace{-0.95mm}
        \caption{Depth completion - condensed calibration plots for ensembling and MC-dropout with $M=16$.}\vspace{-3.5mm}
        \label{fig:depth_completion_calibration_plots}
\end{figure}

In depth completion, we are given an image $x_{\mathrm{img}} \in  \mathbb{R}^{h \times w \times 3}$ from a forward-facing camera and an associated \emph{sparse} depth map $x_{\mathrm{sparse}} \in \mathbb{R}^{h \times w}$. Only non-zero pixels of $x_{\mathrm{sparse}}$ correspond to LiDAR depth measurements, projected onto the image plane. The goal is to predict a dense depth map $y \in \mathbb{R}^{h \times w}$ of the scene. We utilize the KITTI depth completion~\cite{Geiger2013IJRR, Uhrig2017THREEDV} and Virtual KITTI~\cite{Gaidon:Virtual:CVPR2016} datasets. KITTI depth completion contains more than $80\thinspace000$ images $x_{\mathrm{img}}$, sparse depth maps $x_{\mathrm{sparse}}$ and semi-dense target maps $y$. There are $1\thinspace000$ selected validation examples, which we use for evaluation. Only about $4\%$ of the pixels in $x_{\mathrm{sparse}}$ are non-zero and thus correspond to depth measurements. The semi-dense target maps are created by merging the LiDAR scans from $11$ consecutive frames into one, producing $y$ in which roughly $30\%$ of the pixels are non-zero. Virtual KITTI contains synthetic images $x_{\mathrm{img}}$ and dense depth maps $x_{\mathrm{dense}}$ extracted from $5$ driving sequences in a virtual world. It contains $2\thinspace126$ unique frames, of which there are $10$ different versions corresponding to various simulated weather and lighting conditions. We take sequence $0002$ as our validation set, leaving a total of $18\thinspace930$ training examples. We create targets $y$ for training by setting all pixels in $x_{\mathrm{dense}}$ corresponding to a depth $> 80$m to $0$, and then also randomly sample $5\%$ of the remaining non-zero pixels uniformly to create $x_{\mathrm{sparse}}$. We use the DNN model presented by Ma et al.~\cite{ma2018self}. The inputs $x_{\mathrm{img}}$, $x_{\mathrm{sparse}}$ are separately processed by initial convolutional layers, concatenated and fed to an encoder-decoder architecture based on ResNet34~\cite{he2016deep}. We employ the Gaussian model (\ref{eq:gaussian_model}) by duplicating the final layer, outputting both $\mu \in \mathbb{R}^{h \times w}$ and $\log\sigma^2 \in \mathbb{R}^{h \times w}$ instead of only the predicted depth map $\hat{y} \in \mathbb{R}^{h \times w}$. We also employ the same basic training procedure as Ma et al.~\cite{ma2018self} to train all our models, see Appendix~\ref{appendix:depth_completion_training_details} for details. For the MC-dropout comparison, we take inspiration from Kendall et al.~\cite{kendall2015bayesian} and place a dropout layer with drop probability $p = 0.5$ after the three last encoder blocks and the four first decoder blocks.

\parsection{Evaluation Metrics}
We evaluate the methods in terms of the \textit{Area Under the Sparsification Error curve (AUSE)} metric, as introduced by Ilg et al.~\cite{ilg2018uncertainty}. AUSE is a \emph{relative} measure of the uncertainty estimation quality, comparing the ordering of predictions induced by the estimated predictive uncertainty (sorted from least to most uncertain) with the ``oracle'' ordering in terms of the true prediction error. The metric thus reveals how well the estimated uncertainty can be used to sort predictions from worst (large true prediction error) to best (small prediction error). We compute AUSE in terms of \textit{Root Mean Squared Error (RMSE)} and based on all pixels in the entire evaluation dataset. A perfect AUSE score can however be achieved even if the true predictive uncertainty is consistently underestimated. As an \emph{absolute} measure of uncertainty estimation quality, we therefore also evaluate the methods in terms of calibration~\cite{brocker2009reliability, vaicenavicius2019evaluating}. In classification, the \textit{Expected Calibration Error (ECE)}~\cite{guo2017calibration, naeini2015obtaining} is a standard metric used to evaluate calibration. A well-calibrated model should then produce classification confidences which match the observed prediction accuracy, meaning that the model is not over-confident (outputting highly confident predictions which are incorrect), nor over-conservative. We here employ a metric that can be considered a natural generalization of ECE to the regression setting. Since our models output the mean $\mu \in \mathbb{R}$ and variance $\sigma^2 \in \mathbb{R}$ of a Gaussian distribution for each pixel, we can construct pixel-wise prediction intervals $\mu \pm \Phi^{-1}(\frac{p + 1}{2})\sigma$ of confidence level $p \in ]0, 1[$, where $\Phi$ is the CDF of the standard normal distribution. When computing the proportion of pixels for which the prediction interval covers the true target $y \in \mathbb{R}$, we expect this value, denoted $\hat{p}$, to equal $p \in ]0, 1[$ for a perfectly calibrated model. We compute the absolute error with respect to perfect calibration, $|p - \hat{p}|$, for $100$ values of $p \in ]0, 1[$ and use the area under this curve as our metric, which we call \textit{Area Under the Calibration Error curve (AUCE)}. Lastly, we also evaluate in terms of the standard RMSE metric.

\parsection{Results}
A comparison of ensembling and MC-dropout in terms of AUSE, AUCE and RMSE on the KITTI depth completion validation dataset is found in Figure~\ref{fig:depth_completion_ause_auce}. We observe in Figure~\ref{fig:depth_completion_ause} that ensembling consistently outperforms MC-dropout in terms of AUSE. However, the curves decrease as a function of $M$ in a similar manner. Sparsification plots and sparsification error curves are found in Appendix~\ref{appendix:depth_completion_additional_results}. A ranking of the methods can be more readily conducted based on Figure~\ref{fig:depth_completion_auce}, where we observe a clearly improving trend as $M$ increases for ensembling, whereas MC-dropout gets progressively worse. This result is qualitatively supported by the calibration plots found in Appendix~\ref{appendix:depth_completion_additional_results} and Figure~\ref{fig:depth_completion_calibration_plots}. Note that $M = 1$ corresponds to the baseline of only estimating aleatoric uncertainty.

\subsection{Street-Scene Semantic Segmentation}
\label{experiments-semantic_segmentation}

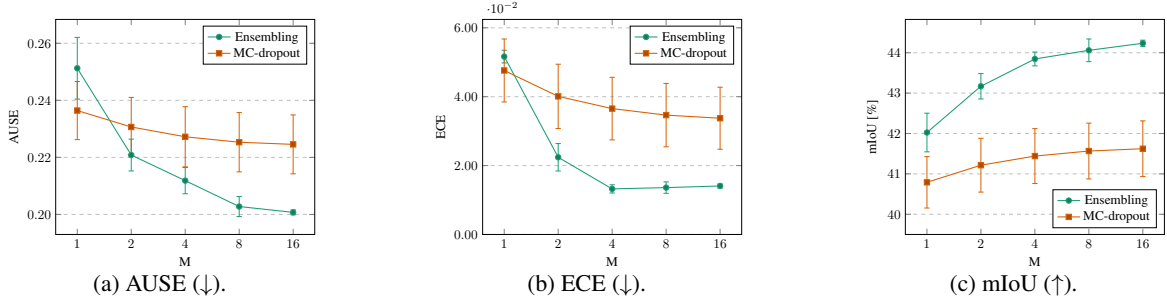
\begin{figure*}
\centering
    \begin{subfigure}[t]{0.32\textwidth}
      \centering
            \begin{tikzpicture}[scale=0.5025, baseline]
                \begin{axis}[
                    xmode=log,
                    log ticks with fixed point,
                    xlabel={M},
                    ylabel={AUSE},
                    xtick={1, 2, 4, 8, 16},
                    legend pos=north east,
                    ymajorgrids=true,
                    grid style=dashed,
                    y tick label style={
                        /pgf/number format/.cd,
                            fixed,
                            fixed zerofill,
                            precision=2,
                        /tikz/.cd
                    },
                    every axis plot/.append style={thick},
                ]
            \addplot
             plot [error bars/.cd, y dir = both, y explicit]
             table[row sep=crcr, x index=0, y index=1, y error index=2,]{
            1 0.251256 0.010828\\
            2 0.220826 0.00557298\\
            4 0.211844 0.00459671\\
            8 0.20277 0.00351327\\
            16 0.200717 0.000875494\\
            };
            \addlegendentry{Ensembling}
            
            \addplot
             plot [error bars/.cd, y dir = both, y explicit]
             table[row sep=crcr, x index=0, y index=1, y error index=2,]{
            1 0.236419 0.0101868\\
            2 0.230663 0.0103596\\
            4 0.227227 0.0105469\\
            8 0.225316 0.010397\\
            16 0.224559 0.010324\\
            };
            \addlegendentry{MC-dropout}
            
            \end{axis}
            \end{tikzpicture}\vspace{-2mm}%
      \caption{AUSE ($\downarrow$).}
      \label{fig:segmentation_ause}
    \end{subfigure}
    \vspace{-0.75mm}
    \begin{subfigure}[t]{0.32\textwidth}
      \centering
            \begin{tikzpicture}[scale=0.5025, baseline]
                \begin{axis}[
                    xmode=log,
                    log ticks with fixed point,
                    xlabel={M},
                    ylabel={ECE},
                    ymin=0.0,
                    xtick={1, 2, 4, 8, 16},
                    legend pos=north east,
                    ymajorgrids=true,
                    grid style=dashed,
                    y tick label style={
                        /pgf/number format/.cd,
                            fixed,
                            fixed zerofill,
                            precision=2,
                        /tikz/.cd
                    },
                    every axis plot/.append style={thick},
                ]
            \addplot
             plot [error bars/.cd, y dir = both, y explicit]
             table[row sep=crcr, x index=0, y index=1, y error index=2,]{
            1 0.0516515 0.00184791\\
            2 0.0224111 0.00398705\\
            4 0.0132305 0.00119393\\
            8 0.0136081 0.00165875\\
            16 0.0140804 0.000583427\\
            };
            \addlegendentry{Ensembling}
            
            \addplot
             plot [error bars/.cd, y dir = both, y explicit]
             table[row sep=crcr, x index=0, y index=1, y error index=2,]{
            1 0.0476137 0.00914593\\
            2 0.0400936 0.00933696\\
            4 0.0365439 0.00908598\\
            8 0.0346533 0.00919817\\
            16 0.0337501 0.00899849\\
            };
            \addlegendentry{MC-dropout} 

            \end{axis}
            \end{tikzpicture}\vspace{-2mm}%
      \caption{ECE ($\downarrow$).}
      \label{fig:segmentation_ece}
    \end{subfigure}
    \vspace{-0.75mm}
    \begin{subfigure}[t]{0.32\textwidth}
      \centering
            \begin{tikzpicture}[scale=0.5025, baseline]
                \begin{axis}[
                    xmode=log,
                    log ticks with fixed point,
                    xlabel={M},
                    ylabel={mIoU [\%]},
                    ymin=39.5,
                    xtick={1, 2, 4, 8, 16},
                    legend pos=south east,
                    ymajorgrids=true,
                    grid style=dashed,
                    every axis plot/.append style={thick},
                ]
            \addplot
             plot [error bars/.cd, y dir = both, y explicit]
             table[row sep=crcr, x index=0, y index=1, y error index=2,]{
            1 42.0244 0.478291\\
            2 43.1698 0.314294\\
            4 43.8462 0.170838\\
            8 44.0595 0.281014\\
            16 44.2341 0.0765837\\
            };
            \addlegendentry{Ensembling}
            
            \addplot
             plot [error bars/.cd, y dir = both, y explicit]
             table[row sep=crcr, x index=0, y index=1, y error index=2,]{
            1 40.7915 0.637076\\
            2 41.2153 0.666358\\
            4 41.4411 0.679762\\
            8 41.566 0.691324\\
            16 41.623 0.689694\\
            };
            \addlegendentry{MC-dropout}
            
            \end{axis}
            \end{tikzpicture}\vspace{-2mm}%
      \caption{mIoU ($\uparrow$).}
      \label{fig:segmentation_miou}
    \end{subfigure}%
    \vspace{-0.75mm}
  \caption{Street-scene semantic segmentation - quantitative results. The plots show a comparison of ensembling and MC-dropout in terms of AUSE, ECE and mIoU on the Cityscapes validation dataset, for different number of samples $M$.}\vspace{-3.5mm}
  \label{fig:segmentation_ause_ece_miou}
\end{figure*}

\begin{figure}
     \centering
     \begin{subfigure}[b]{0.48\columnwidth}
         \centering
         \includegraphics[width=\textwidth]{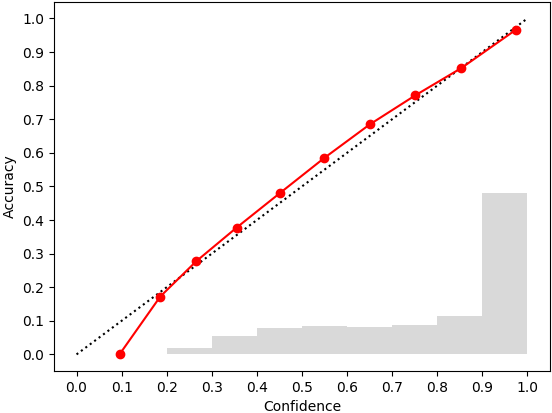}
         \caption{Ensembling.}
     \end{subfigure}
     \vspace{-0.95mm}
     \begin{subfigure}[b]{0.48\columnwidth}
         \centering
         \includegraphics[width=\textwidth]{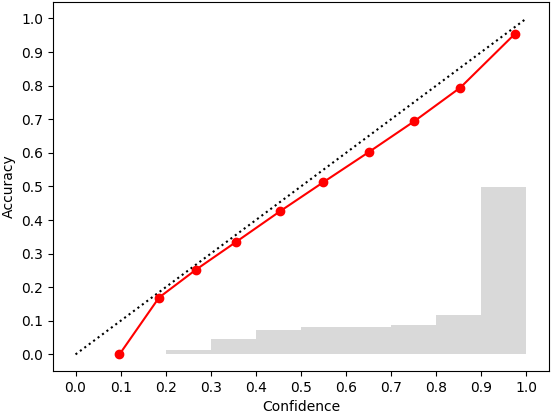}
         \caption{MC-dropout.}
     \end{subfigure}
     \vspace{-0.95mm}
        \caption{Street-scene semantic segmentation - example reliability diagrams for the two methods with $M = 16$.}\vspace{-3.5mm}
        \label{fig:segmentation_rel_diagrams}
\end{figure}

In this task, we are given an image $x \in \mathbb{R}^{h \times w \times 3}$ from a forward-facing camera. The goal is to predict $y$ of size $h \times w$, in which each pixel is assigned to one of $C$ different class labels (road, sidewalk, car, etc.). We utilize the popular Cityscapes~\cite{cordts2016cityscapes} and recent Synscapes~\cite{wrenninge2018synscapes} datasets. Cityscapes contains $5\thinspace000$ finely annotated images, mainly collected in various German cities. The annotations includes $30$ class labels, but only $C = 19$ are used in the training of models. Its validation set contains 500 examples, which we use for evaluation. Synscapes contains $25\thinspace000$ synthetic images, all captured in virtual urban environments. To match the size of Cityscapes, we randomly select $2\thinspace975$ of these for training and $500$ for validation. The images are annotated with the same class labels as Cityscapes. We use the DeepLabv3 DNN model presented by Chen et al.~\cite{chen2017rethinking}. The input image $x$ is processed by a ResNet101~\cite{he2016deep}, outputting a feature map of stride 8. The feature map is further processed by an ASPP module and a $1 \times 1$ convolutional layer, outputting logits at 1/8 of the original resolution. These are then upsampled to image resolution using bilinear interpolation. The conventional Categorical model~(\ref{eq:categorical_model}) is thus used for each pixel. We base our implementation on the one by Yuan and Wang~\cite{yuan2018ocnet}, and also follow the same basic training procedure, see Appendix~\ref{appendix:street-scene_semantic_segmentation_training_details} for details. For reference, the model obtains an mIoU~\cite{long2015fully} of $76.04\%$ when trained on Cityscapes and evaluated on its validation set. For the MC-dropout comparison, we take inspiration from Mukhoti and Gal~\cite{mukhoti2018evaluating} and place a dropout layer with $p = 0.5$ after the four last ResNet blocks.

\parsection{Evaluation Metrics}
As for depth completion, we evaluate the methods in terms of the AUSE metric. In this classification setting, we compare the ``oracle'' ordering of predictions with the one induced by the predictive entropy. We compute AUSE in terms of Brier score and based on all pixels in the evaluation dataset. We also evaluate in terms of calibration by the ECE metric~\cite{guo2017calibration, naeini2015obtaining}. All predictions are here partitioned into $L$ bins based on the maximum assigned confidence. For each bin, the difference between the average predicted confidence and the actual accuracy is then computed, and ECE is obtained as the weighted average of these differences. We use $L = 10$ bins of equal size.

\parsection{Results}
A comparison of ensembling and MC-dropout in terms of AUSE, ECE and mIoU on the Cityscapes validation dataset is found in Figure~\ref{fig:segmentation_ause_ece_miou}. We observe that the metrics clearly improve as functions of $M$ for both ensembling and MC-dropout, demonstrating the importance of epistemic uncertainty estimation. The rate of improvement is generally greater for ensembling. For ECE, we observe in Figure~\ref{fig:segmentation_ece} a drastic improvement for ensembling as $M$ is increased, followed by a distinct plateau. According to the condensed reliability diagrams in Appendix~\ref{appendix:street-scene_semantic_segmentation_additional_results}, this corresponds to a transition from clear model over-confidence to slight over-conservatism. For MC-dropout, the corresponding diagrams suggest a stagnation while the model still is somewhat over-confident. Example reliability diagrams for $M = 16$ are shown in Figure~\ref{fig:segmentation_rel_diagrams}, in which this over-confidence for MC-dropout can be observed. Note that the relatively low mIoU scores reported in Figure~\ref{fig:segmentation_miou}, obtained by models trained exclusively on Synscapes, are expected~\cite{wrenninge2018synscapes} and caused by the intentionally challenging domain gap between synthetic and real-world data.
\section{Discussion \& Conclusion}
\label{discussion_and_conclusion}

We proposed a comprehensive evaluation framework for \emph{scalable} epistemic uncertainty estimation methods in deep learning. The proposed framework is specifically designed to test the robustness required in real-world computer vision applications. We applied our proposed framework and provided the first properly extensive and conclusive comparison of ensembling and MC-dropout, the results of which demonstrates that ensembling consistently provides more reliable and practically useful uncertainty estimates. We attribute the success of ensembling to its ability, due to the random initialization, to capture the important aspect of multi-modality present in the posterior distribution $p(\theta | \mathcal{D})$ of DNNs. MC-dropout has a large design-space compared to ensembling, and while careful tuning of MC-dropout potentially could close the performance gap on individual tasks, the simplicity and general applicability of ensembling must be considered key strengths. The main drawback of both methods is the computational cost at test time that grows linearly with $M$, limiting real-time applicability. Here, future work includes exploring the effect of model pruning techniques~\cite{yang2017designing, han2015deep} on predictive uncertainty quality. For ensembling, sharing early stages of the DNN among ensemble members is also an interesting future direction. A weakness of ensembling is the additional training required, which also scales linearly with $M$. The training of different ensemble members can however be performed in parallel, making it less of an issue in practice given appropriate computing infrastructure. In conclusion, our work suggests that ensembling should be considered the new go-to method for \emph{scalable} epistemic uncertainty estimation.

\parsection{Acknowledgments}
This research was supported by the Swedish Foundation for Strategic Research via the project \emph{ASSEMBLE} and by the Swedish Research Council via the project \emph{Learning flexible models for nonlinear dynamics}.

{\small
\bibliographystyle{ieee_fullname}
\bibliography{references}
}

\newpage

\onecolumn

\section*{Supplementary Material}
This is the supplementary material for \textbf{\textit{Evaluating Scalable Bayesian Deep Learning Methods for Robust Computer Vision}}. It consists of Appendix~\ref{appendix:gaussian_approximation_of_the_predictive_distribution}-\ref{appendix:street-scene_semantic_segmentation}.

\appendix
\begin{appendices}
\section{Approximating a Mixture of Gaussian Distributions}
\label{appendix:gaussian_approximation_of_the_predictive_distribution}

For the Gaussian model~(\ref{eq:gaussian_model}), $\hat{p}(y^\star | x^\star, \mathcal{D})$ in~(\ref{eq:approx_predictive_distribution_Bayesian}) is a uniformly weighted mixture of Gaussian distributions. We approximate this mixture with a single Gaussian parameterized by the mixture mean and variance:
\begin{equation*}
\begin{split}
& \hat{p}(y^\star | x^\star, \mathcal{D}) = \frac{1}{M} \sum_{i=1}^{M}  p(y^\star | x^\star, \theta^{(i)}), \quad \theta^{(i)} \sim q(\theta), \\
& \hat{p}(y^\star | x^\star, \mathcal{D}) = \frac{1}{M} \sum_{i=1}^{M} \mathcal{N}(y^\star; \mu_{\theta^{(i)}}(x^\star),\,\sigma^2_{\theta^{(i)}}(x^\star)), \quad \theta^{(i)} \sim q(\theta),\\
& \hat{p}(y^\star | x^\star, \mathcal{D}) \approx \mathcal{N}(y^\star; \hat{\mu}(x^\star),\,\hat{\sigma}^2(x^\star)),
\end{split}
\end{equation*}
\begin{equation*}
\hat{\mu}(x) = \frac{1}{M} \sum_{i=1}^{M} \mu_{\theta^{(i)}}(x), \quad  \hat{\sigma}^2(x) = \frac{1}{M} \sum_{i=1}^{M} \bigg( \big(\mu_{\theta^{(i)}}(x) - \hat{\mu}(x)\big)^2 + \sigma^2_{\theta^{(i)}}(x) \bigg), \quad \theta^{(i)} \sim q(\theta).
\end{equation*}
\section{Illustrative Toy Problems}
\label{appendix:illustrative_toy_problems}

In this appendix, further details on the illustrative toy problems experiments (Section~\ref{experiments-illustrative_toy_problems}) are provided.

\subsection{Experimental Setup}
\label{appendix:illustrative_toy_problem_experimental_setup}

Figure~\ref{fig:toy_problems_quantative_results_regression} (regression) shows $D_\mathrm{KL}(p~\parallel~p_{\mathrm{HMC}})$ computed on $[-7, 7]$. All training data was given in $[-3, 3]$.

Figure~\ref{fig:toy_problems_quantative_results_classification} (classification) shows $D_\mathrm{KL}(p~\parallel~p_{\mathrm{HMC}})$ computed on the region $-6 \leq  x_1 \leq 6$, $-6 \leq  x_2 \leq 6$. All training data was given in the region $0 \leq  x_1 \leq 3$, $-3 \leq  x_2 \leq 3$.

For regression, $D_\mathrm{KL}(p~\parallel~p_{\mathrm{HMC}})$ is computed using the formula for KL divergence between two Gaussian distributions $p_1(x) = \mathcal{N}(x; \mu_1, \sigma_1^2)$, $p_2(x) = \mathcal{N}(x; \mu_2, \sigma_2^2)$:
\begin{equation*}
    D_\mathrm{KL}(p_1~\parallel~p_2) = \log \frac{\sigma_2}{\sigma_1} + \frac{\sigma_1^2 + (\mu_1 - \mu_2)^2}{2\sigma_2^2} - \frac{1}{2}.
\end{equation*}

For classification, $D_\mathrm{KL}(p~\parallel~p_{\mathrm{HMC}})$ is computed using the formula for KL divergence between two discrete distributions $q_1(x)$, $q_2(x)$:
\begin{equation*}
    D_\mathrm{KL}(q_1~\parallel~q_2) = \sum_{x \in \mathcal{X}}^{} q_1(x) \log \frac{q_1(x)}{q_2(x)}.
\end{equation*}

For both regression and classification, HMC with prior $p(\theta)~=~\mathcal{N}(0, I_{P})$ and $M=1\thinspace000$ samples is implemented using Pyro~\cite{bingham2018pyro}. Specifically, we use pyro.infer.mcmc.MCMC with pyro.infer.mcmc.NUTS as kernel, $\text{num}\_\text{samples} = 1\thinspace000$ and $\text{warmup}\_\text{steps} = 1\thinspace000$.

\subsection{Implementation Details}
\label{appendix:illustrative_toy_problem_details}

For regression, we use the Gaussian model~(\ref{eq:gaussian_model}) with two separate feed-forward neural networks outputting $\mu_{\theta}(x)$ and $\log\sigma^2_{\theta}(x)$. Both neural networks have $2$ hidden layers of size $10$.

For classification, we use the Categorical model~(\ref{eq:categorical_model}) with a feed-forward neural network with $2$ hidden layers of size $10$.

For the MC-dropout comparison, we place a dropout layer after the first hidden layer of each neural network. For regression, we use a drop probability $p = 0.2$. For classification, we use $p = 0.1$.

For ensembling, we train all ensemble models for $150$ epochs with the Adam optimizer, a batch size of $32$ and a fixed learning rate of $0.001$. 

For MC-dropout, we train models for $300$ epochs with the Adam optimizer, a batch size of $32$ and a fixed learning rate of $0.001$.

For ensembling and MC-dropout, we minimize the MAP objective $-\log p(Y | X, \theta)p(\theta)$. In our case where the model parameters $\theta \in \mathbb{R}^P$ and $p(\theta)~=~\mathcal{N}(0, I_{P})$, this corresponds to the following loss for regression:
\begin{equation*}
    L(\theta) = \frac{1}{N} \sum_{i=1}^{N} \frac{(y_i - \hat{\mu}(x_i))^2}{\hat{\sigma}^{2}(x_i)} + \log \hat{\sigma}^{2}(x_i) + \frac{1}{N} \theta^{\Transp}\theta.
\end{equation*}
For classification, where $y_i = \rvect{y_{i, 1} \dots y_{i, C}}^{\Transp}$ (one-hot encoded) and $\hat{s}(x_i) = \rvect{\hat{s}(x_i)_1 \dots \hat{s}(x_i)_C}^{\Transp}$ is the $\Softmax$ output, it corresponds to the following loss:
\begin{equation*}
    L(\theta) = -\frac{1}{N} \sum_{i=1}^{N} \sum_{k=1}^{C} y_{i, k} \log \hat{s}(x_i)_k + \frac{1}{2N} \theta^{\Transp}\theta.
\end{equation*}

For SGLD, we extract samples from the parameter trajectory given by the update equation:
\begin{equation*}
   \theta_{t+1} = \theta_{t} - \alpha_{t} \nabla_{\theta}\widetilde{U}(\theta_{t}) + \sqrt{2\alpha_{t}}\epsilon_{t},
\end{equation*}
where $\epsilon_{t} \sim \mathcal{N}(0, 1)$, $\nabla_{\theta}\widetilde{U}(\theta)$ is the stochastic gradient of $U(\theta) = -\log p(Y | X, \theta)p(\theta)$ and $\alpha_t$ is the stepsize. We run it for a total number of steps corresponding to $256 \cdot 150$ epochs with a batch size of $32$. The stepsize $\alpha_t$ is decayed according to:
\begin{equation*}
    \alpha_t = \alpha_0(1 - \frac{t}{T})^{0.9}, \quad t = 1, 2, \dots, T,
\end{equation*}
where $T$ is the total number of steps, $\alpha_0 = 0.01$ (the initial stepsize) for regression and $\alpha_0 = 0.05$ for classification. $M \in \{8, 16, 32, 64, 128, 256\}$ samples are extracted starting at step $t = \text{int}(0.75T)$, ending at step $t = T$ and spread out evenly between. 

For SGHMC, we extract samples from the parameter trajectory given by the update equation:
\begin{equation*}
    \begin{split}
        & \theta_{t+1} = \theta_{t} + r_t,\\
        & r_{t+1} = (1 - \eta)r_t - \alpha_t \nabla_{\theta}\widetilde{U}(\theta_{t}) + \sqrt{2\eta\alpha_{t}}\epsilon_{t},
    \end{split}
\end{equation*}
where $\epsilon_{t} \sim \mathcal{N}(0, 1)$, $\nabla_{\theta}\widetilde{U}(\theta)$ is the stochastic gradient of $U(\theta) = -\log p(Y | X, \theta)p(\theta)$, $\alpha_t$ is the stepsize and $\eta = 0.1$. We run it for a total number of steps corresponding to $256 \cdot 150$ epochs with a batch size of $32$. The stepsize $\alpha_t$ is decayed according to:
\begin{equation*}
    \alpha_t = \alpha_0(1 - \frac{t}{T})^{0.9}, \quad t = 1, 2, \dots, T,
\end{equation*}
where $T$ is the total number of steps, $\alpha_0 = 0.001$ (the initial stepsize) for regression and $\alpha_0 = 0.01$ for classification. $M \in \{8, 16, 32, 64, 128, 256\}$ samples are extracted starting at step $t = \text{int}(0.75T)$, ending at step $t = T$ and spread out evenly between.

For all models, we randomly initialize the parameters $\theta$ using the default initializer in PyTorch.

\subsection{Description of Results}
\label{appendix:illustrative_toy_problems_results_description}

The results in Figure~\ref{fig:toy_problems_quantative_results_regression}, \ref{fig:toy_problems_quantative_results_classification} were obtained in the following way:
\begin{itemize}
    \item \textbf{Ensembling:} $1024$ models were trained using the same training procedure, the mean and standard deviation was computed based on $1024/M$ unique sets of models for $M~\in~\{8, 16, 32, 64, 128, 256\}$.  
    
    \item \textbf{MC-dropout:} $10$ models were trained using the same training procedure, based on which the mean and standard deviation was computed. 
    
    \item \textbf{SGLD:} $6$ models were trained using the same training procedure, based on which the mean and standard deviation was computed.
    
    \item \textbf{SGHMC:} $6$ models were trained using the same training procedure, based on which the mean and standard deviation was computed. 
\end{itemize}

\subsection{Additional Results}
\label{appendix:illustrative_toy_problem_additional_results}

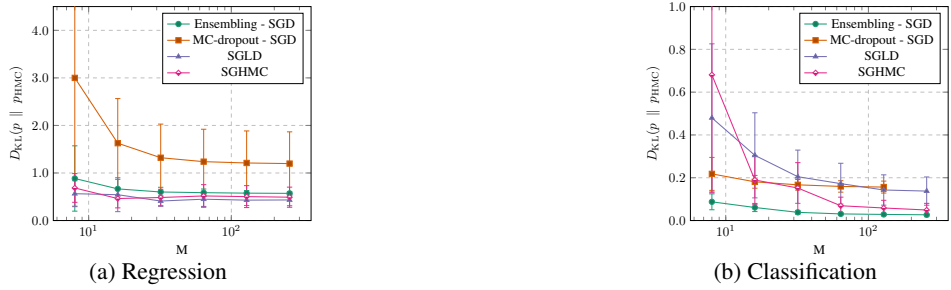
\begin{figure}
\centering
    \begin{subfigure}[t]{0.48\textwidth}
      \centering
            \begin{tikzpicture}[scale=0.50, baseline]
                \begin{axis}[
                    xmode=log,
                    xlabel={M},
                    ylabel={$D_\mathrm{KL}(p~\parallel~p_{\mathrm{HMC}})$},
                    ymin=0.0, ymax=4.5,
                    legend pos=north east,
                    ymajorgrids=true,
                    xmajorgrids=true,
                    grid style=dashed,
                    y tick label style={
                        /pgf/number format/.cd,
                            fixed,
                            fixed zerofill,
                            precision=1,
                        /tikz/.cd
                    },
                    every axis plot/.append style={thick},
                ]
            \addplot
             plot [error bars/.cd, y dir = both, y explicit]
             table[row sep=crcr, x index=0, y index=1, y error index=2,]{
            8 0.884296 0.687031\\
            16 0.665706 0.201289\\
            32 0.600385 0.0990562\\
            64 0.586071 0.077102\\
            128 0.57588 0.0547483\\
            256 0.571333 0.0382014\\
            };
            \addlegendentry{Ensembling - SGD} 
            
            \addplot
             plot [error bars/.cd, y dir = both, y explicit]
             table[row sep=crcr, x index=0, y index=1, y error index=2,]{
            8 3.00014 2.69764\\
            16 1.62673 0.937929\\
            32 1.32121 0.706895\\
            64 1.23829 0.681647\\
            128 1.21055 0.673814\\
            256 1.19745 0.667877\\
            };
            \addlegendentry{MC-dropout - SGD} 
            
            \addplot
             plot [error bars/.cd, y dir = both, y explicit]
             table[row sep=crcr, x index=0, y index=1, y error index=2,]{
            8 0.564368 0.272595\\
            16 0.543245 0.356657\\
            32 0.410131 0.111204\\
            64 0.449033 0.152189\\
            128 0.428836 0.113887\\
            256 0.433093 0.121152\\
            };
            \addlegendentry{SGLD} 
            
            \addplot
             plot [error bars/.cd, y dir = both, y explicit]
             table[row sep=crcr, x index=0, y index=1, y error index=2,]{
            8 0.684939 0.301885\\
            16 0.46253 0.196306\\
            32 0.484642 0.17294\\
            64 0.516942 0.237645\\
            128 0.503064 0.230595\\
            256 0.490817 0.211457\\
            };
            \addlegendentry{SGHMC} 
            
            \end{axis}
            \end{tikzpicture}\vspace{-2mm}%
      \caption{Regression}
    \end{subfigure}
    \begin{subfigure}[t]{0.48\textwidth}
      \centering
            \begin{tikzpicture}[scale=0.50, baseline]
                \begin{axis}[
                    xmode=log,
                    xlabel={M},
                    ylabel={$D_\mathrm{KL}(p~\parallel~p_{\mathrm{HMC}})$},
                    ymin=0.0, ymax=1.0,
                    legend pos=north east,
                    ymajorgrids=true,
                    xmajorgrids=true,
                    grid style=dashed,
                    y tick label style={
                        /pgf/number format/.cd,
                            fixed,
                            fixed zerofill,
                            precision=1,
                        /tikz/.cd
                    },
                    every axis plot/.append style={thick},
                ]
            \addplot 
             plot [error bars/.cd, y dir = both, y explicit]
             table[row sep=crcr, x index=0, y index=1, y error index=2,]{
            8 0.0874148 0.0370799\\
            16 0.0609295 0.0190791\\
            32 0.0383158 0.00900817\\
            64 0.0309371 0.00525672\\
            128 0.0285367 0.00256969\\
            256 0.0268412 0.00196269\\
            };
            \addlegendentry{Ensembling - SGD} 
            
            \addplot
             plot [error bars/.cd, y dir = both, y explicit]
             table[row sep=crcr, x index=0, y index=1, y error index=2,]{
            8 0.218017 0.0770154\\
            16 0.180943 0.0299291\\
            32 0.166855 0.024802\\
            64 0.159344 0.0268131\\
            128 0.156366 0.02794\\
            };
            \addlegendentry{MC-dropout - SGD} 
            
            \addplot
             plot [error bars/.cd, y dir = both, y explicit]
             table[row sep=crcr, x index=0, y index=1, y error index=2,]{
            8 0.479821 0.345935\\
            16 0.304685 0.198831\\
            32 0.204824 0.124369\\
            64 0.172374 0.0952615\\
            128 0.142729 0.0707726\\
            256 0.137614 0.0660084\\
            };
            \addlegendentry{SGLD} 
            
            \addplot
             plot [error bars/.cd, y dir = both, y explicit]
             table[row sep=crcr, x index=0, y index=1, y error index=2,]{
            8 0.680847 0.550977\\
            16 0.189506 0.113788\\
            32 0.151961 0.118518\\
            64 0.0690482 0.0401916\\
            128 0.0586131 0.0359652\\
            256 0.0491939 0.0310982\\
            };
            \addlegendentry{SGHMC} 
            
            \end{axis}
            \end{tikzpicture}\vspace{-2mm}%
      \caption{Classification}
    \end{subfigure}
  \caption{Illustrative toy problems, quantitative results. SGD is used for ensembling and MC-dropout instead of Adam.}
  \label{fig:toy_problems_quantative_results_SGD}
\end{figure}

Figure~\ref{fig:toy_problems_quantative_results_SGD} and Figure~\ref{fig:toy_problems_quantative_results_SGDMOM} show the same comparison as Figure~\ref{fig:toy_problems_quantative_results_regression}, \ref{fig:toy_problems_quantative_results_classification}, but using SGD and SGD with momentum for ensembling and MC-dropout, respectively. We observe that ensembling consistently outperforms the compared methods for classification, but that SGLD and SGHMC has better performance for regression in these cases. SGLD and SGHMC are however trained for 256 times longer than each ensemble model, complicating the comparison somewhat. If SGLD and SGHMC instead are trained for just 64 times longer than each ensemble model, we observe in Figure~\ref{fig:toy_problems_quantative_results_SGDMOM_less_training} that they are consistently outperformed by ensembling.

\begin{figure}
\centering
    \begin{subfigure}[t]{0.48\textwidth}
      \centering
            \begin{tikzpicture}[scale=0.50, baseline]
                \begin{axis}[
                    xmode=log,
                    xlabel={M},
                    ylabel={$D_\mathrm{KL}(p~\parallel~p_{\mathrm{HMC}})$},
                    ymin=0.0, ymax=8.5,
                    legend pos=north east,
                    ymajorgrids=true,
                    xmajorgrids=true,
                    grid style=dashed,
                    y tick label style={
                        /pgf/number format/.cd,
                            fixed,
                            fixed zerofill,
                            precision=1,
                        /tikz/.cd
                    },
                    every axis plot/.append style={thick},
                ]
                
            \addplot
             plot [error bars/.cd, y dir = both, y explicit]
             table[row sep=crcr, x index=0, y index=1, y error index=2,]{
            8 15.716 12.902\\
            16 12.523 10.952\\
            32 8.1706 6.9662\\
            64 5.06537 4.5256\\
            128 3.32757 2.55369\\
            256 2.98532 2.16758\\
            };
            \addlegendentry{$p = 0.05$} 
                
            \addplot
             plot [error bars/.cd, y dir = both, y explicit]
             table[row sep=crcr, x index=0, y index=1, y error index=2,]{
            8 5.44736 4.22986\\
            16 2.24801 2.05129\\
            32 1.30724 0.726443\\
            64 1.14724 0.54791\\
            128 1.09269 0.502092\\
            256 1.0683 0.478516\\
            };
            \addlegendentry{$p = 0.1$} 
            
            \addplot
             plot [error bars/.cd, y dir = both, y explicit]
             table[row sep=crcr, x index=0, y index=1, y error index=2,]{
            8 1.62725 1.16544\\
            16 1.20143 0.77387\\
            32 1.07685 0.623371\\
            64 1.01024 0.538445\\
            128 0.969594 0.502551\\
            256 0.94703 0.491882\\
            };
            \addlegendentry{*$p = 0.2$} 
            
            \end{axis}
            \end{tikzpicture}\vspace{-2mm}%
      \caption{Regression}
    \end{subfigure}
    \begin{subfigure}[t]{0.48\textwidth}
      \centering
            \begin{tikzpicture}[scale=0.50, baseline]
                \begin{axis}[
                    xmode=log,
                    xlabel={M},
                    ylabel={$D_\mathrm{KL}(p~\parallel~p_{\mathrm{HMC}})$},
                    ymin=0.0, ymax=1.25,
                    legend pos=north east,
                    ymajorgrids=true,
                    xmajorgrids=true,
                    grid style=dashed,
                    y tick label style={
                        /pgf/number format/.cd,
                            fixed,
                            fixed zerofill,
                            precision=1,
                        /tikz/.cd
                    },
                    every axis plot/.append style={thick},
                ]

            \addplot
             plot [error bars/.cd, y dir = both, y explicit]
             table[row sep=crcr, x index=0, y index=1, y error index=2,]{
            8 0.997535 0.90124\\
            16 0.724178 0.704072\\
            32 0.452422 0.404351\\
            64 0.28174 0.209364\\
            128 0.237568 0.164092\\
            };
            \addlegendentry{$p = 0.05$} 
                
            \addplot
             plot [error bars/.cd, y dir = both, y explicit]
             table[row sep=crcr, x index=0, y index=1, y error index=2,]{
            8 0.312933 0.141196\\
            16 0.210738 0.072815\\
            32 0.148762 0.0419286\\
            64 0.129146 0.0369342\\
            128 0.122003 0.0358212\\
            };
            \addlegendentry{*$p = 0.1$} 
                
            \addplot
             plot [error bars/.cd, y dir = both, y explicit]
             table[row sep=crcr, x index=0, y index=1, y error index=2,]{
            8 0.249316 0.155723\\
            16 0.180024 0.0605716\\
            32 0.157055 0.0377527\\
            64 0.148327 0.0315233\\
            128 0.144578 0.0292545\\
            };
            \addlegendentry{$p = 0.2$} 
            
            \addplot
             plot [error bars/.cd, y dir = both, y explicit]
             table[row sep=crcr, x index=0, y index=1, y error index=2,]{
            8 0.269312 0.0231495\\
            16 0.258393 0.0135039\\
            32 0.255614 0.0133051\\
            64 0.253753 0.0128161\\
            128 0.253105 0.0131291\\
            };
            \addlegendentry{$p = 0.5$} 
            
            \end{axis}
            \end{tikzpicture}\vspace{-2mm}%
      \caption{Classification}
    \end{subfigure}
  \caption{Illustrative toy problems, quantitative results. MC-dropout using Adam.}
  \label{fig:toy_problems_quantative_results_mcdropout_adam}
\end{figure}
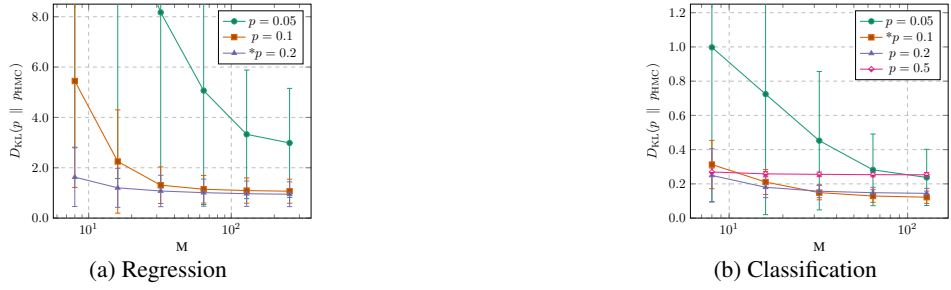

For MC-dropout using Adam, we also varied the drop probability $p$ and chose the best performing variant. These results are found in Figure~\ref{fig:toy_problems_quantative_results_mcdropout_adam}, in which * marks the chosen variant.

\begin{figure}
\centering
    \begin{subfigure}[t]{0.48\textwidth}
      \centering
            \begin{tikzpicture}[scale=0.50, baseline]
                \begin{axis}[
                    xmode=log,
                    xlabel={M},
                    ylabel={$D_\mathrm{KL}(p~\parallel~p_{\mathrm{HMC}})$},
                    ymin=0.0, ymax=4.5,
                    legend pos=north east,
                    ymajorgrids=true,
                    xmajorgrids=true,
                    grid style=dashed,
                    y tick label style={
                        /pgf/number format/.cd,
                            fixed,
                            fixed zerofill,
                            precision=1,
                        /tikz/.cd
                    },
                    every axis plot/.append style={thick},
                ]
            \addplot
             plot [error bars/.cd, y dir = both, y explicit]
             table[row sep=crcr, x index=0, y index=1, y error index=2,]{
            8 1.31327 1.21182\\
            16 0.873464 0.291195\\
            32 0.764788 0.197846\\
            64 0.722799 0.127398\\
            128 0.70546 0.0846264\\
            256 0.698855 0.07611\\
            };
            \addlegendentry{Ensembling - SGDMOM} 
            
            \addplot
             plot [error bars/.cd, y dir = both, y explicit]
             table[row sep=crcr, x index=0, y index=1, y error index=2,]{
            8 3.15305 2.6672\\
            16 1.54568 0.729629\\
            32 1.07238 0.386907\\
            64 0.943156 0.353\\
            128 0.899017 0.352454\\
            256 0.885846 0.354435\\
            };
            \addlegendentry{MC-dropout - SGDMOM} 
            
            \addplot
             plot [error bars/.cd, y dir = both, y explicit]
             table[row sep=crcr, x index=0, y index=1, y error index=2,]{
            8 0.564368 0.272595\\
            16 0.543245 0.356657\\
            32 0.410131 0.111204\\
            64 0.449033 0.152189\\
            128 0.428836 0.113887\\
            256 0.433093 0.121152\\
            };
            \addlegendentry{SGLD} 
            
            \addplot
             plot [error bars/.cd, y dir = both, y explicit]
             table[row sep=crcr, x index=0, y index=1, y error index=2,]{
            8 0.684939 0.301885\\
            16 0.46253 0.196306\\
            32 0.484642 0.17294\\
            64 0.516942 0.237645\\
            128 0.503064 0.230595\\
            256 0.490817 0.211457\\
            };
            \addlegendentry{SGHMC} 
            
            \end{axis}
            \end{tikzpicture}\vspace{-2mm}%
      \caption{Regression}
    \end{subfigure}
    \begin{subfigure}[t]{0.48\textwidth}
      \centering
            \begin{tikzpicture}[scale=0.50, baseline]
                \begin{axis}[
                    xmode=log,
                    xlabel={M},
                    ylabel={$D_\mathrm{KL}(p~\parallel~p_{\mathrm{HMC}})$},
                    ymin=0.0, ymax=1.0,
                    legend pos=north east,
                    ymajorgrids=true,
                    xmajorgrids=true,
                    grid style=dashed,
                    y tick label style={
                        /pgf/number format/.cd,
                            fixed,
                            fixed zerofill,
                            precision=1,
                        /tikz/.cd
                    },
                    every axis plot/.append style={thick},
                ]
            \addplot 
             plot [error bars/.cd, y dir = both, y explicit]
             table[row sep=crcr, x index=0, y index=1, y error index=2,]{
            8 0.214353 0.168936\\
            16 0.109072 0.0627925\\
            32 0.0547409 0.0320485\\
            64 0.033561 0.0155624\\
            128 0.0273158 0.009843\\
            256 0.0199156 0.000930452\\
            };
            \addlegendentry{Ensembling - SGDMOM} 
            
            \addplot
             plot [error bars/.cd, y dir = both, y explicit]
             table[row sep=crcr, x index=0, y index=1, y error index=2,]{
            8 0.343153 0.189491\\
            16 0.229002 0.102032\\
            32 0.175187 0.0587089\\
            64 0.151353 0.0402061\\
            128 0.142109 0.0371594\\
            };
            \addlegendentry{MC-dropout - SGDMOM} 
            
            \addplot
             plot [error bars/.cd, y dir = both, y explicit]
             table[row sep=crcr, x index=0, y index=1, y error index=2,]{
            8 0.479821 0.345935\\
            16 0.304685 0.198831\\
            32 0.204824 0.124369\\
            64 0.172374 0.0952615\\
            128 0.142729 0.0707726\\
            256 0.137614 0.0660084\\
            };
            \addlegendentry{SGLD} 
            
            \addplot
             plot [error bars/.cd, y dir = both, y explicit]
             table[row sep=crcr, x index=0, y index=1, y error index=2,]{
            8 0.680847 0.550977\\
            16 0.189506 0.113788\\
            32 0.151961 0.118518\\
            64 0.0690482 0.0401916\\
            128 0.0586131 0.0359652\\
            256 0.0491939 0.0310982\\
            };
            \addlegendentry{SGHMC} 
            
            \end{axis}
            \end{tikzpicture}\vspace{-2mm}%
      \caption{Classification}
    \end{subfigure}
  \caption{Illustrative toy problems, quantitative results. SGD with momentum is used for ensembling and MC-dropout instead of Adam.}
  \label{fig:toy_problems_quantative_results_SGDMOM}
\end{figure}
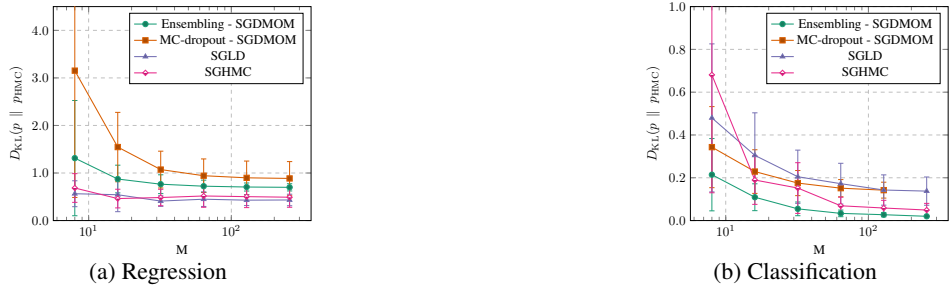

\begin{figure}
\centering
            \begin{tikzpicture}[scale=0.50, baseline]
                \begin{axis}[
                    xmode=log,
                    xlabel={M},
                    ylabel={$D_\mathrm{KL}(p~\parallel~p_{\mathrm{HMC}})$},
                    ymin=0.0, ymax=11,
                    legend pos=north east,
                    ymajorgrids=true,
                    xmajorgrids=true,
                    grid style=dashed,
                    y tick label style={
                        /pgf/number format/.cd,
                            fixed,
                            fixed zerofill,
                            precision=1,
                        /tikz/.cd
                    },
                    every axis plot/.append style={thick},
                ]
            \addplot
             plot [error bars/.cd, y dir = both, y explicit]
             table[row sep=crcr, x index=0, y index=1, y error index=2,]{
            8 1.31327 1.21182\\
            16 0.873464 0.291195\\
            32 0.764788 0.197846\\
            64 0.722799 0.127398\\
            128 0.70546 0.0846264\\
            256 0.698855 0.07611\\
            };
            \addlegendentry{Ensembling - SGDMOM} 
            
            \addplot
             plot [error bars/.cd, y dir = both, y explicit]
             table[row sep=crcr, x index=0, y index=1, y error index=2,]{
            8 3.15305 2.6672\\
            16 1.54568 0.729629\\
            32 1.07238 0.386907\\
            64 0.943156 0.353\\
            128 0.899017 0.352454\\
            256 0.885846 0.354435\\
            };
            \addlegendentry{MC-dropout - SGDMOM} 
            
            \addplot
             plot [error bars/.cd, y dir = both, y explicit]
             table[row sep=crcr, x index=0, y index=1, y error index=2,]{
            8 1.7203 1.5348\\
            16 1.19723 0.790489\\
            32 1.37389 1.17095\\
            64 1.16624 0.87977\\
            128 1.1056 0.733211\\
            256 1.08937 0.720269\\
            };
            \addlegendentry{SGLD-64} 
            
            \addplot
             plot [error bars/.cd, y dir = both, y explicit]
             table[row sep=crcr, x index=0, y index=1, y error index=2,]{
            8 7.0836 5.9555\\
            16 3.09207 2.16595\\
            32 6.05611 5.1274\\
            64 4.57992 3.78068\\
            128 4.70943 3.21936\\
            256 3.84556 2.73831\\
            };
            \addlegendentry{SGHMC-64} 
            
            \end{axis}
            \end{tikzpicture}\vspace{-2mm}%
  \caption{Illustrative toy regression problem, quantitative results. SGD with momentum is used for ensembling and MC-dropout instead of Adam. Less training for SGLD and SGHMC.}
  \label{fig:toy_problems_quantative_results_SGDMOM_less_training}
\end{figure}
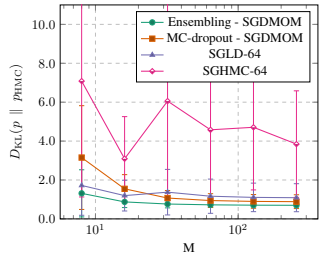

\subsection{Qualitative Results}
\label{appendix:illustrative_toy_problem_qualitative_results}

\begin{figure}
     \centering
     \begin{subfigure}[b]{0.195\textwidth}
         \centering
         \includegraphics[width=\textwidth]{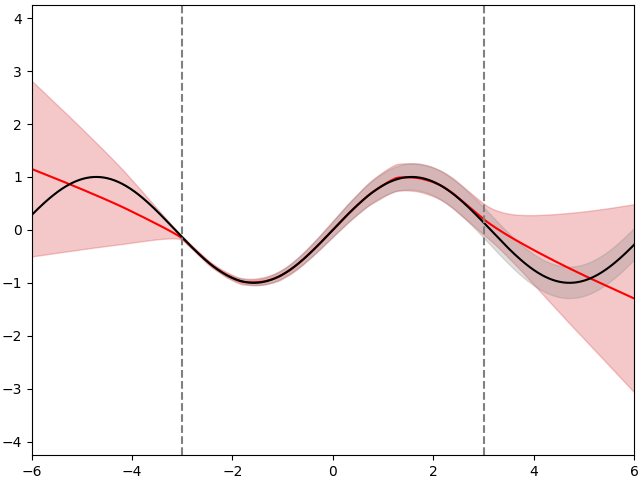}
     \end{subfigure}
     \begin{subfigure}[b]{0.195\textwidth}
         \centering
         \includegraphics[width=\textwidth]{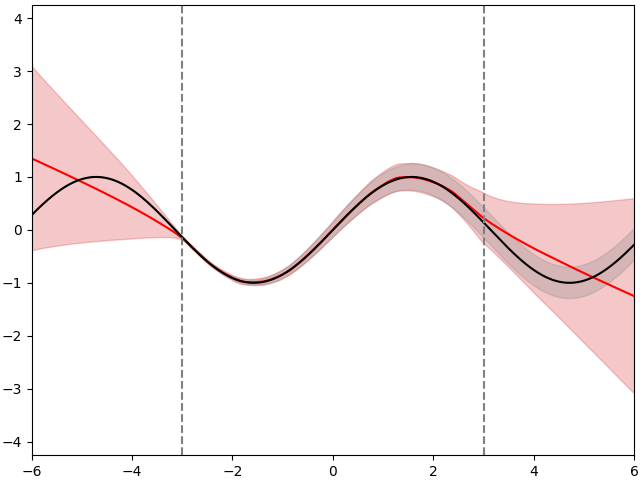}
     \end{subfigure}
     \begin{subfigure}[b]{0.195\textwidth}
         \centering
         \includegraphics[width=\textwidth]{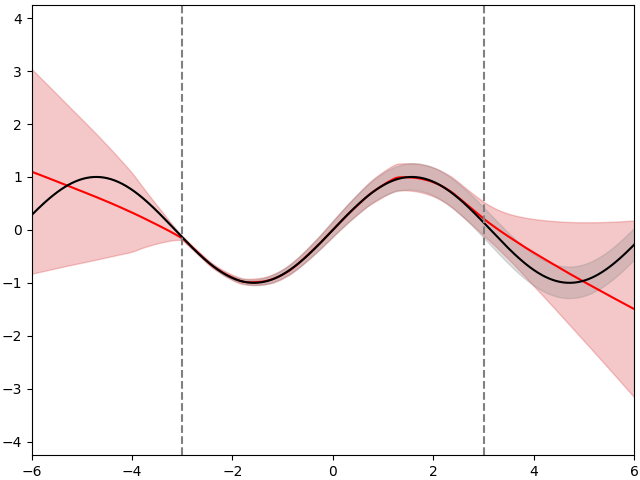}
     \end{subfigure}
     \begin{subfigure}[b]{0.195\textwidth}
         \centering
         \includegraphics[width=\textwidth]{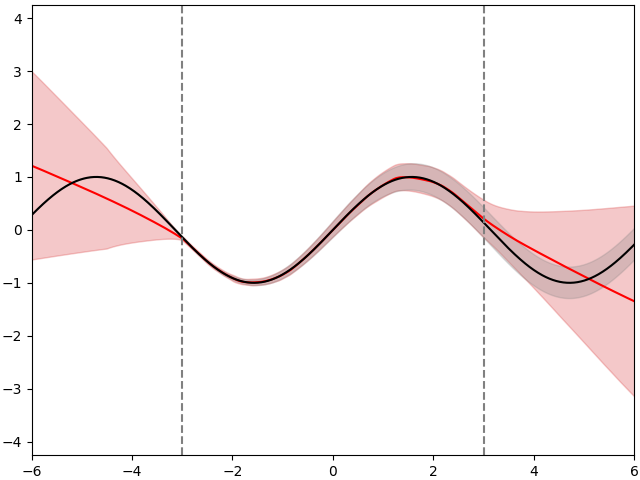}
     \end{subfigure}
     \begin{subfigure}[b]{0.195\textwidth}
         \centering
         \includegraphics[width=\textwidth]{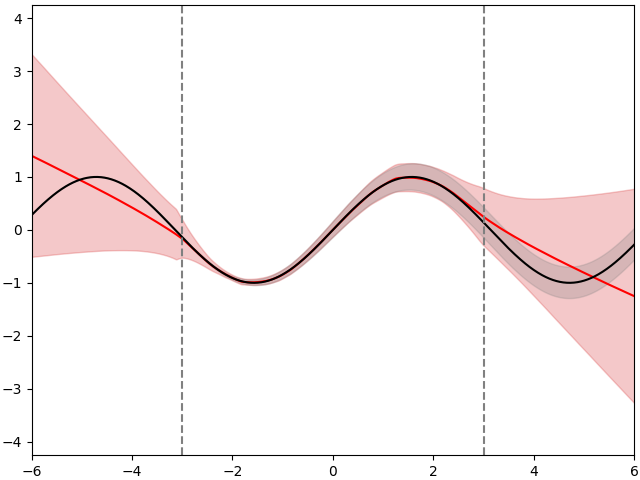}
     \end{subfigure}
        \caption{Toy regression problem, ensembling, $M=64$. Examples of predictive distributions.} 
        \label{fig:toy_regression_problem_ensembling_m64}
\end{figure}

\begin{figure}
     \centering
     \begin{subfigure}[b]{0.195\textwidth}
         \centering
         \includegraphics[width=\textwidth]{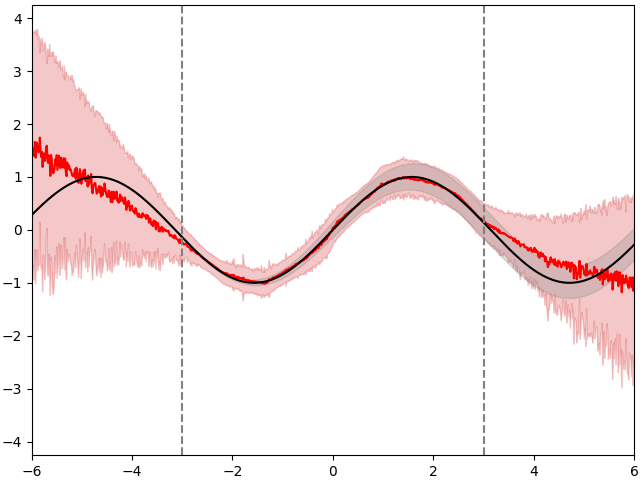}
     \end{subfigure}
     \begin{subfigure}[b]{0.195\textwidth}
         \centering
         \includegraphics[width=\textwidth]{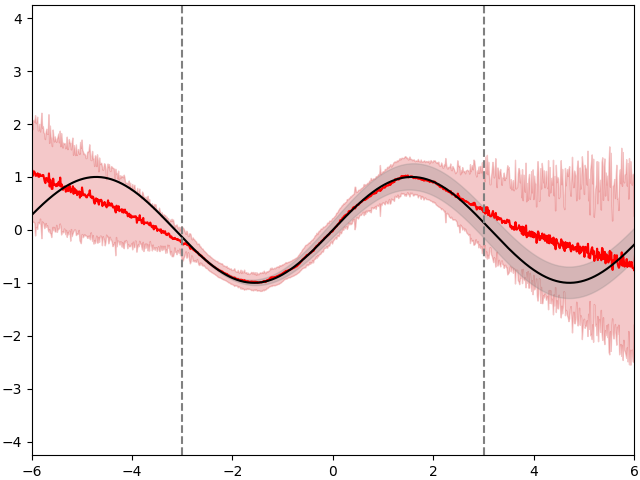}
     \end{subfigure}
     \begin{subfigure}[b]{0.195\textwidth}
         \centering
         \includegraphics[width=\textwidth]{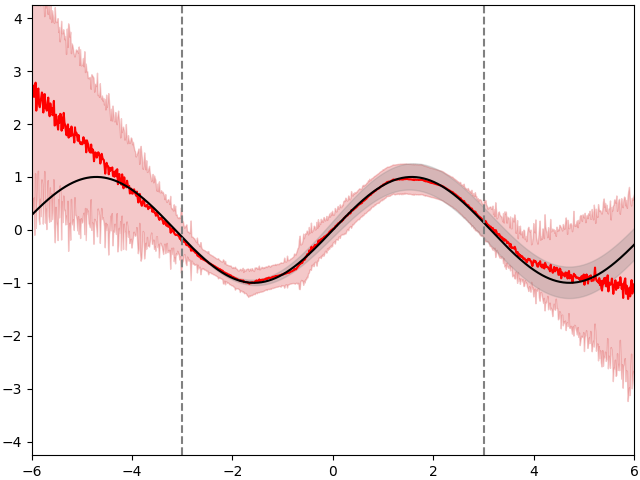}
     \end{subfigure}
     \begin{subfigure}[b]{0.195\textwidth}
         \centering
         \includegraphics[width=\textwidth]{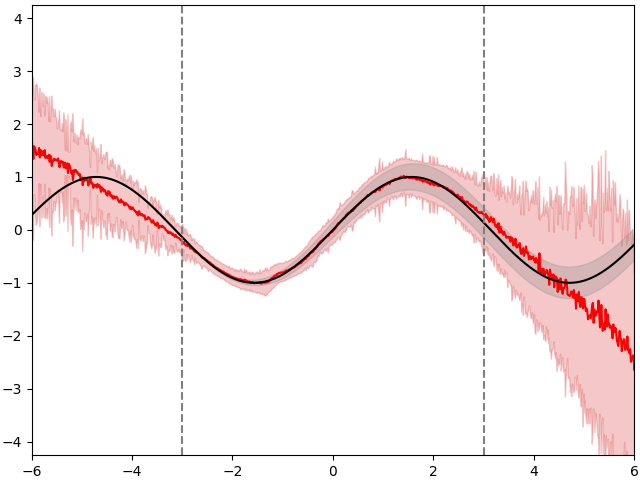}
     \end{subfigure}
     \begin{subfigure}[b]{0.195\textwidth}
         \centering
         \includegraphics[width=\textwidth]{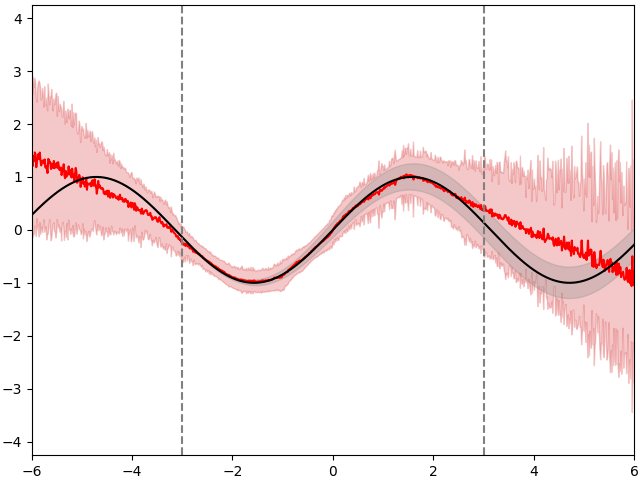}
     \end{subfigure}
        \caption{Toy regression problem, MC-dropout, $M=64$. Examples of predictive distributions.} 
        \label{fig:toy_regression_problem_mcdropout_m64}
\end{figure}

\begin{figure}
     \centering
     \begin{subfigure}[b]{0.195\textwidth}
         \centering
         \includegraphics[width=\textwidth]{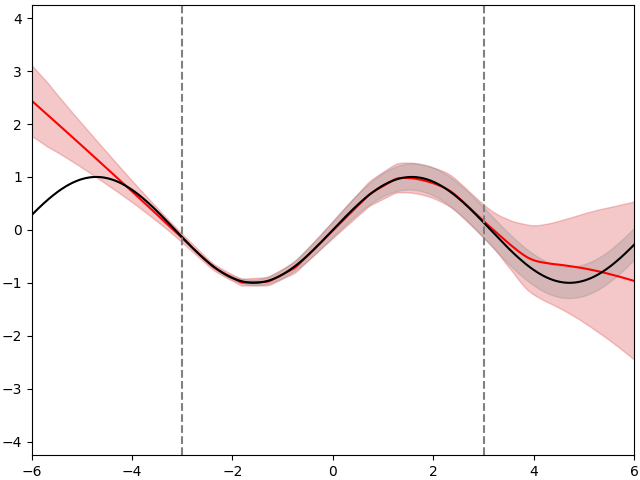}
     \end{subfigure}
     \begin{subfigure}[b]{0.195\textwidth}
         \centering
         \includegraphics[width=\textwidth]{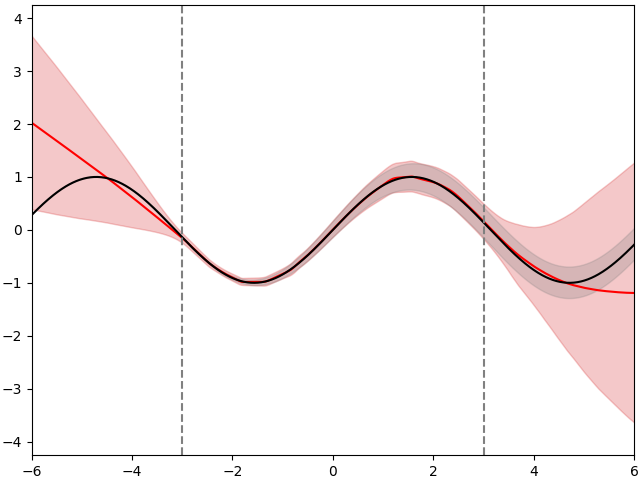}
     \end{subfigure}
     \begin{subfigure}[b]{0.195\textwidth}
         \centering
         \includegraphics[width=\textwidth]{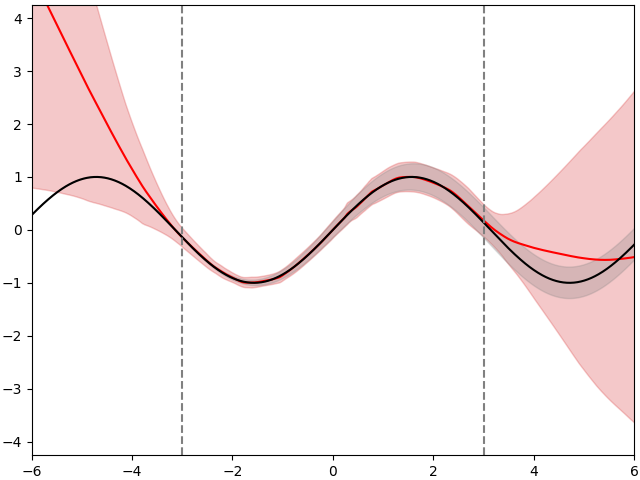}
     \end{subfigure}
     \begin{subfigure}[b]{0.195\textwidth}
         \centering
         \includegraphics[width=\textwidth]{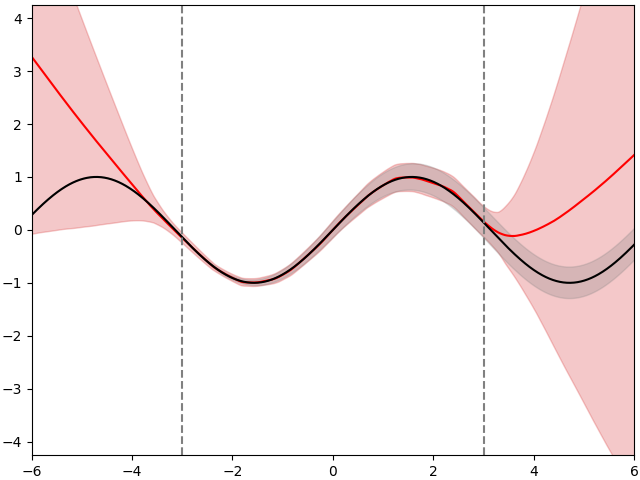}
     \end{subfigure}
     \begin{subfigure}[b]{0.195\textwidth}
         \centering
         \includegraphics[width=\textwidth]{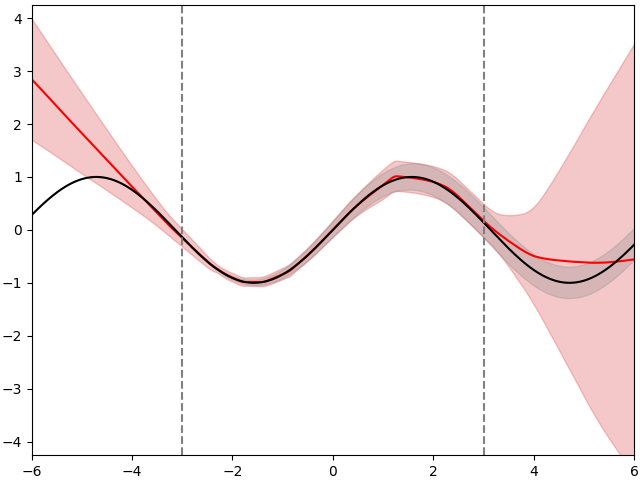}
     \end{subfigure}
        \caption{Toy regression problem, SGLD, $M=64$. Examples of predictive distributions.} 
        \label{fig:toy_regression_problem_sgld_m64}
\end{figure}

\begin{figure}
     \centering
     \begin{subfigure}[b]{0.195\textwidth}
         \centering
         \includegraphics[width=\textwidth]{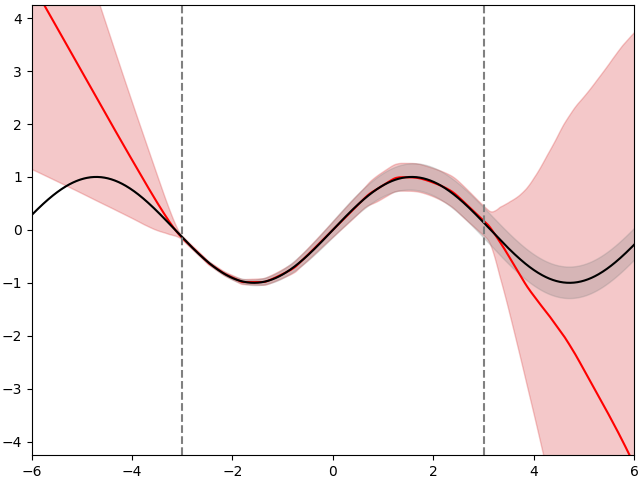}
     \end{subfigure}
     \begin{subfigure}[b]{0.195\textwidth}
         \centering
         \includegraphics[width=\textwidth]{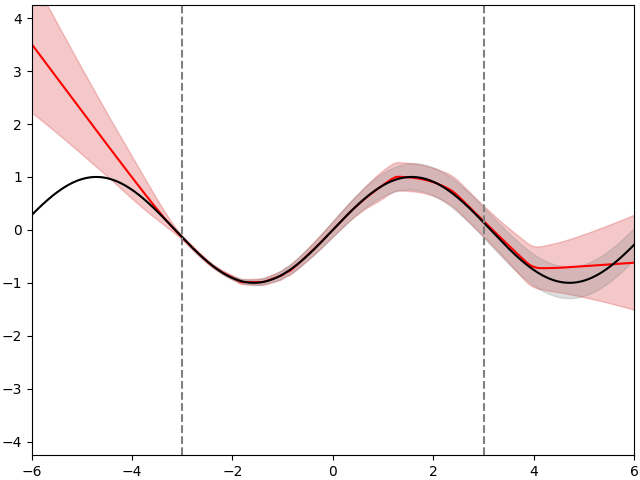}
     \end{subfigure}
     \begin{subfigure}[b]{0.195\textwidth}
         \centering
         \includegraphics[width=\textwidth]{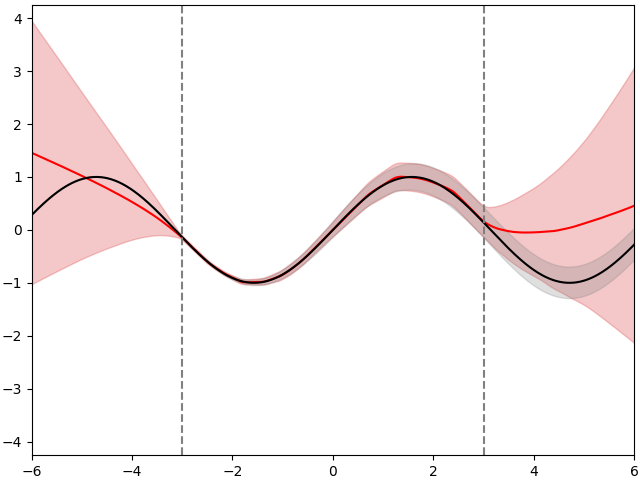}
     \end{subfigure}
     \begin{subfigure}[b]{0.195\textwidth}
         \centering
         \includegraphics[width=\textwidth]{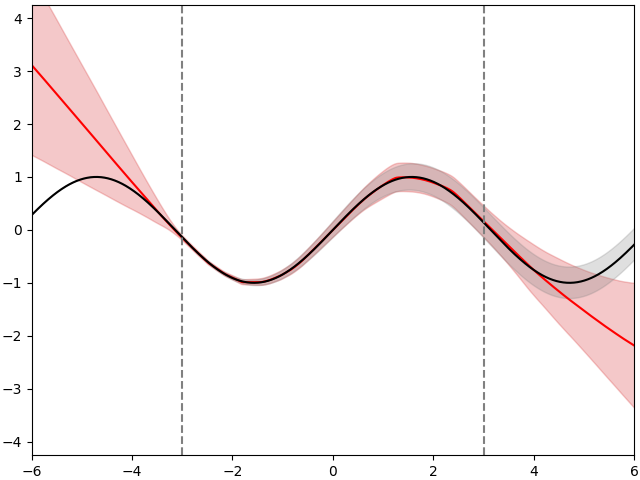}
     \end{subfigure}
     \begin{subfigure}[b]{0.195\textwidth}
         \centering
         \includegraphics[width=\textwidth]{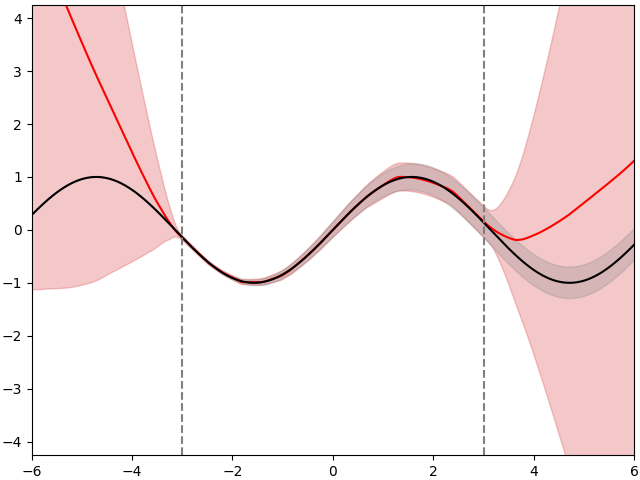}
     \end{subfigure}
        \caption{Toy regression problem, SGHMC, $M=64$. Examples of predictive distributions.} 
        \label{fig:toy_regression_problem_sghmc_m64}
\end{figure}

\begin{figure}
     \centering
     \begin{subfigure}[b]{0.195\textwidth}
         \centering
         \includegraphics[width=\textwidth]{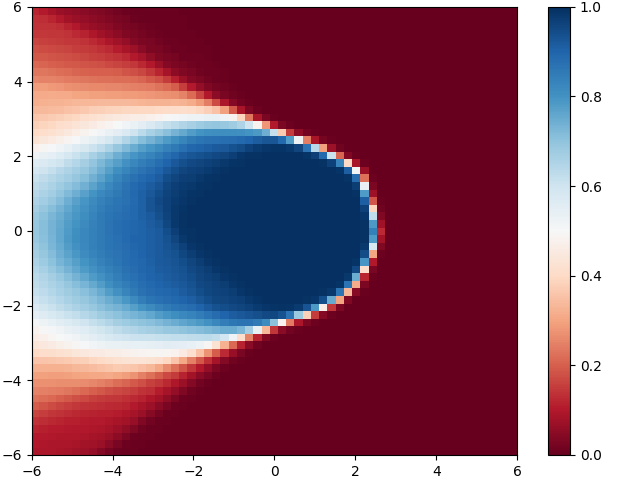}
     \end{subfigure}
     \begin{subfigure}[b]{0.195\textwidth}
         \centering
         \includegraphics[width=\textwidth]{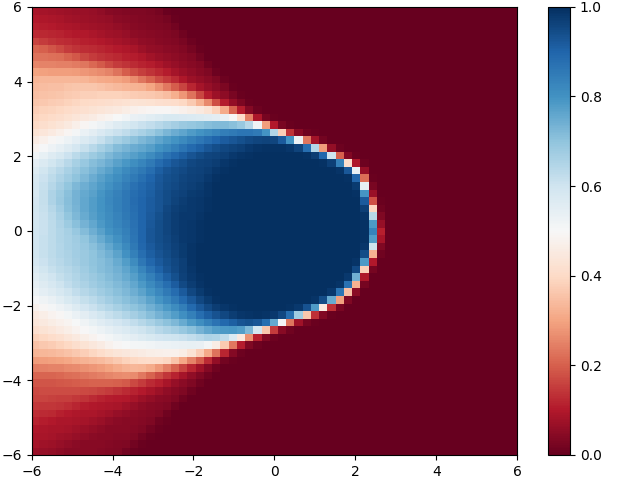}
     \end{subfigure}
     \begin{subfigure}[b]{0.195\textwidth}
         \centering
         \includegraphics[width=\textwidth]{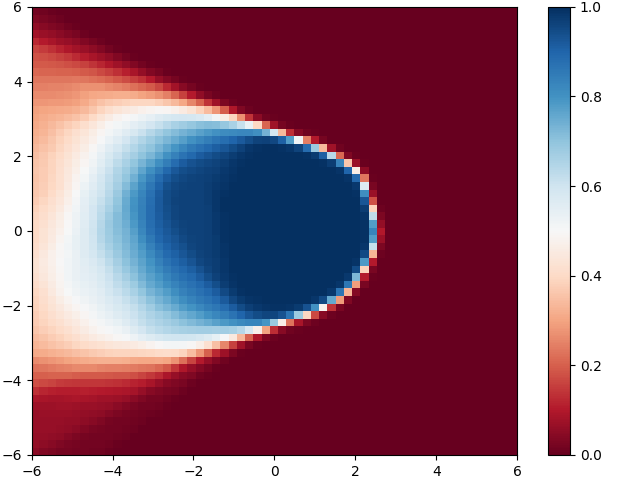}
     \end{subfigure}
     \begin{subfigure}[b]{0.195\textwidth}
         \centering
         \includegraphics[width=\textwidth]{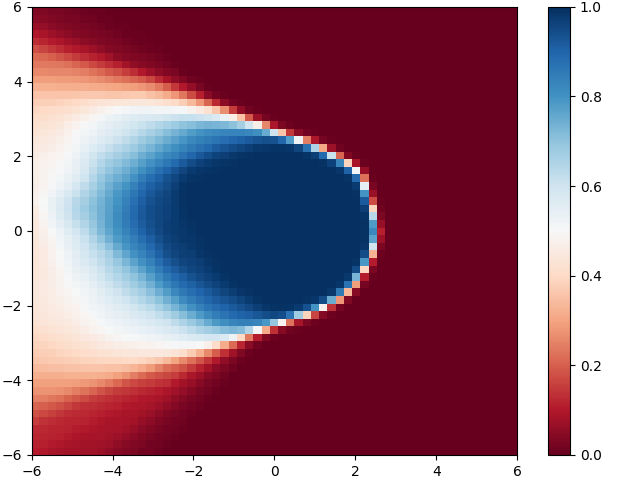}
     \end{subfigure}
     \begin{subfigure}[b]{0.195\textwidth}
         \centering
         \includegraphics[width=\textwidth]{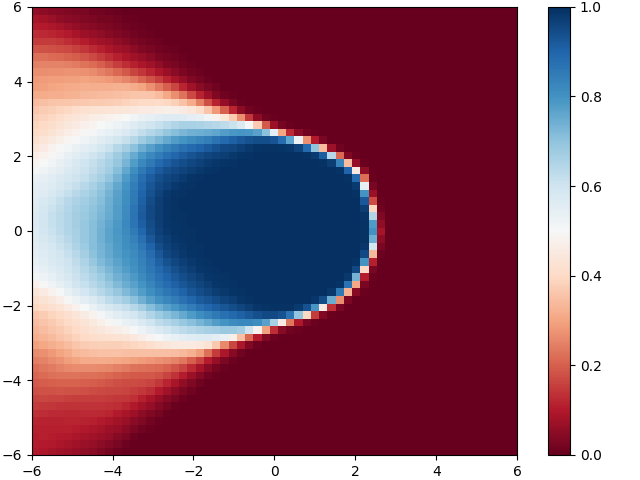}
     \end{subfigure}
        \caption{Toy classification problem, ensembling, $M=64$. Examples of predictive distributions.} 
        \label{fig:toy_classification_problem_ensembling_m64}
\end{figure}

\begin{figure}
     \centering
     \begin{subfigure}[b]{0.195\textwidth}
         \centering
         \includegraphics[width=\textwidth]{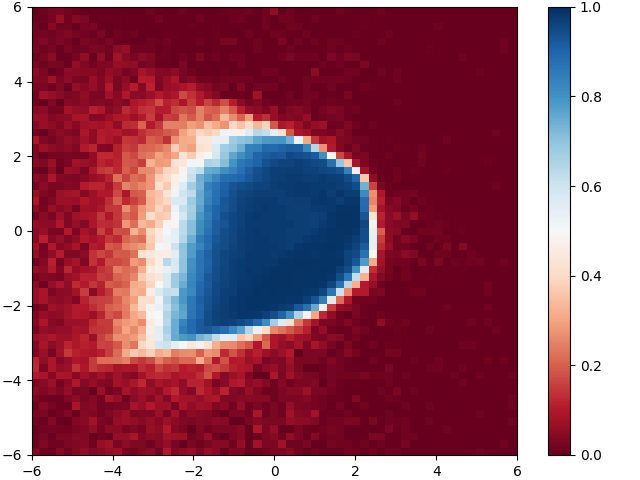}
     \end{subfigure}
     \begin{subfigure}[b]{0.195\textwidth}
         \centering
         \includegraphics[width=\textwidth]{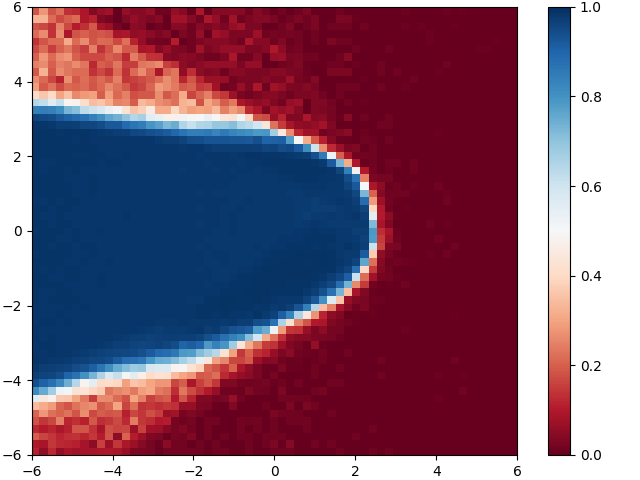}
     \end{subfigure}
     \begin{subfigure}[b]{0.195\textwidth}
         \centering
         \includegraphics[width=\textwidth]{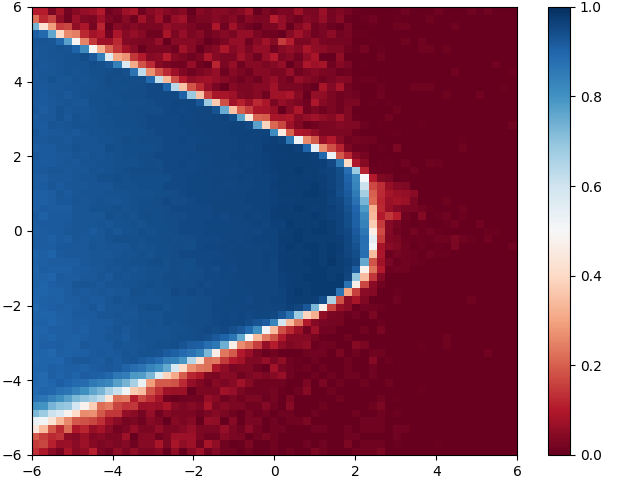}
     \end{subfigure}
     \begin{subfigure}[b]{0.195\textwidth}
         \centering
         \includegraphics[width=\textwidth]{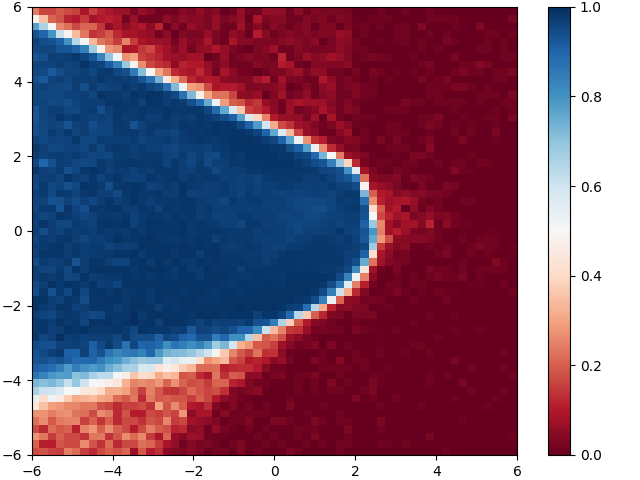}
     \end{subfigure}
     \begin{subfigure}[b]{0.195\textwidth}
         \centering
         \includegraphics[width=\textwidth]{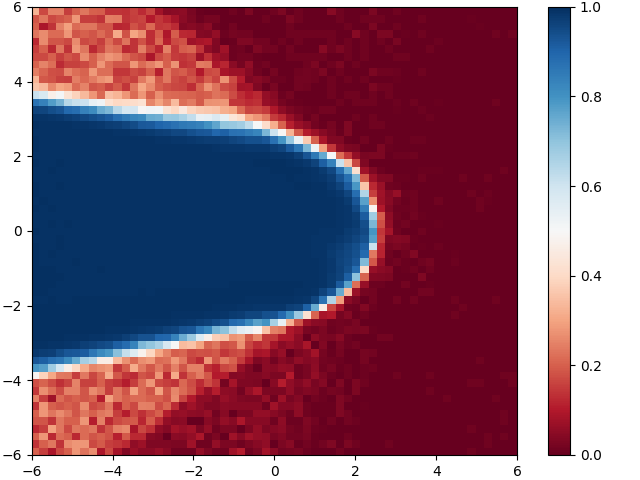}
     \end{subfigure}
        \caption{Toy classification problem, MC-dropout, $M=64$. Examples of predictive distributions.}
        \label{fig:toy_classification_problem_mcdropout_m64}
\end{figure}

\begin{figure}
     \centering
     \begin{subfigure}[b]{0.195\textwidth}
         \centering
         \includegraphics[width=\textwidth]{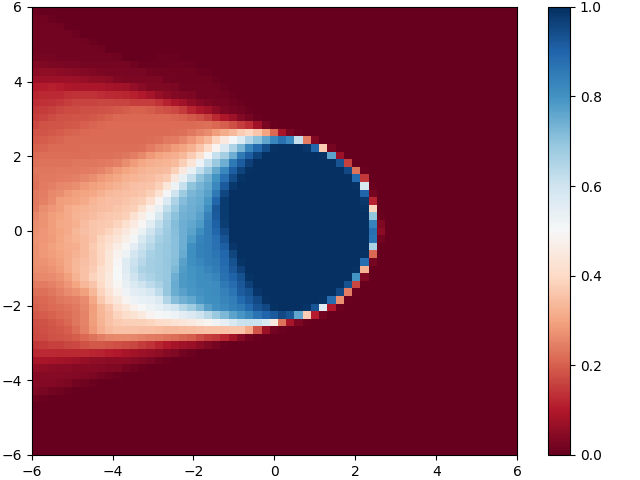}
     \end{subfigure}
     \begin{subfigure}[b]{0.195\textwidth}
         \centering
         \includegraphics[width=\textwidth]{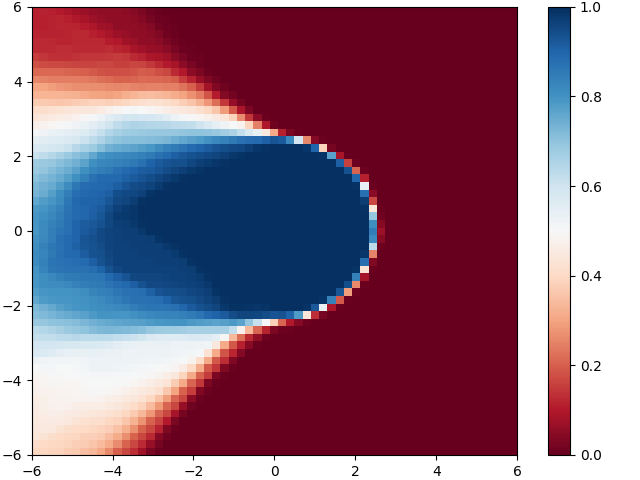}
     \end{subfigure}
     \begin{subfigure}[b]{0.195\textwidth}
         \centering
         \includegraphics[width=\textwidth]{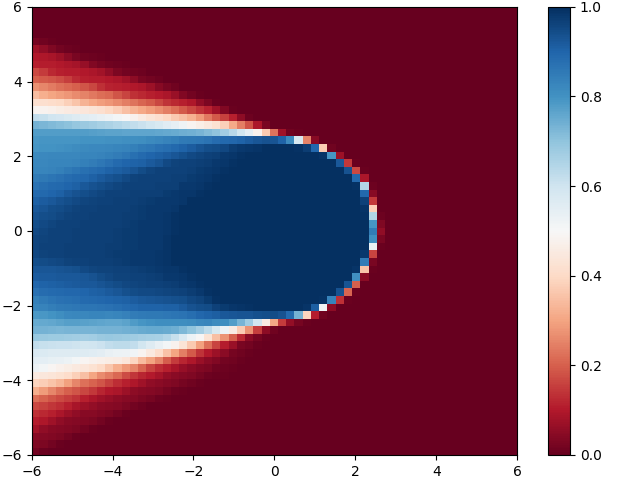}
     \end{subfigure}
     \begin{subfigure}[b]{0.195\textwidth}
         \centering
         \includegraphics[width=\textwidth]{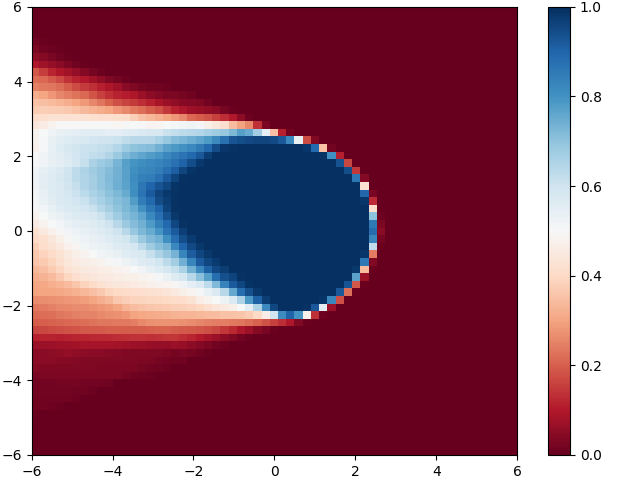}
     \end{subfigure}
     \begin{subfigure}[b]{0.195\textwidth}
         \centering
         \includegraphics[width=\textwidth]{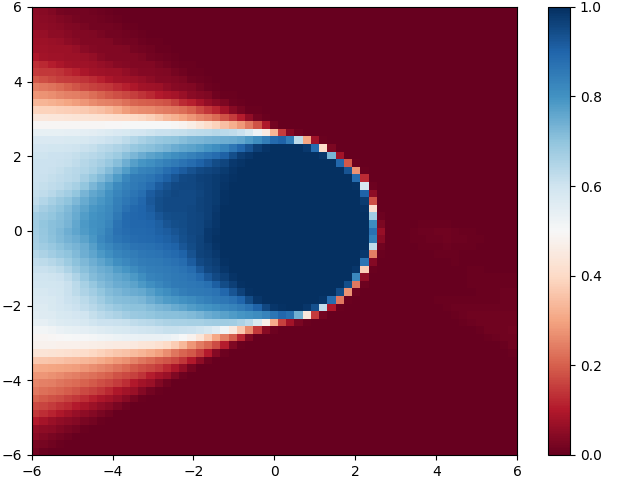}
     \end{subfigure}
        \caption{Toy classification problem, SGLD, $M=64$. Examples of predictive distributions.}
        \label{fig:toy_classification_problem_sgld_m64}
\end{figure}

\begin{figure}
     \centering
     \begin{subfigure}[b]{0.195\textwidth}
         \centering
         \includegraphics[width=\textwidth]{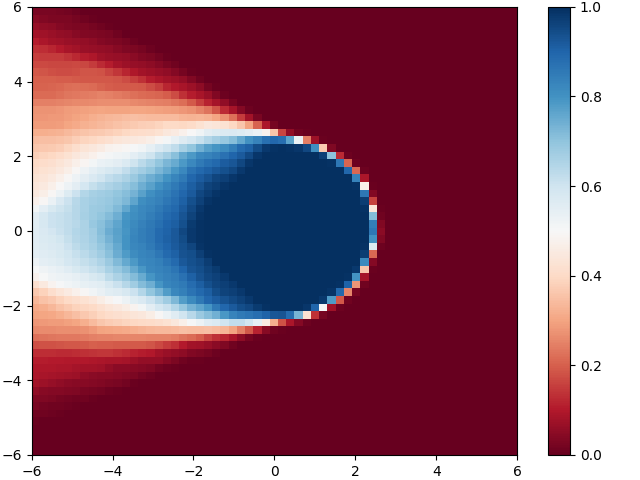}
     \end{subfigure}
     \begin{subfigure}[b]{0.195\textwidth}
         \centering
         \includegraphics[width=\textwidth]{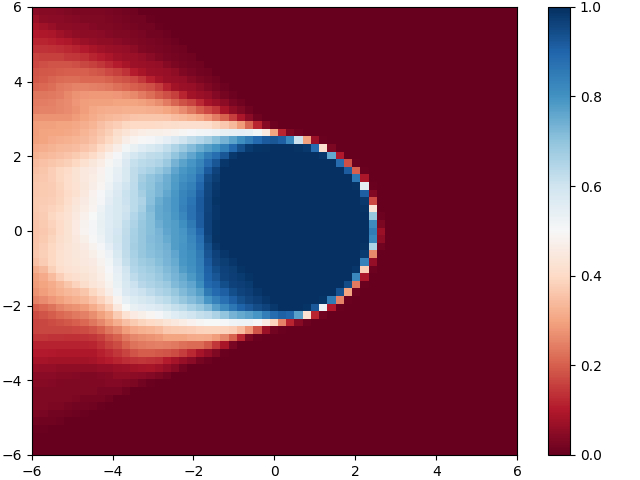}
     \end{subfigure}
     \begin{subfigure}[b]{0.195\textwidth}
         \centering
         \includegraphics[width=\textwidth]{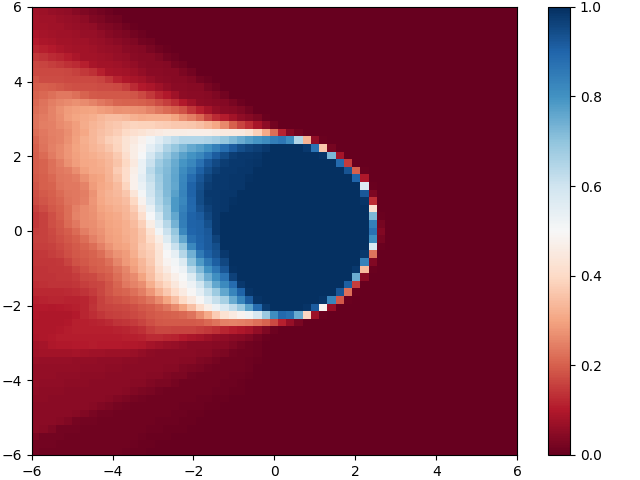}
     \end{subfigure}
     \begin{subfigure}[b]{0.195\textwidth}
         \centering
         \includegraphics[width=\textwidth]{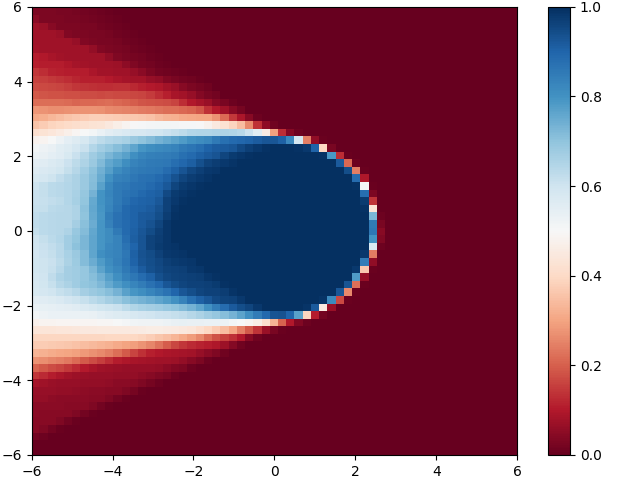}
     \end{subfigure}
     \begin{subfigure}[b]{0.195\textwidth}
         \centering
         \includegraphics[width=\textwidth]{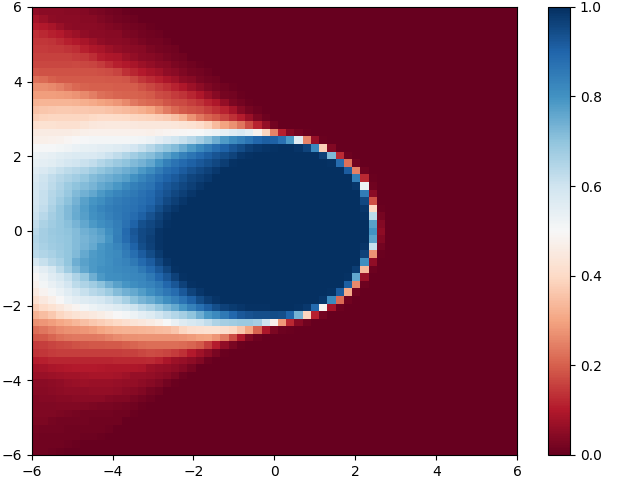}
     \end{subfigure}
        \caption{Toy classification problem, SGHMC, $M=64$. Examples of predictive distributions.}
        \label{fig:toy_classification_problem_sghmc_m64}
\end{figure}

Here, we show visualizations of predictive distributions obtained by the different methods. Figure~\ref{fig:toy_regression_problem_ensembling_m64}, \ref{fig:toy_classification_problem_ensembling_m64} for ensembling, Figure~\ref{fig:toy_regression_problem_mcdropout_m64}, \ref{fig:toy_classification_problem_mcdropout_m64} for MC-dropout, Figure~\ref{fig:toy_regression_problem_sgld_m64}, \ref{fig:toy_classification_problem_sgld_m64} for SGLD, and Figure~\ref{fig:toy_regression_problem_sghmc_m64}, \ref{fig:toy_classification_problem_sghmc_m64} for SGHMC.

\clearpage
\section{Depth Completion}
\label{appendix:depth_completion}

In this appendix, further details on the depth completion experiments (Section~\ref{experiments-depth_completion}) are provided.

\subsection{Training Details}
\label{appendix:depth_completion_training_details}

For both ensembling and MC-dropout, we train all models for $40\thinspace000$ steps with the Adam optimizer, a batch size of $4$, a fixed learning rate of $10^{-5}$ and weight decay of $0.0005$. We use a smaller batch size and train for fewer steps than Ma et al.~\cite{ma2018self} to enable an extensive evaluation with repeated experiments. For the same reason, we also train on randomly selected image crops of size $352 \times 352$. The only other data augmentation used is random flipping along the vertical axis. We follow Ma et al.~\cite{ma2018self} and randomly initialize all network weights from $\mathcal{N}(0, 10^{-3})$ and all network biases with $0$s. Models are trained on a single NVIDIA TITAN Xp GPU with $12$GB of RAM.

\subsection{Description of Results}
\label{appendix:depth_completion_results_description}

The results in Figure~\ref{fig:depth_completion_ause_auce} (Section~\ref{experiments-depth_completion}) were obtained in the following way:
\begin{itemize}
    \item \textbf{Ensembling:} $33$ models were trained using the same training procedure, the mean and standard deviation was computed based on $32$ ($M = 1$), $16$ ($M = 2, 4, 8, 16$) or $4$ ($M = 32$) sets of randomly drawn models. The same set could not be drawn more than once.  
    
    \item \textbf{MC-dropout:} $16$ models were trained using the same training procedure, based on which the mean and standard deviation was computed. 
\end{itemize}

\subsection{Additional Results}
\label{appendix:depth_completion_additional_results}

\begin{figure}
     \centering
     \begin{subfigure}[b]{0.49\textwidth}
         \centering
         \includegraphics[width=\textwidth]{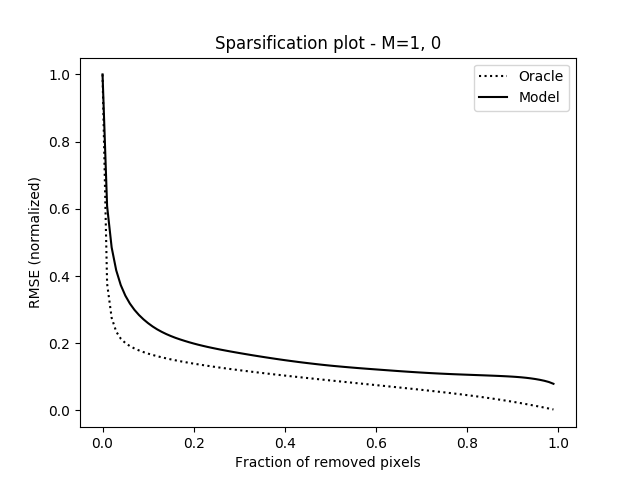}
         \caption{$M = 1$.}
     \end{subfigure}
     \begin{subfigure}[b]{0.49\textwidth}
         \centering
         \includegraphics[width=\textwidth]{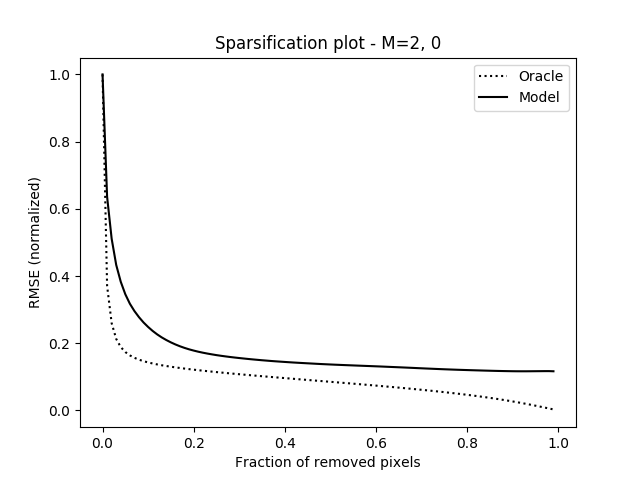}
         \caption{$M = 2$.}
     \end{subfigure}
     \begin{subfigure}[b]{0.49\textwidth}
         \centering
         \includegraphics[width=\textwidth]{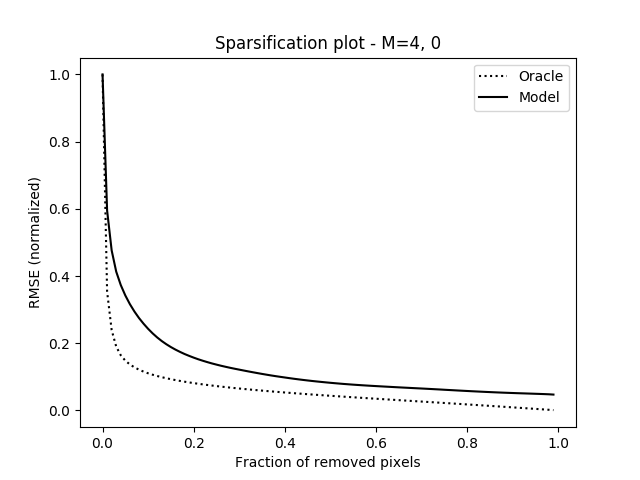}
         \caption{$M = 4$.}
     \end{subfigure}
     \begin{subfigure}[b]{0.49\textwidth}
         \centering
         \includegraphics[width=\textwidth]{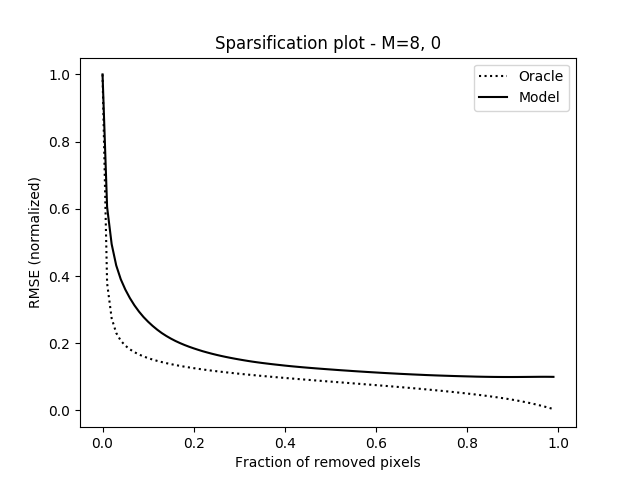}
         \caption{$M = 8$.}
     \end{subfigure}
     \begin{subfigure}[b]{0.49\textwidth}
         \centering
         \includegraphics[width=\textwidth]{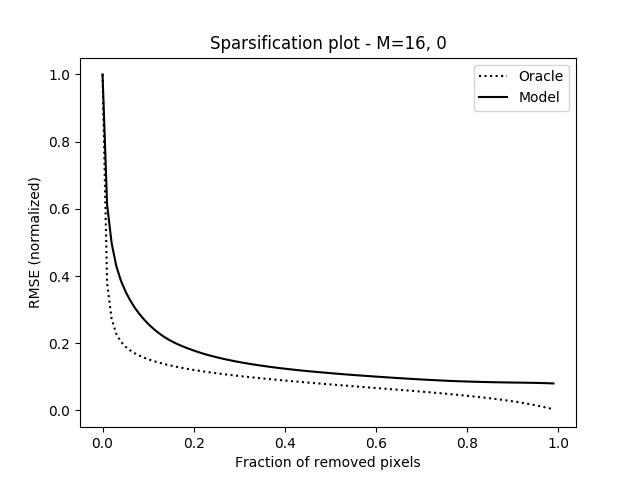}
         \caption{$M = 16$.}
     \end{subfigure}
     \begin{subfigure}[b]{0.49\textwidth}
         \centering
         \includegraphics[width=\textwidth]{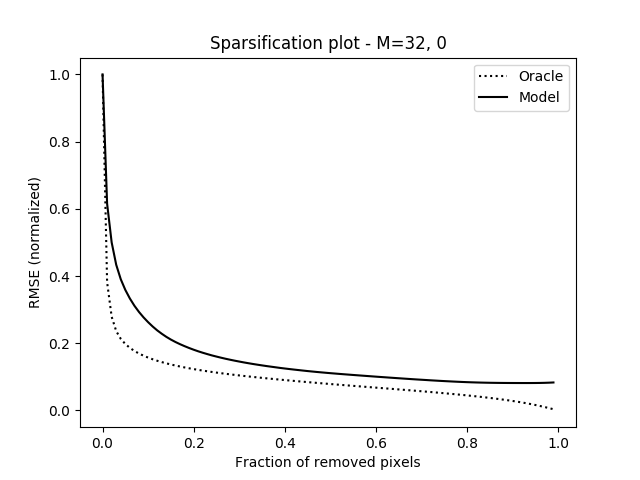}
         \caption{$M = 32$.}
     \end{subfigure}
        \caption{Results for ensembling on the KITTI depth completion validation dataset. Examples of sparsification plots.}
        \label{fig:depth_completion_sparsification_plots_ensembling}
\end{figure}

\begin{figure}
     \centering
     \begin{subfigure}[b]{0.49\textwidth}
         \centering
         \includegraphics[width=\textwidth]{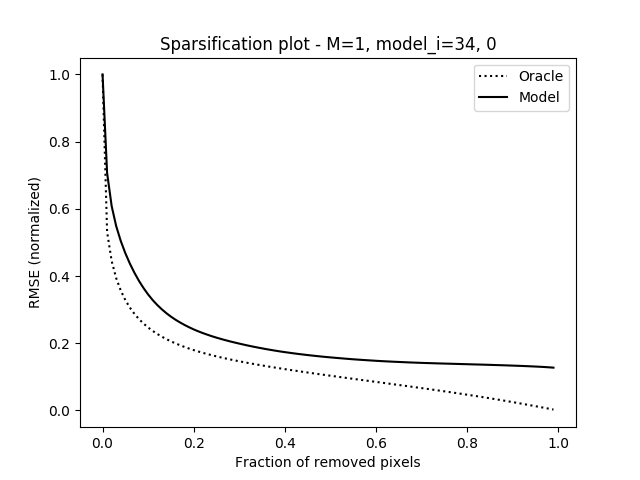}
         \caption{$M = 1$.}
     \end{subfigure}
     \begin{subfigure}[b]{0.49\textwidth}
         \centering
         \includegraphics[width=\textwidth]{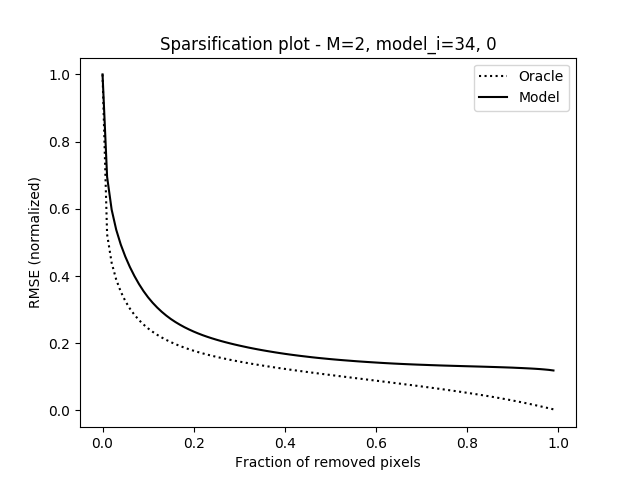}
         \caption{$M = 2$.}
     \end{subfigure}
     \begin{subfigure}[b]{0.49\textwidth}
         \centering
         \includegraphics[width=\textwidth]{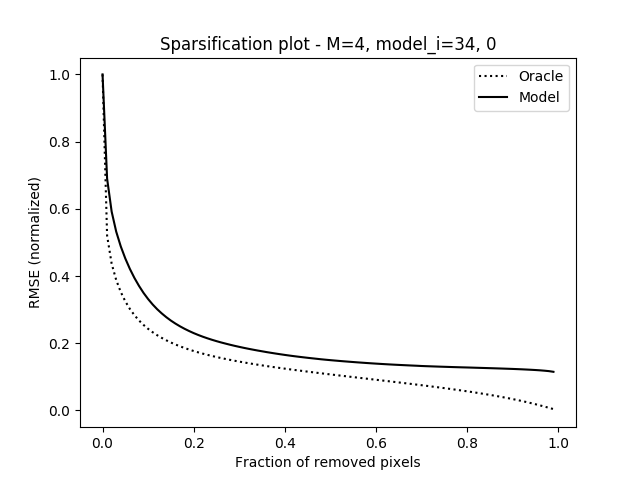}
         \caption{$M = 4$.}
     \end{subfigure}
     \begin{subfigure}[b]{0.49\textwidth}
         \centering
         \includegraphics[width=\textwidth]{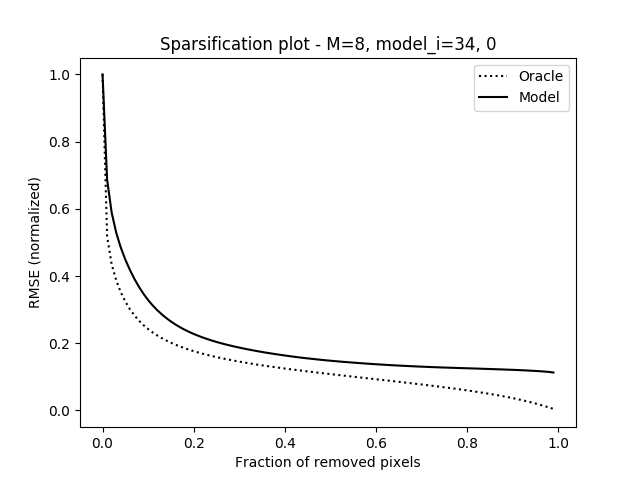}
         \caption{$M = 8$.}
     \end{subfigure}
     \begin{subfigure}[b]{0.49\textwidth}
         \centering
         \includegraphics[width=\textwidth]{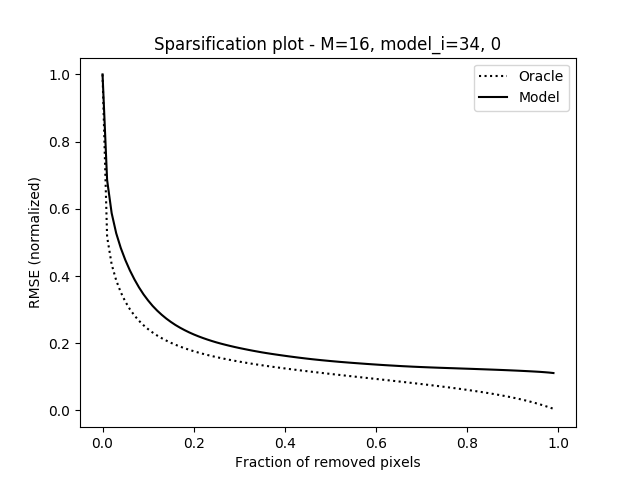}
         \caption{$M = 16$.}
     \end{subfigure}
     \begin{subfigure}[b]{0.49\textwidth}
         \centering
         \includegraphics[width=\textwidth]{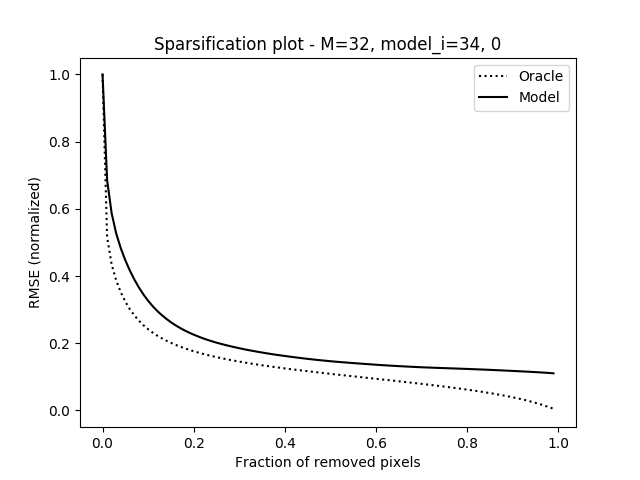}
         \caption{$M = 32$.}
     \end{subfigure}
        \caption{Results for MC-dropout on the KITTI depth completion validation dataset. Examples of sparsification plots.}
        \label{fig:depth_completion_sparsification_plots_mcdropout}
\end{figure}

\begin{figure}
     \centering
     \begin{subfigure}[b]{0.49\textwidth}
         \centering
         \includegraphics[width=\textwidth]{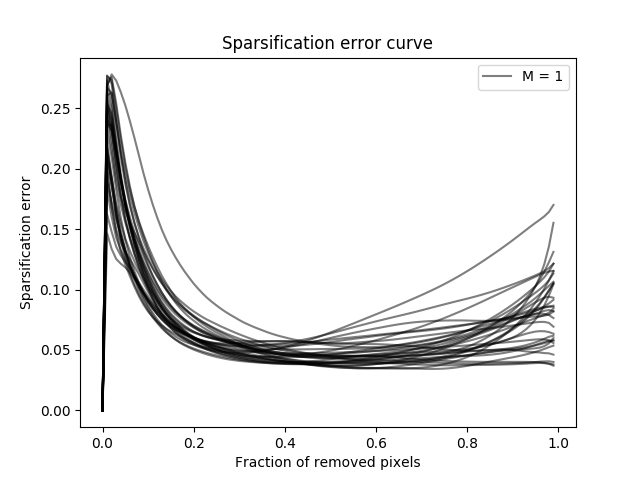}
         \caption{$M = 1$.}
     \end{subfigure}
     \begin{subfigure}[b]{0.49\textwidth}
         \centering
         \includegraphics[width=\textwidth]{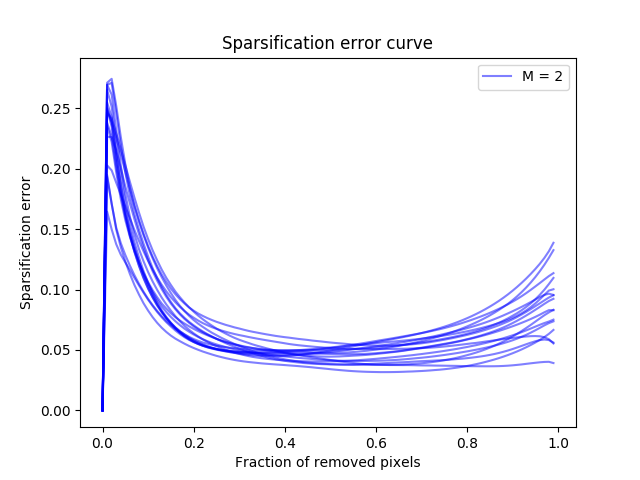}
         \caption{$M = 2$.}
     \end{subfigure}
     \begin{subfigure}[b]{0.49\textwidth}
         \centering
         \includegraphics[width=\textwidth]{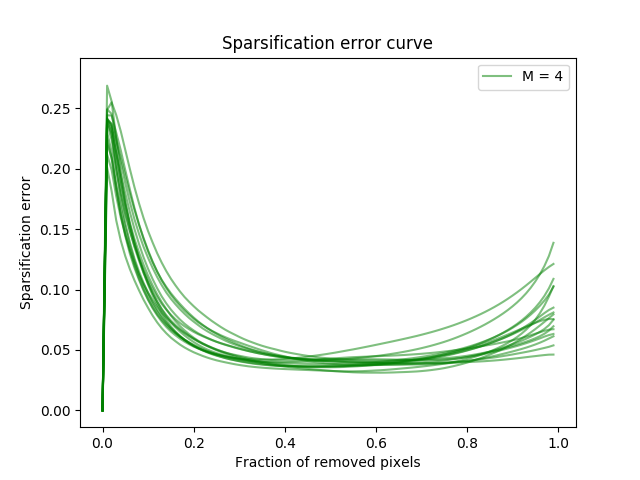}
         \caption{$M = 4$.}
     \end{subfigure}
     \begin{subfigure}[b]{0.49\textwidth}
         \centering
         \includegraphics[width=\textwidth]{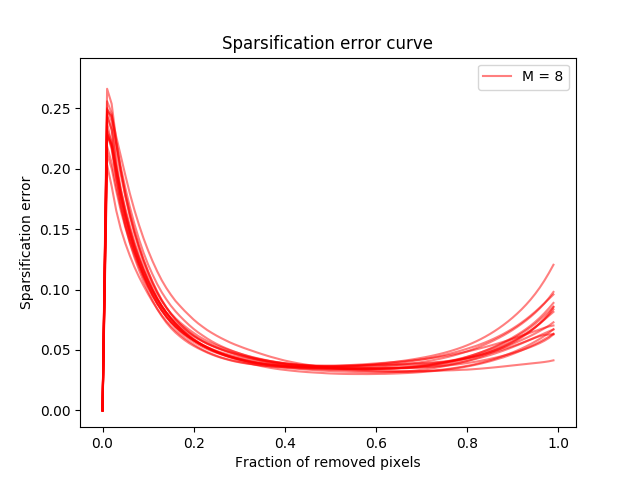}
         \caption{$M = 8$.}
     \end{subfigure}
     \begin{subfigure}[b]{0.49\textwidth}
         \centering
         \includegraphics[width=\textwidth]{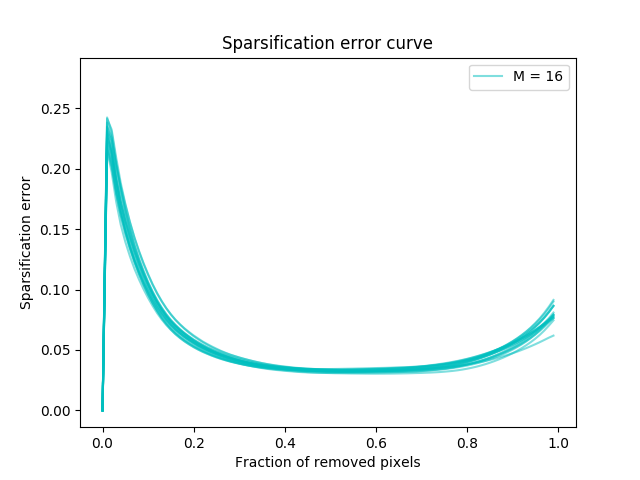}
         \caption{$M = 16$.}
     \end{subfigure}
     \begin{subfigure}[b]{0.49\textwidth}
         \centering
         \includegraphics[width=\textwidth]{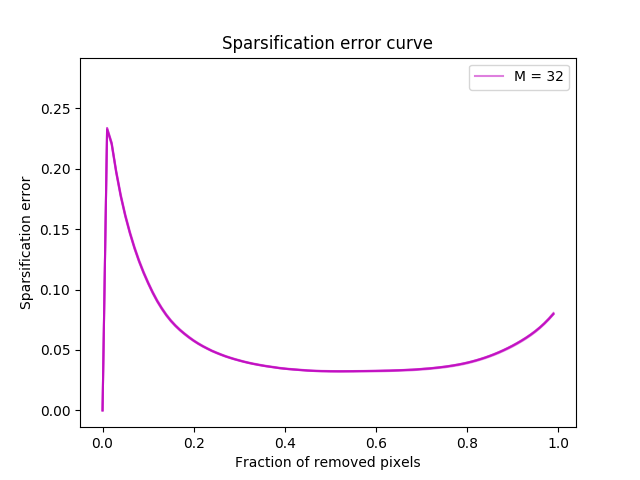}
         \caption{$M = 32$.}
     \end{subfigure}
        \caption{Results for ensembling on the KITTI depth completion validation dataset. Condensed sparsification error curves.}
        \label{fig:depth_completion_sparsification_error_curves_ensembling}
\end{figure}

\begin{figure}
     \centering
     \begin{subfigure}[b]{0.49\textwidth}
         \centering
         \includegraphics[width=\textwidth]{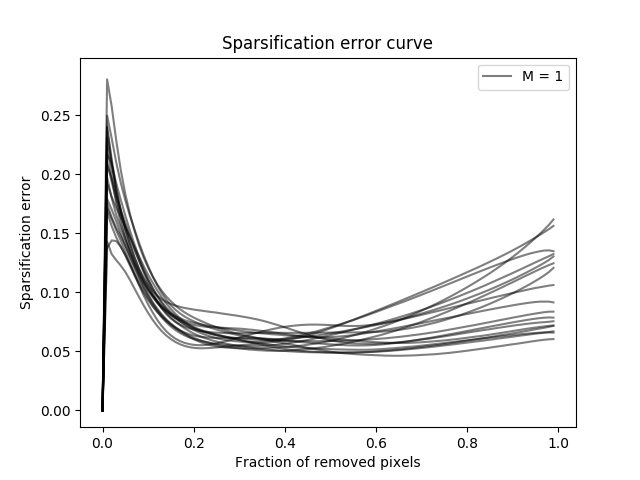}
         \caption{$M = 1$.}
     \end{subfigure}
     \begin{subfigure}[b]{0.49\textwidth}
         \centering
         \includegraphics[width=\textwidth]{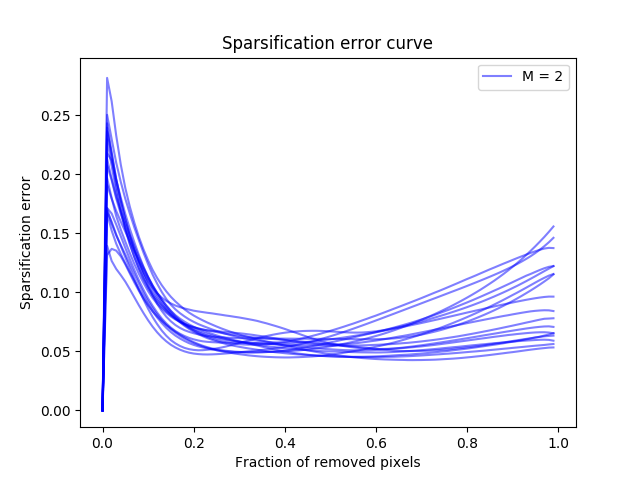}
         \caption{$M = 2$.}
     \end{subfigure}
     \begin{subfigure}[b]{0.49\textwidth}
         \centering
         \includegraphics[width=\textwidth]{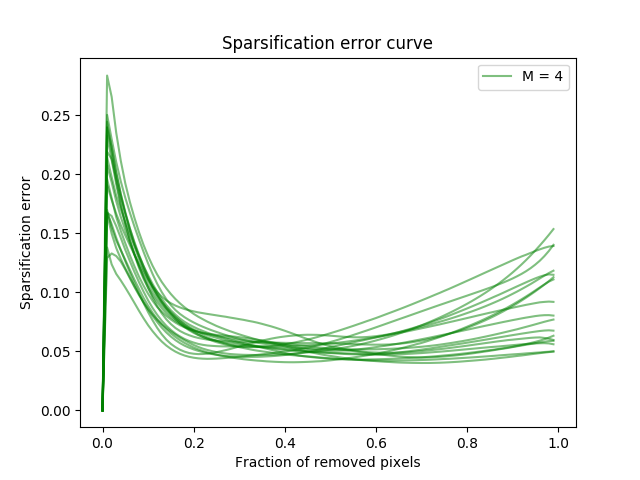}
         \caption{$M = 4$.}
     \end{subfigure}
     \begin{subfigure}[b]{0.49\textwidth}
         \centering
         \includegraphics[width=\textwidth]{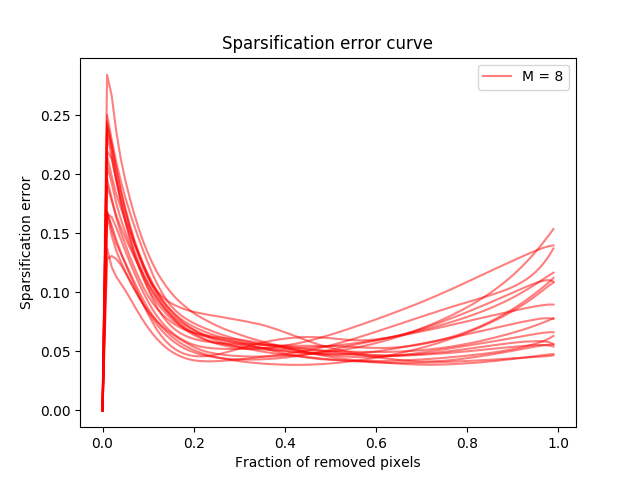}
         \caption{$M = 8$.}
     \end{subfigure}
     \begin{subfigure}[b]{0.49\textwidth}
         \centering
         \includegraphics[width=\textwidth]{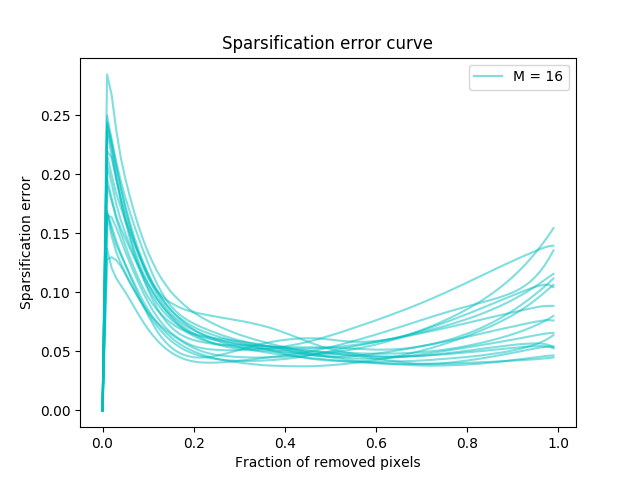}
         \caption{$M = 16$.}
     \end{subfigure}
     \begin{subfigure}[b]{0.49\textwidth}
         \centering
         \includegraphics[width=\textwidth]{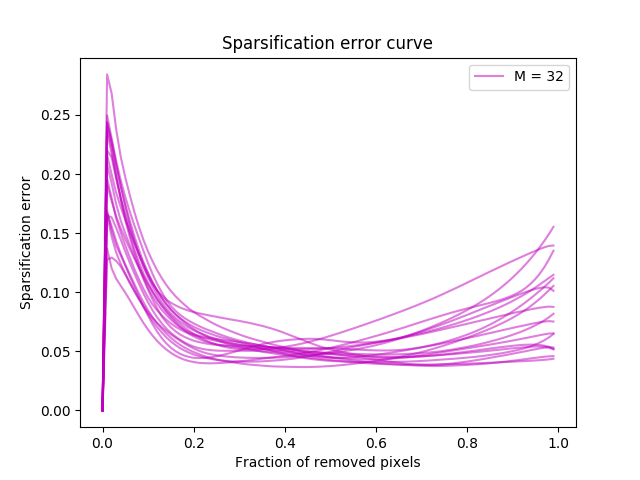}
         \caption{$M = 32$.}
     \end{subfigure}
        \caption{Results for MC-dropout on the KITTI depth completion validation dataset. Condensed sparsification error curves.}
        \label{fig:depth_completion_sparsification_error_curves_mcdropout}
\end{figure}

\begin{figure}
     \centering
     \begin{subfigure}[b]{0.49\textwidth}
         \centering
         \includegraphics[width=\textwidth]{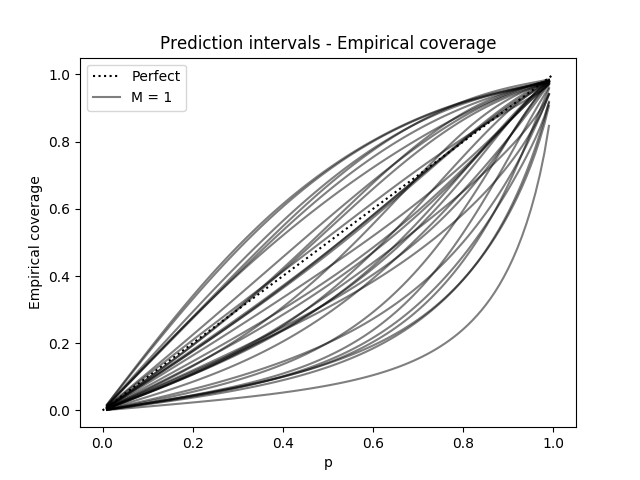}
         \caption{$M = 1$.}
     \end{subfigure}
     \begin{subfigure}[b]{0.49\textwidth}
         \centering
         \includegraphics[width=\textwidth]{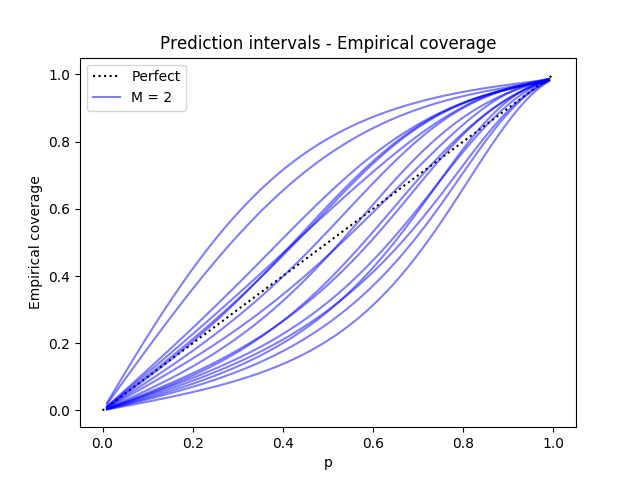}
         \caption{$M = 2$.}
     \end{subfigure}
     \begin{subfigure}[b]{0.49\textwidth}
         \centering
         \includegraphics[width=\textwidth]{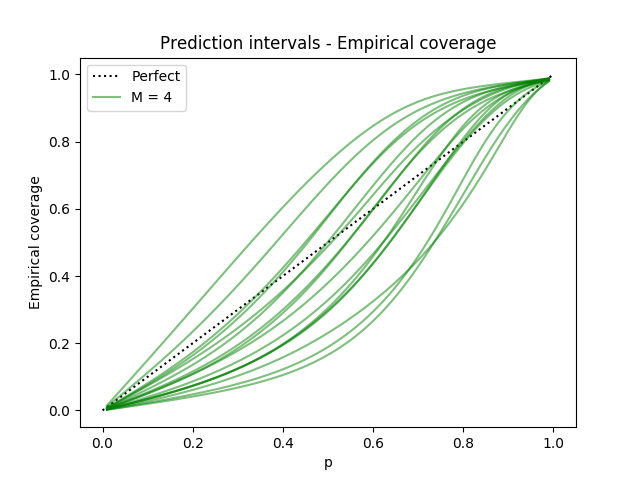}
         \caption{$M = 4$.}
     \end{subfigure}
     \begin{subfigure}[b]{0.49\textwidth}
         \centering
         \includegraphics[width=\textwidth]{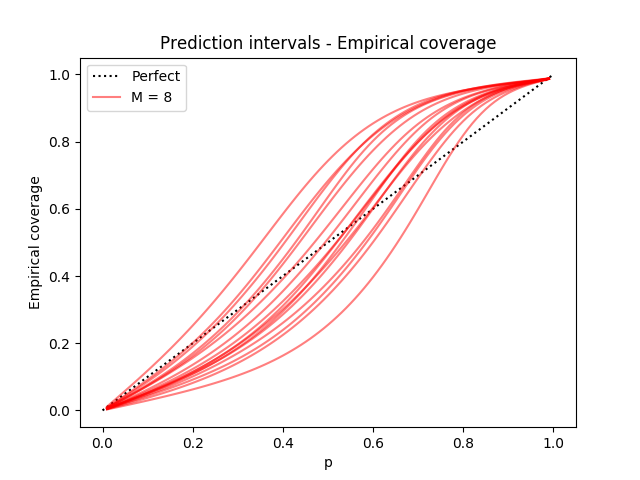}
         \caption{$M = 8$.}
     \end{subfigure}
     \begin{subfigure}[b]{0.49\textwidth}
         \centering
         \includegraphics[width=\textwidth]{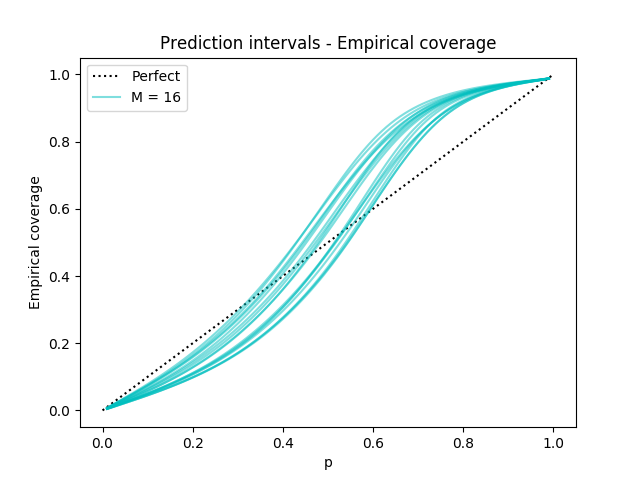}
         \caption{$M = 16$.}
     \end{subfigure}
     \begin{subfigure}[b]{0.49\textwidth}
         \centering
         \includegraphics[width=\textwidth]{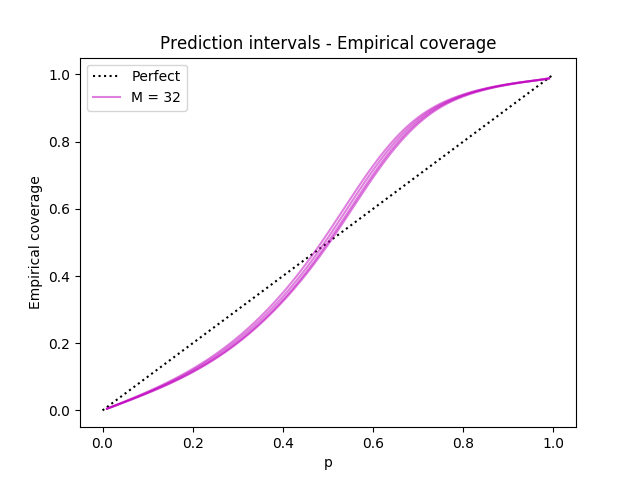}
         \caption{$M = 32$.}
     \end{subfigure}
        \caption{Results for ensembling on the KITTI depth completion validation dataset. Condensed calibration plots.}
        \label{fig:depth_completion_calibration_plots_ensembling}
\end{figure}

\begin{figure}
     \centering
     \begin{subfigure}[b]{0.49\textwidth}
         \centering
         \includegraphics[width=\textwidth]{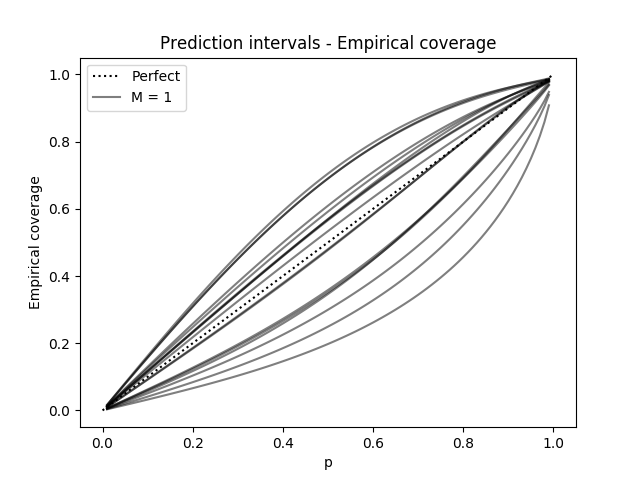}
         \caption{$M = 1$.}
     \end{subfigure}
     \begin{subfigure}[b]{0.49\textwidth}
         \centering
         \includegraphics[width=\textwidth]{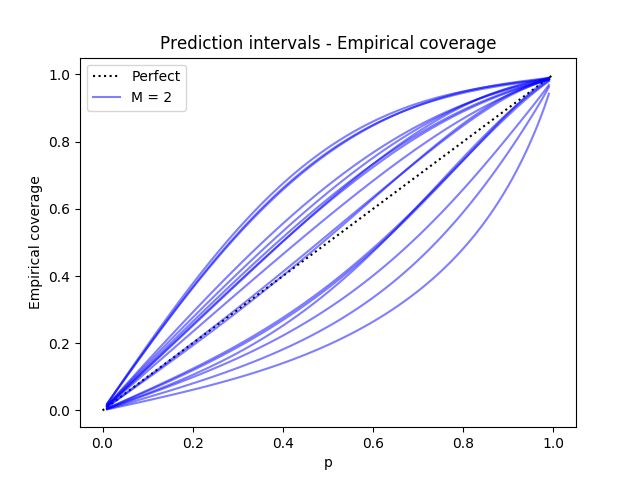}
         \caption{$M = 2$.}
     \end{subfigure}
     \begin{subfigure}[b]{0.49\textwidth}
         \centering
         \includegraphics[width=\textwidth]{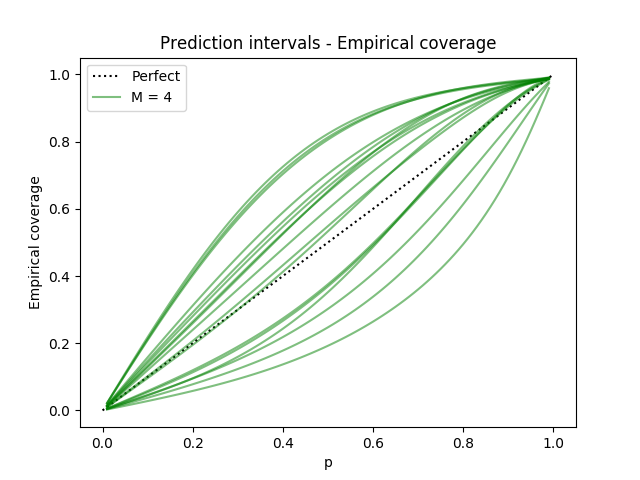}
         \caption{$M = 4$.}
     \end{subfigure}
     \begin{subfigure}[b]{0.49\textwidth}
         \centering
         \includegraphics[width=\textwidth]{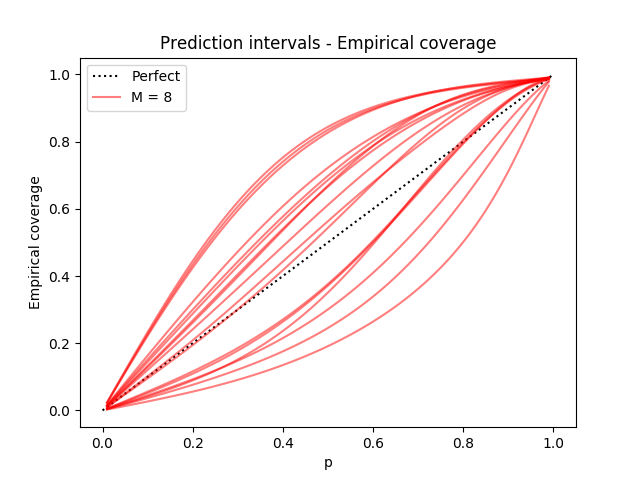}
         \caption{$M = 8$.}
     \end{subfigure}
     \begin{subfigure}[b]{0.49\textwidth}
         \centering
         \includegraphics[width=\textwidth]{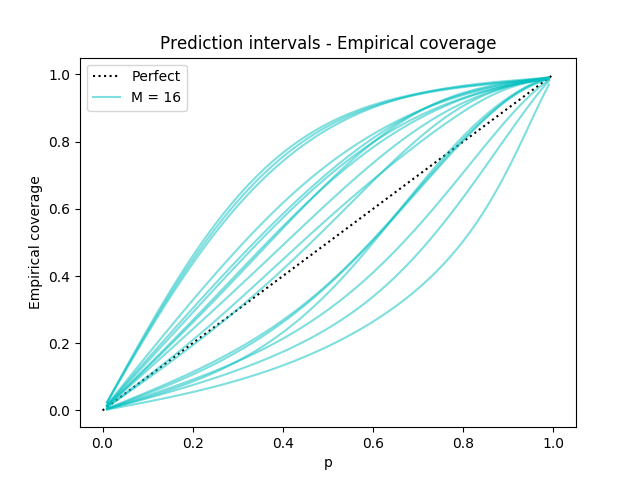}
         \caption{$M = 16$.}
     \end{subfigure}
        \caption{Results for MC-dropout on the KITTI depth completion validation dataset. Condensed calibration plots.}
        \label{fig:depth_completion_calibration_plots_mcdropout}
\end{figure}

Here, we show sparsification plots, sparsification error curves and calibration plots. Examples of sparsification plots are found in Figure~\ref{fig:depth_completion_sparsification_plots_ensembling} for ensembling and Figure~\ref{fig:depth_completion_sparsification_plots_mcdropout} for MC-dropout. Condensed sparsification error curves are found in Figure~\ref{fig:depth_completion_sparsification_error_curves_ensembling} for ensembling and Figure~\ref{fig:depth_completion_sparsification_error_curves_mcdropout} for MC-dropout. Condensed calibration plots are found in Figure~\ref{fig:depth_completion_calibration_plots_ensembling} for ensembling and Figure~\ref{fig:depth_completion_calibration_plots_mcdropout} for MC-dropout.

\clearpage
\section{Street-Scene Semantic Segmentation}
\label{appendix:street-scene_semantic_segmentation}

In this appendix, further details on the street-scene semantic segmentation experiments (Section~\ref{experiments-semantic_segmentation}) are provided.

\subsection{Training Details}
\label{appendix:street-scene_semantic_segmentation_training_details}

For ensembling, we train all ensemble models for $40\thinspace000$ steps with SGD + momentum ($0.9$), a batch size of $8$ and weight decay of $0.0005$. The learning rate $\alpha_t$ is decayed according to:
\begin{equation*}
    \alpha_t = \alpha_0(1 - \frac{t}{T})^{0.9}, \quad t = 1, 2, \dots, T,
\end{equation*}
where $T = 40\thinspace000$ and $\alpha_0 = 0.01$ (the initial learning rate). We train on randomly selected image crops of size $512 \times 512$. We choose a smaller crop size than Yuan and Wang~\cite{yuan2018ocnet} to enable an extensive evaluation with repeated experiments. The only other data augmentation used is random flipping along the vertical axis and random scaling in the range $[0.5, 1.5]$. The ResNet101 backbone is initialized with weights\footnote{\url{http://sceneparsing.csail.mit.edu/model/pretrained_resnet/resnet101-imagenet.pth}.} from a model pretrained on the ImageNet dataset, all other model parameters are randomly initialized using the default initializer in PyTorch. Models are trained on two NVIDIA TITAN Xp GPUs with $12$GB of RAM each. For MC-dropout, models are instead trained for $60\thinspace000$ steps.

\subsection{Description of Results}
\label{appendix:street-scene_semantic_segmentation_results_description}

The results in Figure~\ref{fig:segmentation_ause_ece_miou} (Section~\ref{experiments-semantic_segmentation}) were obtained in the following way:
\begin{itemize}
    \item \textbf{Ensembling:} $26$ models were trained using the same training procedure, the mean and standard deviation was computed based on $8$ sets of randomly drawn models for $M~\in~\{1, 2, 4, 8, 16\}$. The same set could not be drawn more than once.  
    
    \item \textbf{MC-dropout:} $8$ models were trained using the same training procedure, based on which the mean and standard deviation was computed. 
\end{itemize}

\subsection{Additional Results}
\label{appendix:street-scene_semantic_segmentation_additional_results}

\begin{figure}
     \centering
     \begin{subfigure}[b]{0.49\textwidth}
         \centering
         \includegraphics[width=\textwidth]{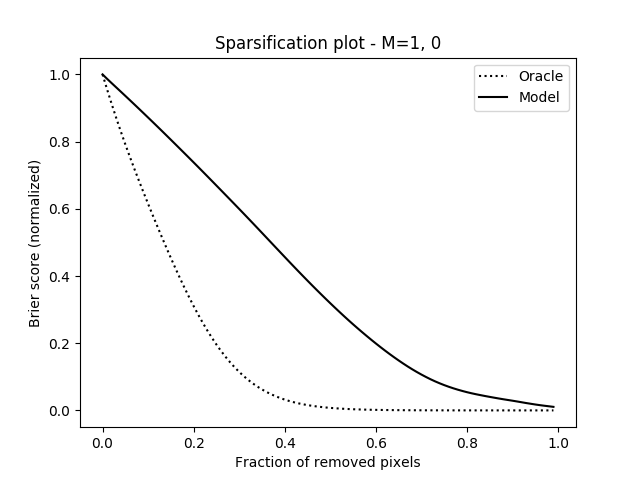}
         \caption{$M = 1$.}
     \end{subfigure}
     \begin{subfigure}[b]{0.49\textwidth}
         \centering
         \includegraphics[width=\textwidth]{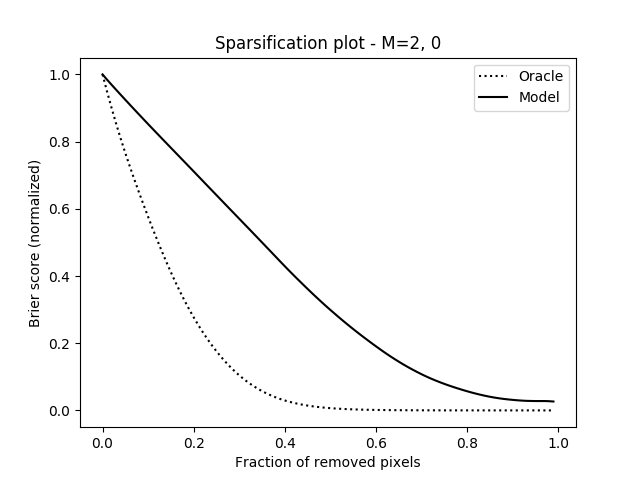}
         \caption{$M = 2$.}
     \end{subfigure}
     \begin{subfigure}[b]{0.49\textwidth}
         \centering
         \includegraphics[width=\textwidth]{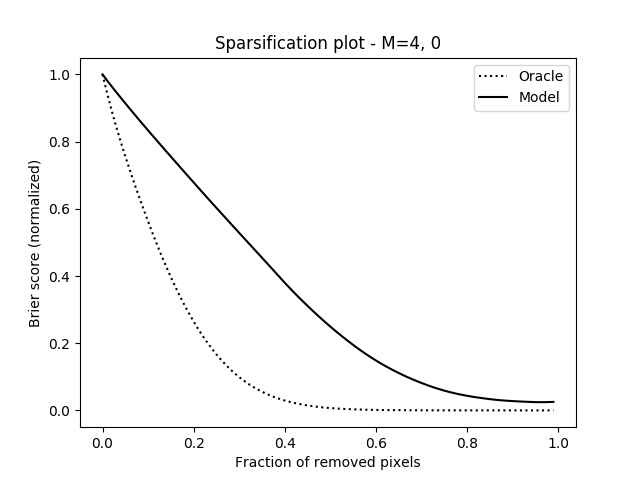}
         \caption{$M = 4$.}
     \end{subfigure}
     \begin{subfigure}[b]{0.49\textwidth}
         \centering
         \includegraphics[width=\textwidth]{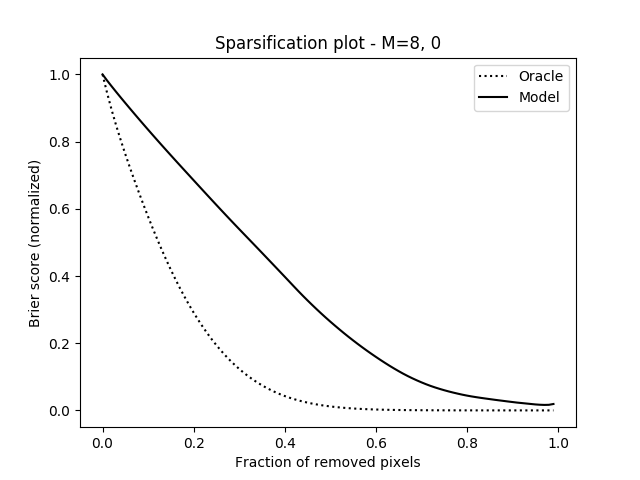}
         \caption{$M = 8$.}
     \end{subfigure}
     \begin{subfigure}[b]{0.49\textwidth}
         \centering
         \includegraphics[width=\textwidth]{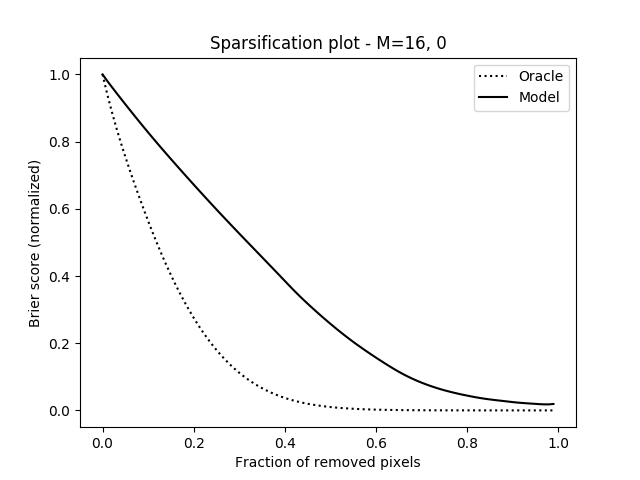}
         \caption{$M = 16$.}
     \end{subfigure}
        \caption{Results for ensembling on the Cityscapes validation dataset. Examples of sparsification plots.}
        \label{fig:segmentation_sparsification_plots_ensembling}
\end{figure}

\begin{figure}
     \centering
     \begin{subfigure}[b]{0.49\textwidth}
         \centering
         \includegraphics[width=\textwidth]{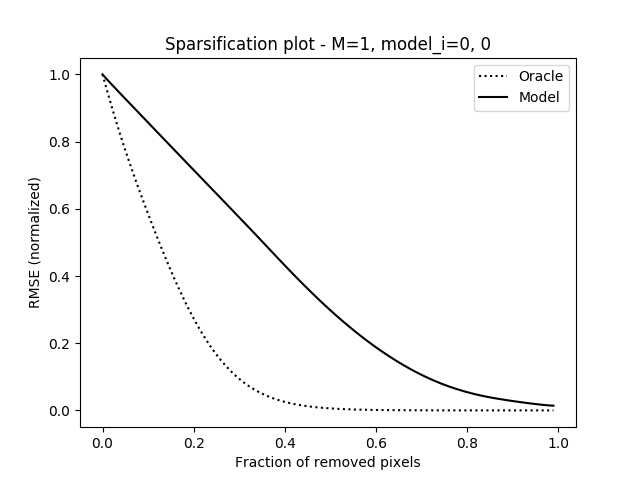}
         \caption{$M = 1$.}
     \end{subfigure}
     \begin{subfigure}[b]{0.49\textwidth}
         \centering
         \includegraphics[width=\textwidth]{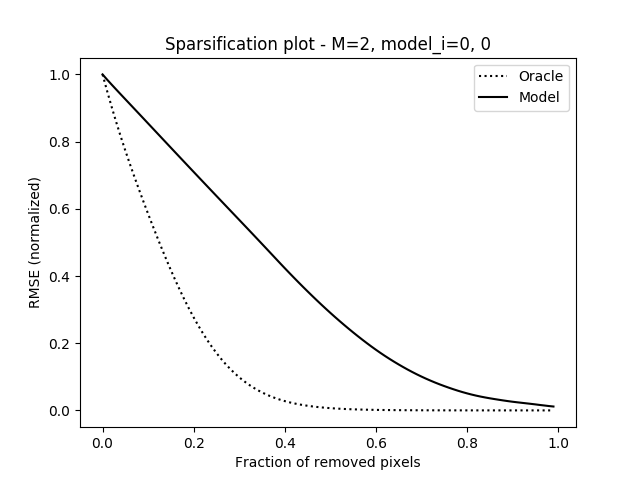}
         \caption{$M = 2$.}
     \end{subfigure}
     \begin{subfigure}[b]{0.49\textwidth}
         \centering
         \includegraphics[width=\textwidth]{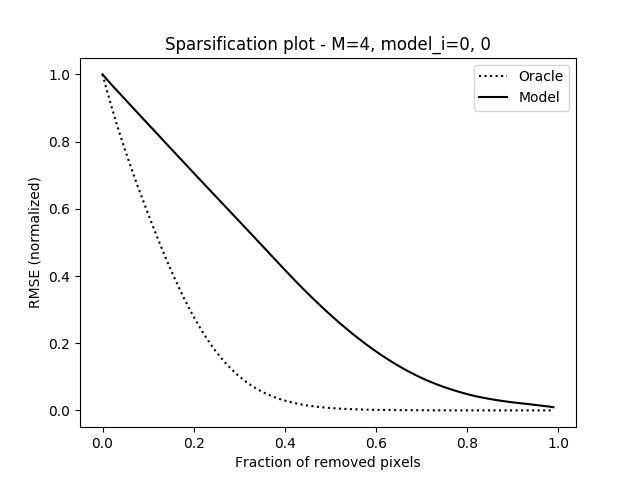}
         \caption{$M = 4$.}
     \end{subfigure}
     \begin{subfigure}[b]{0.49\textwidth}
         \centering
         \includegraphics[width=\textwidth]{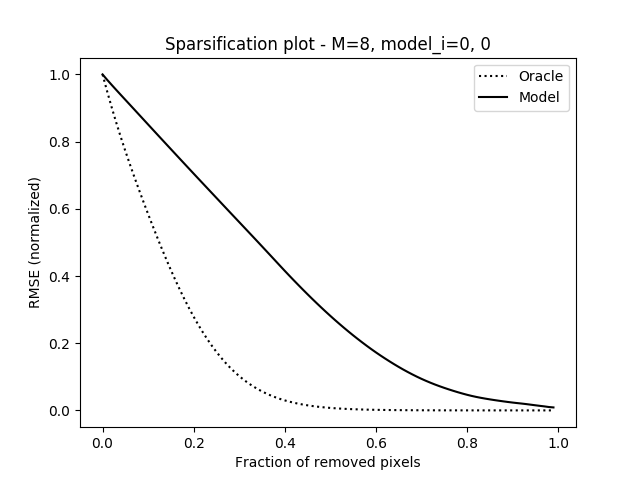}
         \caption{$M = 8$.}
     \end{subfigure}
     \begin{subfigure}[b]{0.49\textwidth}
         \centering
         \includegraphics[width=\textwidth]{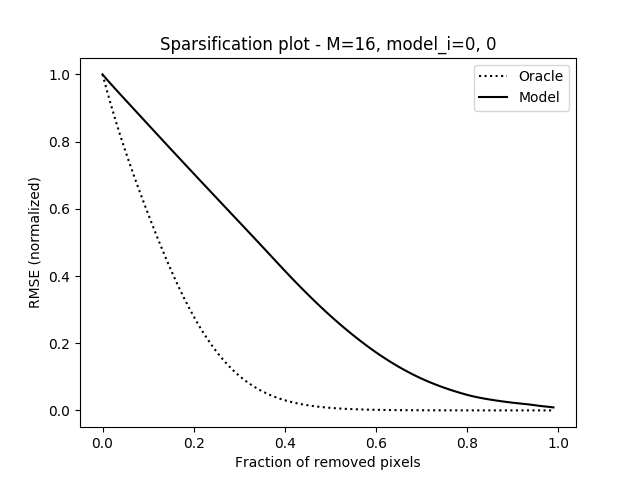}
         \caption{$M = 16$.}
     \end{subfigure}
        \caption{Results for MC-dropout on the Cityscapes validation dataset. Examples of sparsification plots.}
        \label{fig:segmentation_sparsification_plots_mcdropout}
\end{figure}

\begin{figure}
     \centering
     \begin{subfigure}[b]{0.49\textwidth}
         \centering
         \includegraphics[width=\textwidth]{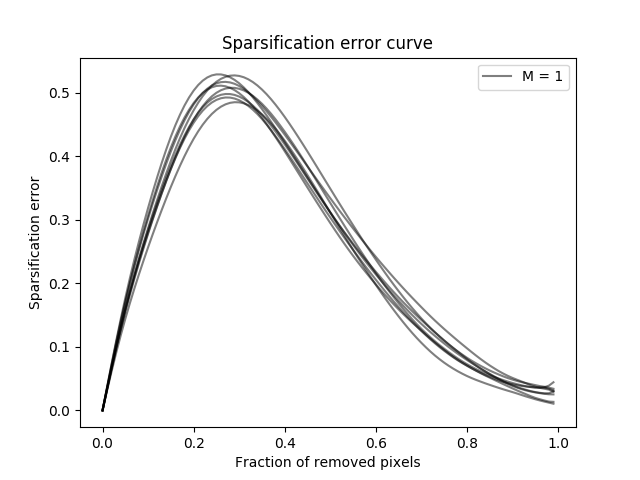}
         \caption{$M = 1$.}
     \end{subfigure}
     \begin{subfigure}[b]{0.49\textwidth}
         \centering
         \includegraphics[width=\textwidth]{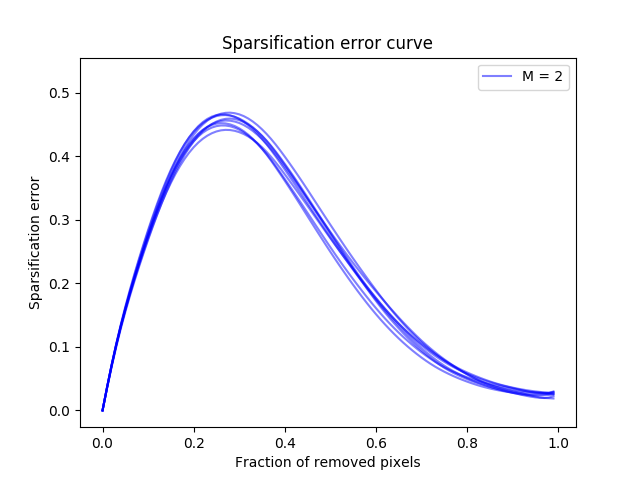}
         \caption{$M = 2$.}
     \end{subfigure}
     \begin{subfigure}[b]{0.49\textwidth}
         \centering
         \includegraphics[width=\textwidth]{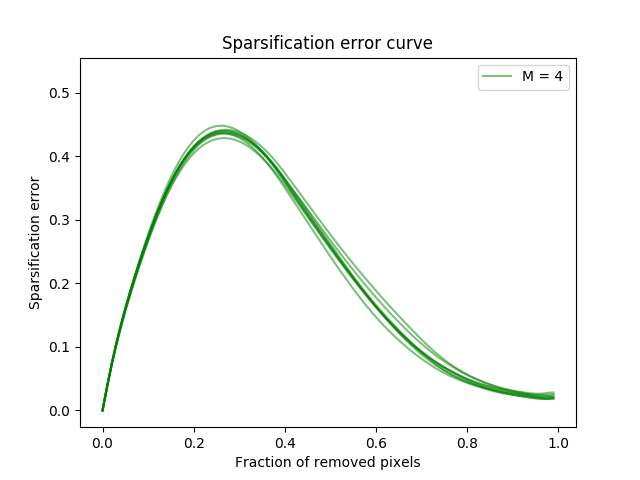}
         \caption{$M = 4$.}
     \end{subfigure}
     \begin{subfigure}[b]{0.49\textwidth}
         \centering
         \includegraphics[width=\textwidth]{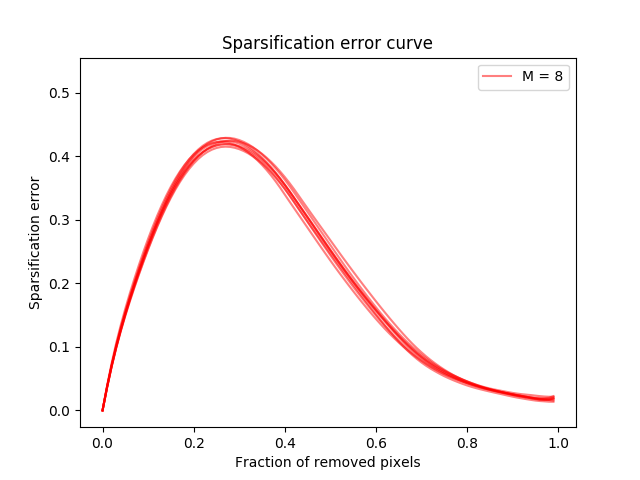}
         \caption{$M = 8$.}
     \end{subfigure}
     \begin{subfigure}[b]{0.49\textwidth}
         \centering
         \includegraphics[width=\textwidth]{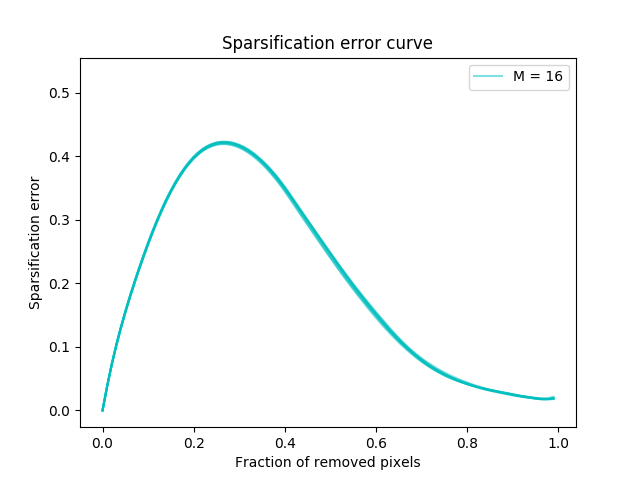}
         \caption{$M = 16$.}
     \end{subfigure}
        \caption{Results for ensembling on the Cityscapes validation dataset. Condensed sparsification error curves.}
        \label{fig:segmentation_sparsification_error_curves_ensembling}
\end{figure}

\begin{figure}
     \centering
     \begin{subfigure}[b]{0.49\textwidth}
         \centering
         \includegraphics[width=\textwidth]{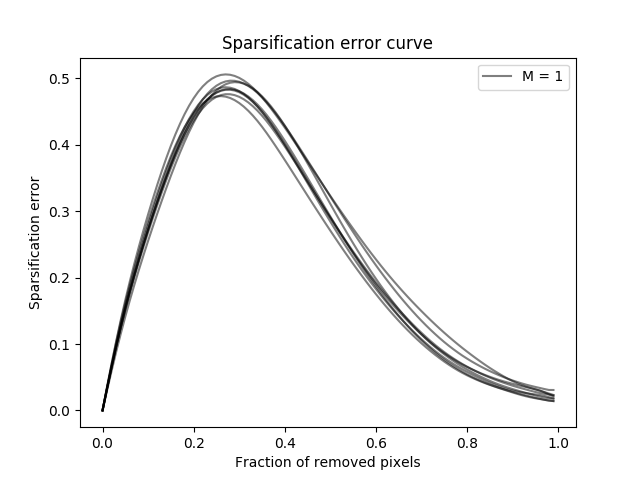}
         \caption{$M = 1$.}
     \end{subfigure}
     \begin{subfigure}[b]{0.49\textwidth}
         \centering
         \includegraphics[width=\textwidth]{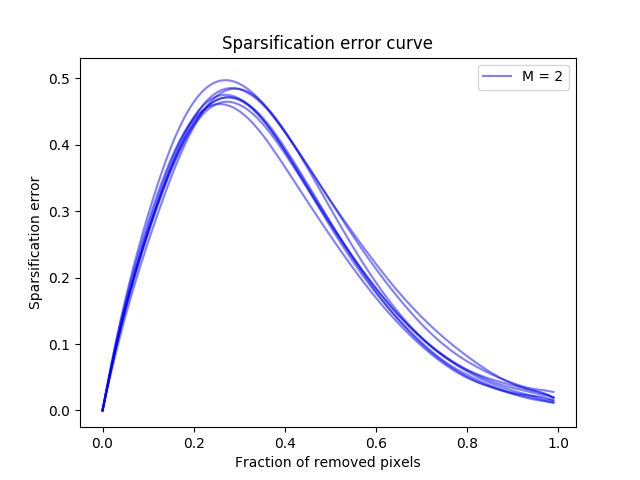}
         \caption{$M = 2$.}
     \end{subfigure}
     \begin{subfigure}[b]{0.49\textwidth}
         \centering
         \includegraphics[width=\textwidth]{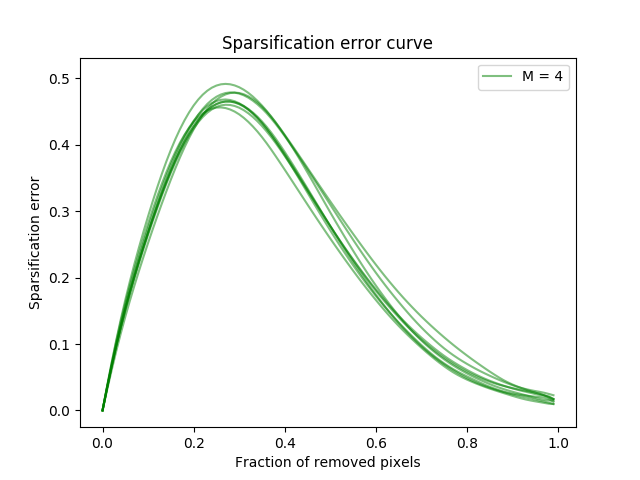}
         \caption{$M = 4$.}
     \end{subfigure}
     \begin{subfigure}[b]{0.49\textwidth}
         \centering
         \includegraphics[width=\textwidth]{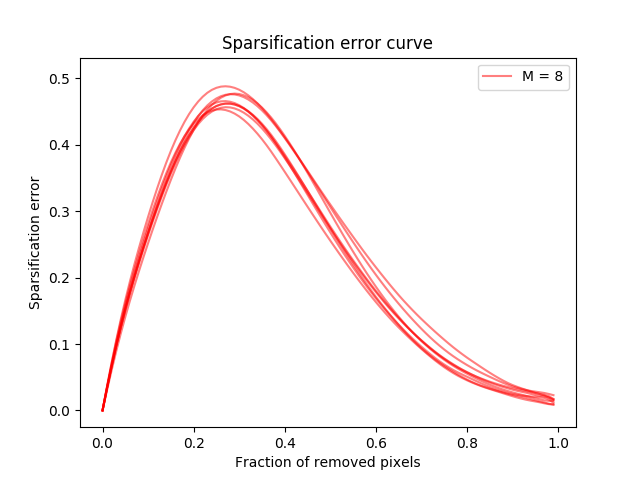}
         \caption{$M = 8$.}
     \end{subfigure}
     \begin{subfigure}[b]{0.49\textwidth}
         \centering
         \includegraphics[width=\textwidth]{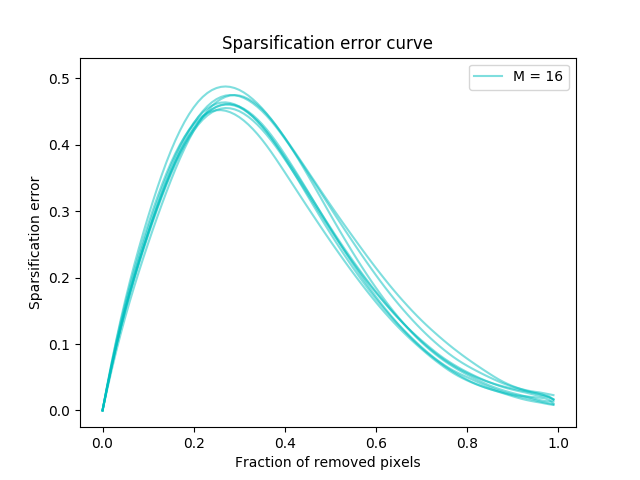}
         \caption{$M = 16$.}
     \end{subfigure}
        \caption{Results for MC-dropout on the Cityscapes validation dataset. Condensed sparsification error curves.}
        \label{fig:segmentation_sparsification_error_curves_mcdropout}
\end{figure}

\begin{figure}
     \centering
     \begin{subfigure}[b]{0.49\textwidth}
         \centering
         \includegraphics[width=\textwidth]{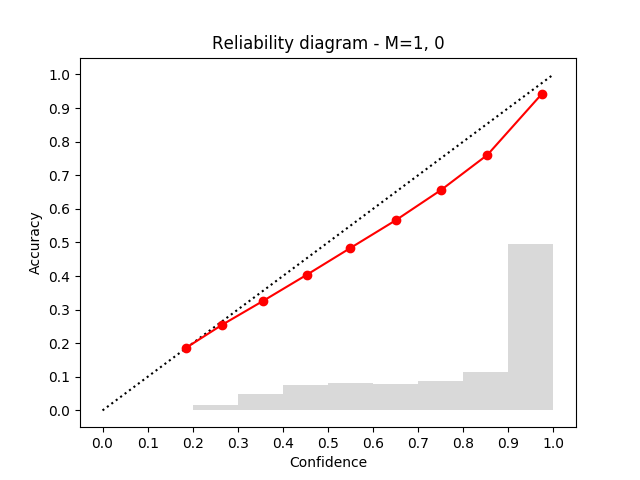}
         \caption{$M = 1$.}
     \end{subfigure}
     \begin{subfigure}[b]{0.49\textwidth}
         \centering
         \includegraphics[width=\textwidth]{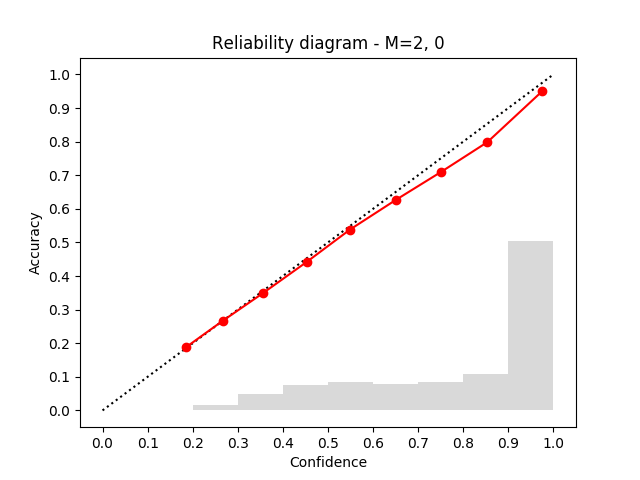}
         \caption{$M = 2$.}
     \end{subfigure}
     \begin{subfigure}[b]{0.49\textwidth}
         \centering
         \includegraphics[width=\textwidth]{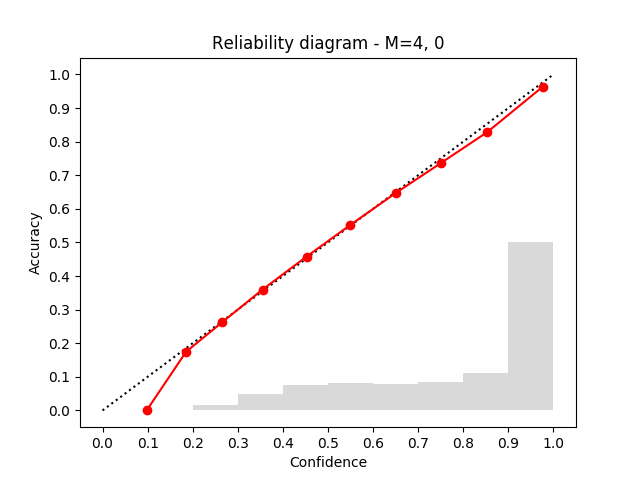}
         \caption{$M = 4$.}
     \end{subfigure}
     \begin{subfigure}[b]{0.49\textwidth}
         \centering
         \includegraphics[width=\textwidth]{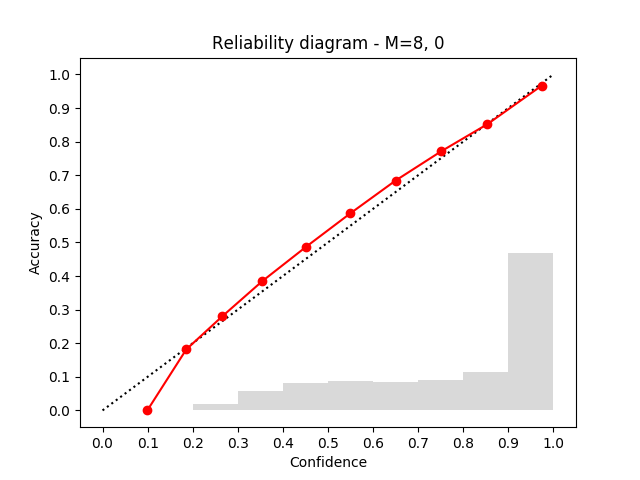}
         \caption{$M = 8$.}
     \end{subfigure}
     \begin{subfigure}[b]{0.49\textwidth}
         \centering
         \includegraphics[width=\textwidth]{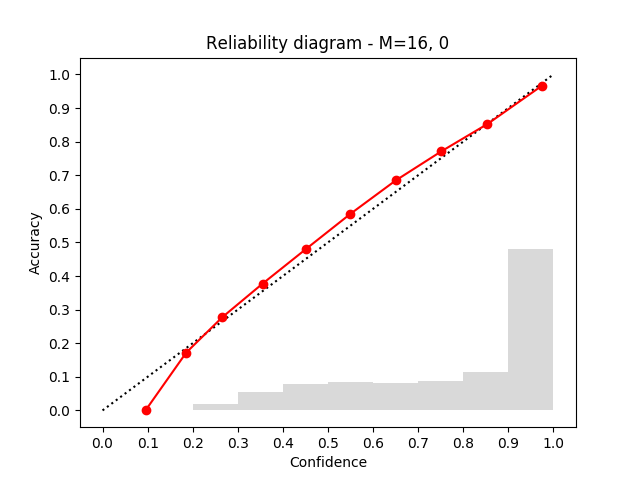}
         \caption{$M = 16$.}
     \end{subfigure}
        \caption{Results for ensembling on the Cityscapes validation dataset. Examples of reliability diagrams with histograms.}
        \label{fig:segmentation_rel_diagrams_ensembling}
\end{figure}

\begin{figure}
     \centering
     \begin{subfigure}[b]{0.49\textwidth}
         \centering
         \includegraphics[width=\textwidth]{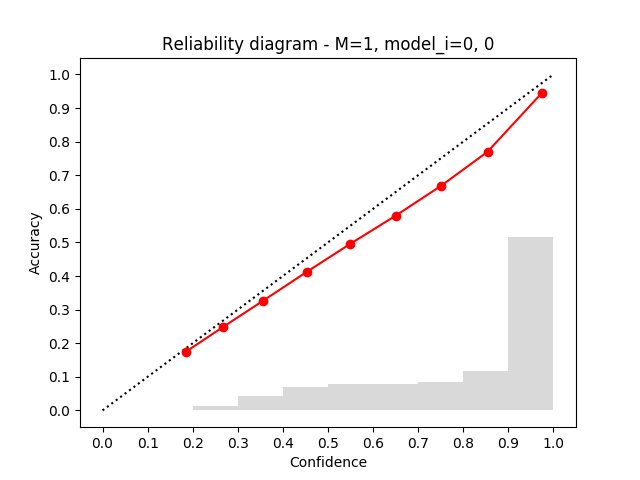}
         \caption{$M = 1$.}
     \end{subfigure}
     \begin{subfigure}[b]{0.49\textwidth}
         \centering
         \includegraphics[width=\textwidth]{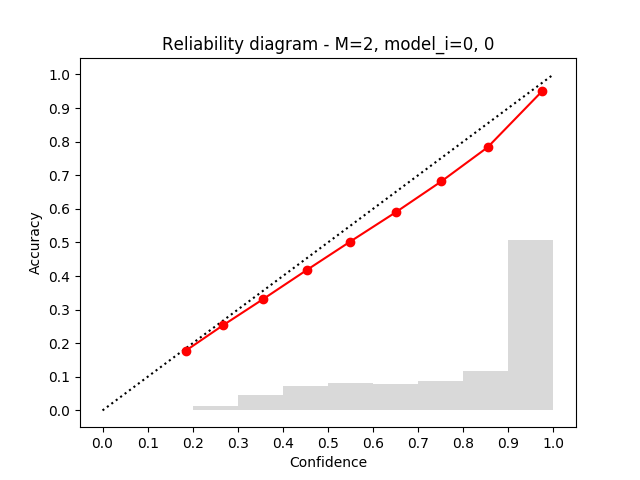}
         \caption{$M = 2$.}
     \end{subfigure}
     \begin{subfigure}[b]{0.49\textwidth}
         \centering
         \includegraphics[width=\textwidth]{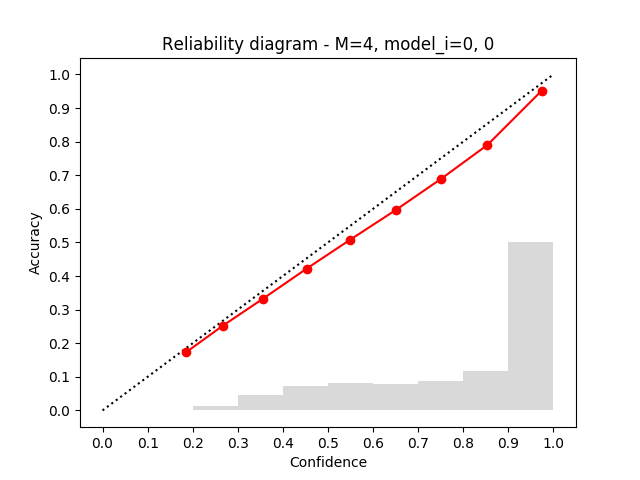}
         \caption{$M = 4$.}
     \end{subfigure}
     \begin{subfigure}[b]{0.49\textwidth}
         \centering
         \includegraphics[width=\textwidth]{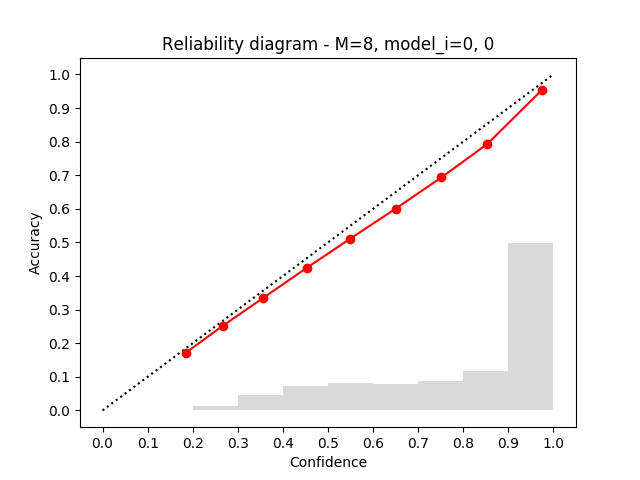}
         \caption{$M = 8$.}
     \end{subfigure}
     \begin{subfigure}[b]{0.49\textwidth}
         \centering
         \includegraphics[width=\textwidth]{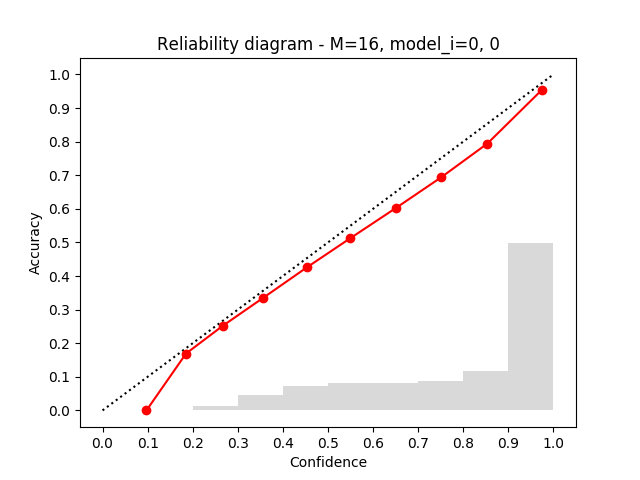}
         \caption{$M = 16$.}
     \end{subfigure}
        \caption{Results for MC-dropout on the Cityscapes validation dataset. Examples of reliability diagrams with histograms.}
        \label{fig:segmentation_rel_diagrams_mcdropout}
\end{figure}

\begin{figure}
     \centering
     \begin{subfigure}[b]{0.49\textwidth}
         \centering
         \includegraphics[width=\textwidth]{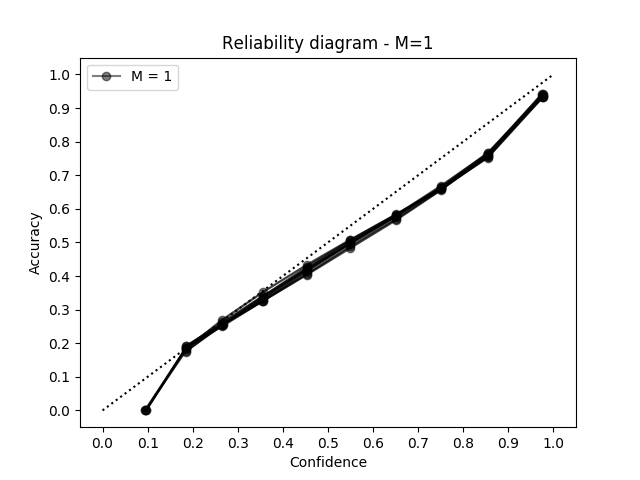}
         \caption{$M = 1$.}
     \end{subfigure}
     \begin{subfigure}[b]{0.49\textwidth}
         \centering
         \includegraphics[width=\textwidth]{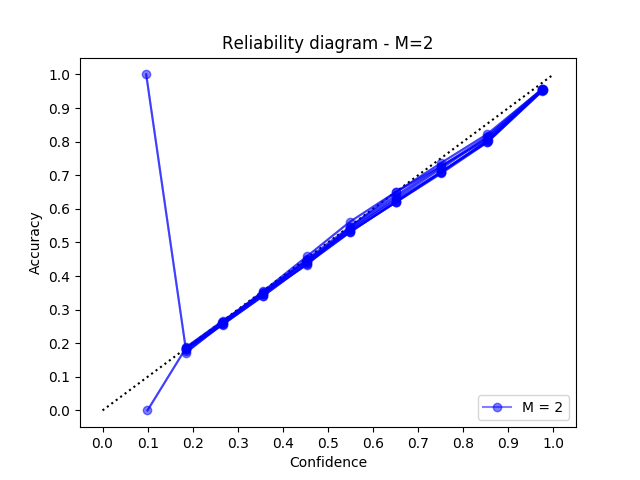}
         \caption{$M = 2$.}
     \end{subfigure}
     \begin{subfigure}[b]{0.49\textwidth}
         \centering
         \includegraphics[width=\textwidth]{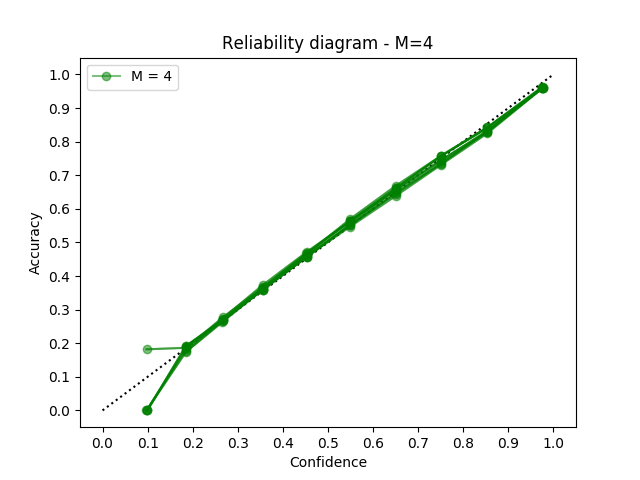}
         \caption{$M = 4$.}
     \end{subfigure}
     \begin{subfigure}[b]{0.49\textwidth}
         \centering
         \includegraphics[width=\textwidth]{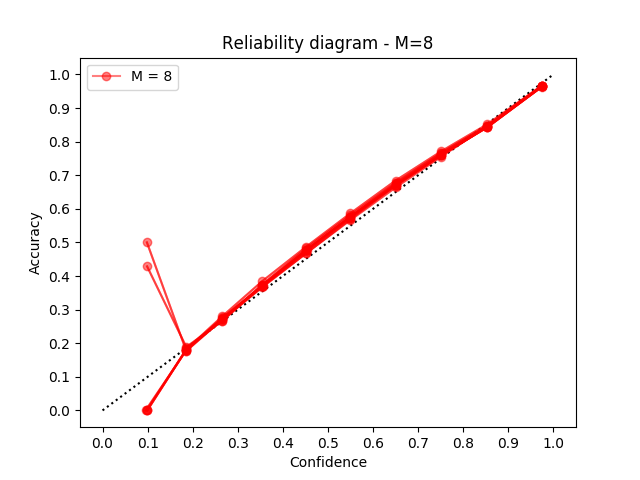}
         \caption{$M = 8$.}
     \end{subfigure}
     \begin{subfigure}[b]{0.49\textwidth}
         \centering
         \includegraphics[width=\textwidth]{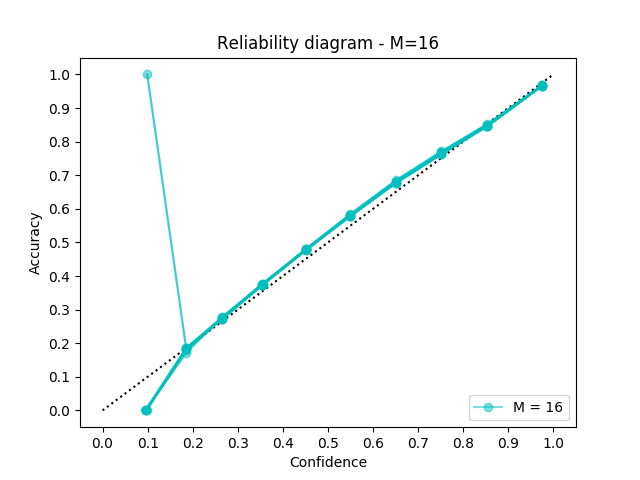}
         \caption{$M = 16$.}
     \end{subfigure}
        \caption{Results for ensembling on the Cityscapes validation dataset. Condensed reliability diagrams.}
        \label{fig:segmentation_condensed_rel_diagrams_ensembling}
\end{figure}

\begin{figure}
     \centering
     \begin{subfigure}[b]{0.49\textwidth}
         \centering
         \includegraphics[width=\textwidth]{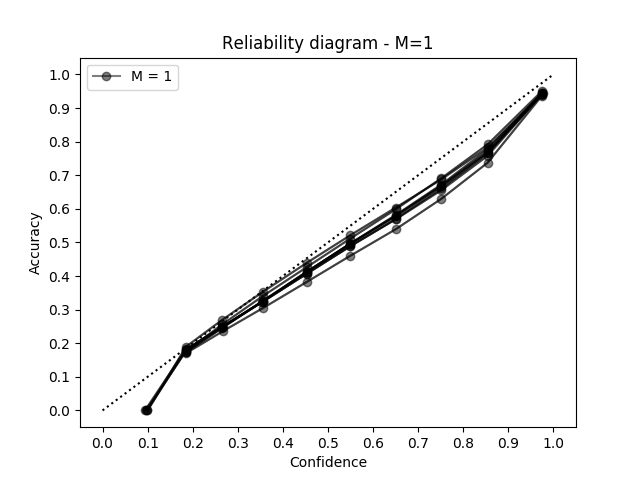}
         \caption{$M = 1$.}
     \end{subfigure}
     \begin{subfigure}[b]{0.49\textwidth}
         \centering
         \includegraphics[width=\textwidth]{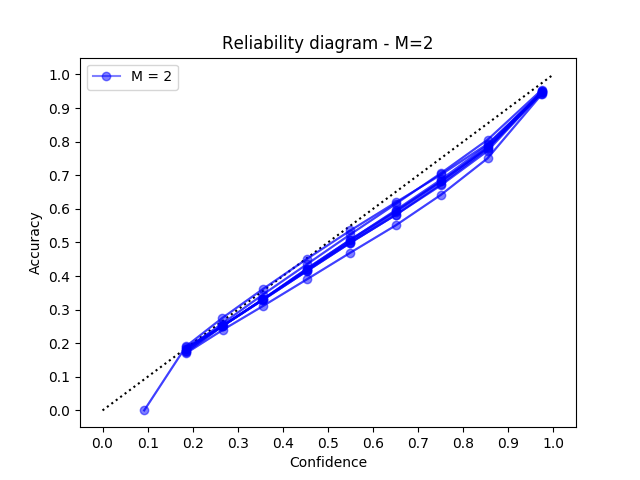}
         \caption{$M = 2$.}
     \end{subfigure}
     \begin{subfigure}[b]{0.49\textwidth}
         \centering
         \includegraphics[width=\textwidth]{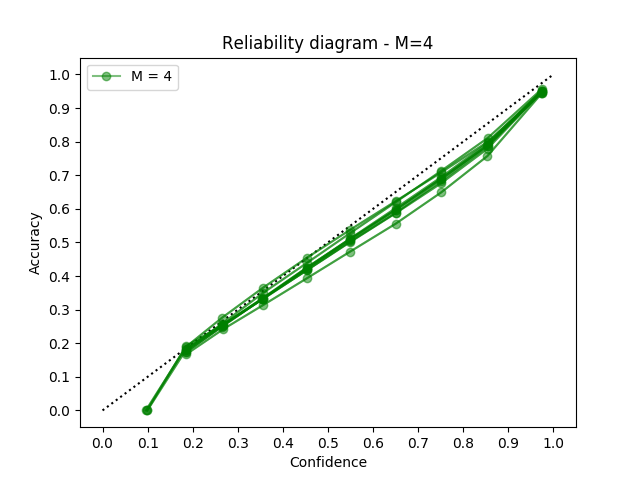}
         \caption{$M = 4$.}
     \end{subfigure}
     \begin{subfigure}[b]{0.49\textwidth}
         \centering
         \includegraphics[width=\textwidth]{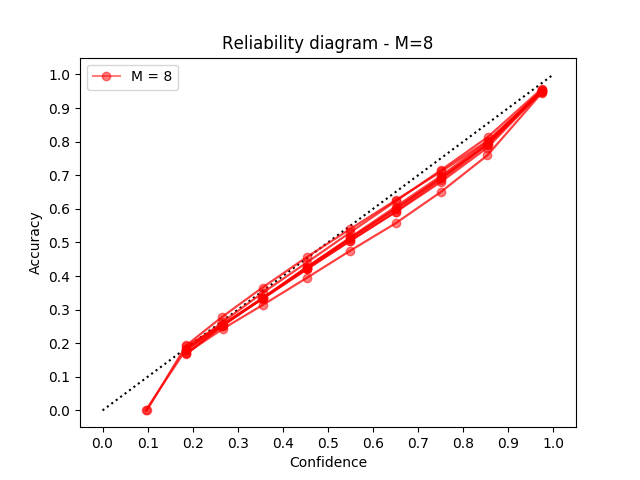}
         \caption{$M = 8$.}
     \end{subfigure}
     \begin{subfigure}[b]{0.49\textwidth}
         \centering
         \includegraphics[width=\textwidth]{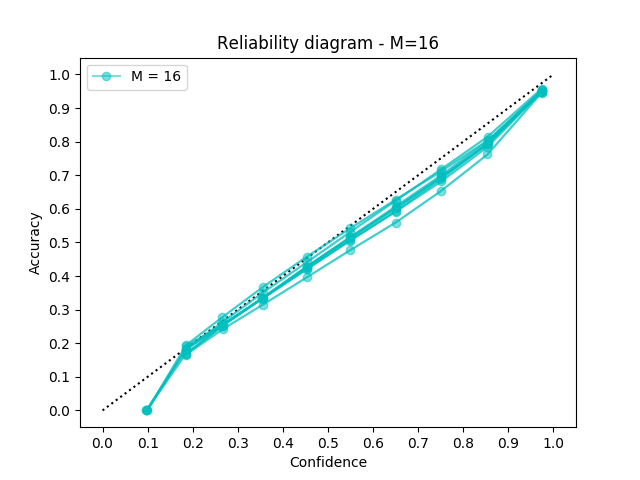}
         \caption{$M = 16$.}
     \end{subfigure}
        \caption{Results for MC-dropout on the Cityscapes validation dataset. Condensed reliability diagrams.}
        \label{fig:segmentation_condensed_rel_diagrams_mcdropout}
\end{figure}

Here, we show sparsification plots, sparsification error curves and reliability diagrams. Examples of sparsification plots are found in Figure~\ref{fig:segmentation_sparsification_plots_ensembling} for ensembling and Figure~\ref{fig:segmentation_sparsification_plots_mcdropout} for MC-dropout. Condensed sparsification error curves are found in Figure~\ref{fig:segmentation_sparsification_error_curves_ensembling} for ensembling and Figure~\ref{fig:segmentation_sparsification_error_curves_mcdropout} for MC-dropout. Examples of reliability diagrams with histograms are found in Figure~\ref{fig:segmentation_rel_diagrams_ensembling} for ensembling and Figure~\ref{fig:segmentation_rel_diagrams_mcdropout} for MC-dropout. Condensed reliability diagrams are found in Figure~\ref{fig:segmentation_condensed_rel_diagrams_ensembling} for ensembling and Figure~\ref{fig:segmentation_condensed_rel_diagrams_mcdropout} for MC-dropout.

\end{appendices}

\end{document}